\documentclass[acmlarge]{acmart}

\AtBeginDocument{%
  }

\renewcommand\footnotetextcopyrightpermission[1]{}
\usepackage[table]{xcolor}
\usepackage{tikz}
\usepackage{graphicx}
\graphicspath{{}{ubicomp_package/source/}}
\usepackage{subcaption}
\usepackage{multirow}
\usepackage{wrapfig}
\makeatletter
\AtBeginEnvironment{wrapfigure}{\let\@vspace\@vspace@orig\let\@vspacer\@vspacer@orig}
\makeatother
\usepackage{amsmath}
\usepackage{microtype}

\usepackage{amssymb}
\makeatletter
\def\input@path{{}{ubicomp_package/source/}}
\makeatother

\begin{document}

\title{Peak-Detector: Explainable Peak Detection via Instruction–Tuned Large Language Models in Physiological Signal}

\author{Jiahui Li}
\email{jl57095@uga.edu}
\orcid{0009-0007-5129-0107}
\affiliation{%
  \institution{University of Georgia}
  \city{Athens}
  \state{Georgia}
  \country{USA}
}

\author{Yida Zhang}
\orcid{0009-0000-5555-3778}
\affiliation{%
  \institution{University of Georgia}
  \city{Athens}
  \state{Georgia}
  \country{USA}}
\email{yida.zhang@uga.edu}

\author{Zixuan Zeng}
\orcid{0009-0005-7051-7421}
\affiliation{%
  \institution{University of Georgia}
  \city{Athens}
  \state{Georgia}
  \country{USA}
}
\email{zxzeng@uga.edu}

\author{Jiayu Chen}
\orcid{0009-0003-0904-4114}
\affiliation{%
 \institution{University of Georgia}
 \city{Athens}
 \state{Georgia}
 \country{USA}}
\email{jiayu.chen@uga.edu}

\author{Yingjian Song}
\orcid{0009-0005-5601-4465}
\affiliation{%
 \institution{University of Georgia}
 \city{Athens}
 \state{Georgia}
 \country{USA}}
\email{ys63016@uga.edu}

\author{Yin Xiao}
\orcid{0009-0005-2247-8042}
\affiliation{%
 \institution{Yixing People’s Hospital}
 \city{Yixing}
 \state{Jiangsu Province}
 \country{China}}
\email{yinxiaosn@163.com}

\author{Nishan Dong}
\affiliation{%
 \institution{Yixing People’s Hospital}
 \city{Yixing}
 \state{Jiangsu Province}
 \country{China}}
\email{staff1820@yxph.com}
\orcid{0009-0006-4463-9246}

\author{Junjie Lu}
\orcid{0000-0001-7547-6378}
\affiliation{%
 \institution{Yixing People’s Hospital}
 \city{Yixing}
 \state{Jiangsu Province}
 \country{China}}
\email{staff806@yxph.com}

\author{Younghoon Kwon}
\orcid{0000-0002-8152-9170}
\affiliation{%
 \institution{University of Washington}
 \city{Seattle}
 \state{Washington}
 \country{USA}}
\email{yhkwon@uw.edu}

\author{Xiang Zhang}
\orcid{0000-0001-5097-2113}
\affiliation{%
 \institution{University of North Carolina at Charlotte}
 \city{Charlotte}
 \state{North Carolina}
 \country{USA}}
\email{xiang.zhang@charlotte.edu}

\author{Jin Lu}
\orcid{0000-0003-1356-0202}
\affiliation{%
 \institution{University of Georgia}
 \city{Athens}
 \state{Georgia}
 \country{USA}}
\email{jin.lu@uga.edu}

\author{Wenzhan Song}
\orcid{0000-0001-8174-1772}
\affiliation{%
 \institution{University of Georgia}
 \city{Athens}
 \state{Georgia}
 \country{USA}}
\email{wsong@uga.edu}

\author{Fei Dou}
\authornote{Corresponding author.}
\orcid{0000-0003-4246-8616}
\affiliation{%
 \institution{University of Georgia}
 \city{Athens}
 \state{Georgia}
 \country{USA}}
\email{fei.dou@uga.edu}

\newcommand{\std}[1]{{\fontsize{7}{8.2}\selectfont$\pm$ #1}}
\newcommand{\val}[2]{#1 \std{#2}}
\newcommand{\valsig}[3]{#1\textsuperscript{#3} \std{#2}}

\renewcommand{\shortauthors}{Li et al.}

\setcopyright{cc}
\setcctype{by}
\begin{abstract}
    Accurate peak detection across diverse cardiac physiological signals, including the Electrocardiogram (ECG), Photoplethysmogram (PPG), Ballistocardiogram (BCG), and Bodyseismography (BSG), is fundamental for cardiovascular monitoring but is often hindered by artifacts and signal variability. Conventional algorithms are typically engineered with expert knowledge for a single signal modality, limiting their generalizability. Conversely, deep learning-based methods often lack interpretability, limiting transparency for expert verification and hindering expert--computer interaction. To address these limitations, we introduce Peak-Detector, a novel framework that 
    leverages instruction-tuned Large Language Models (LLMs) for robust, cross-modal, and explainable peak detection. A core innovation of our framework is a "peak-representation" technique that transforms time-series data into a condensed format, preserving critical event information while significantly reducing signal length. This representation provides a crucial inductive bias, guiding the LLM to reason over physiologically meaningful events rather than raw, noisy data. The model is optimized through a two-stage process: supervised fine-tuning (SFT) followed by reinforcement learning (RL) with a multi-objective reward function. The model's self-explanation capabilities are cultivated by fine-tuning on a custom-built Peak-Explanation dataset. Across four modalities—ECG, PPG, BCG, and BSG—spanning seven datasets (six public benchmarks plus one real-world cohort), Peak-Detector demonstrates strong cross-modal performance, achieving best or tied-best detection under clinically relevant temporal tolerance. Beyond accuracy, the generated rationales surface failure modes and support verification and error analysis. Together, these results indicate a transparent and generalizable framework for trustworthy peak analysis and cardiovascular metric extraction.
    
\end{abstract}




\maketitle

\section{INTRODUCTION}

Continuous cardiovascular monitoring is an important component of mobile and ubiquitous health, supporting the longitudinal assessment of cardiac function across wearable, contactless, and clinical sensing settings. Unobtrusive sensing modalities, particularly the Electrocardiogram (ECG), Photoplethysmogram (PPG), Ballistocardiogram (BCG), and Bodyseismography (BSG), have become cornerstone technologies for the longitudinal assessment of an individual's health status~\citep{gordon1877certain,kim2016ballistocardiogram,song2024real,penzel2016modulations,temko2017accurate}. The ability of these sensors to facilitate timely identification of cardiac arrhythmias—such as Premature Atrial Contractions (PACs), Premature Ventricular Contractions (PVCs), and Atrial Fibrillation (AFib)—is critical for preventing adverse health outcomes. However, the efficacy of these monitoring systems hinges on the accurate and robust detection of key prominent points, or "peaks," within the collected physiological data.

\begin{wrapfigure}{r}{0.38\textwidth}
        \centering
        \includegraphics[width=\linewidth]{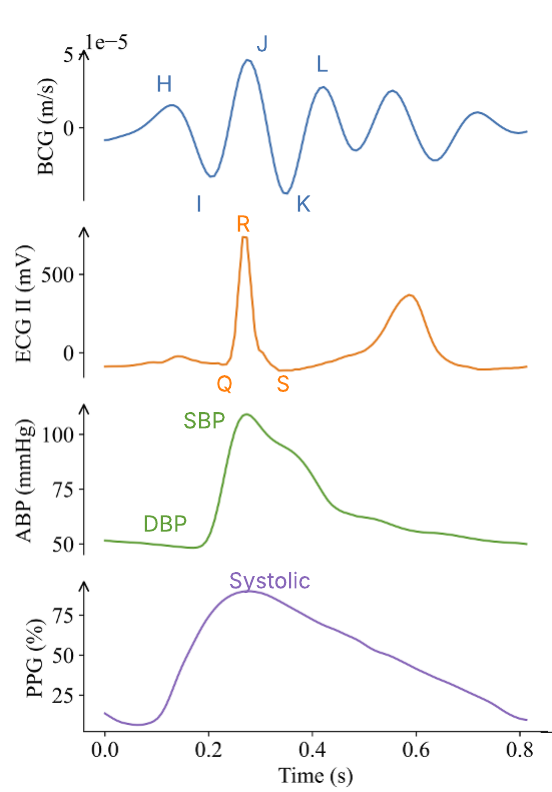}
        \caption{Physiology signals with labelled peaks}
        \Description{Annotated physiological signal examples with marked fiducial peaks.}
        \label{fig:physiological_signals}
\end{wrapfigure}

The primary challenge in this domain lies in the inherent heterogeneity of these signal modalities. Each captures a different facet of the cardiac cycle through distinct physical principles, resulting in unique signal morphologies (Fig~\ref{fig:physiological_signals}).  
The ECG records the heart's electrical activity, with the prominent R-peak signifying peak ventricular depolarization~\citep{setiawidayat2018new}. The PPG waveform features a systolic peak, which represents the point of maximum blood volume in the peripheral tissue during ventricular contraction~\cite{castaneda2018review}. The BCG provides a contactless method by measuring the body's subtle vibrations from cardiac ejections, characterized by a prominent J-peak~\citep{li2024ballistocardiogram,su2009ballistocardiogram,azhaginiyan2019denoising}. The BSG delivers richer information content by effectively containing features of both BCG (whole-body movement) and SCG (direct precordial vibrations)~\cite{song2024real,pitafi2025contactless,song2025multi}. This diversity in signal origin and quality necessitates a versatile and resilient approach to peak detection~\citep{warmerdam2018hierarchical,fung2004continuous}.

Existing peak-detection methods remain limited in this regard.
Traditional signal-processing approaches rely on modality-specific domain knowledge, employing handcrafted features and heuristics that are often brittle and require meticulous parameter tuning~\citep{chakraborty2020robust,elgendi2013systolic,kuntamalla2014efficient}. Consequently, these methods lack generalizability across different signal types. 
While deep learning models offer greater adaptability by learning features directly from data, they are often criticized as "black boxes."~\cite{kazemi2022robust,sarkar2021cardiogan,xu2024medical,castelvecchi2016can}. This lack of transparency and interpretability creates a significant barrier to trust and adoption in critical clinical applications, where understanding the model's reasoning is as important as its output.

To address the limitations mentioned, we introduce Peak-Detector, a framework that reformulates cardiac peak detection as a language-guided reasoning task using LLMs. By leveraging the advanced reasoning and linguistic capabilities of LLMs \cite{bi2024deepseek, wang2024llm}, our approach utilizes a novel Peak Representation that converts sparse, high-frequency physiological signals into condensed symbolic sequences. This shift effectively addresses the performance degradation typically associated with processing long, continuous numerical sequences \cite{fons2024evaluating}. To bolster interpretability, we developed a Peak-Explanation Dataset via a specialized data generation pipeline designed to enhance the model’s self-explanatory capabilities.
The model is trained using a two-stage instruction-tuning strategy. The first stage, Supervised Fine-Tuning (SFT), establishes a reliable and correctly formatted output structure. The second stage employs Reinforcement Learning (RL) with Group Relative Policy Optimization (GRPO), optimizing the model against a multi-objective reward function. This function is designed to concurrently enhance format validity, heart-rate consistency, positional accuracy, and detection completeness, ensuring a robust and physiologically consistent output. 

We position \textit{Peak-Detector} not as a strict real-time, always-on edge model for battery-constrained wearables, but as an explainable and higher-fidelity analysis component within broader cardiovascular sensing workflows. Under this framing, lightweight front-end methods may be used for continuous screening, while \textit{Peak-Detector} can be invoked selectively for flagged windows, uncertain segments, degraded signals, or retrospective summaries where interpretability and auditability are especially important. This perspective is particularly relevant in ubiquitous sensing pipelines, where deployment decisions must balance signal quality, compute budget, and the need for trustworthy downstream metric extraction.

In summary, this work makes the following contributions:

\begin{itemize}
    \item We introduce a modality-agnostic \textbf{Peak Representation} that summarizes physiological signals as timestamped local extrema with signal value, enabling token-efficient, auditable reasoning. It effectively compresses raw ECG, PPG, and BCG data by 87\%, 97\%, and 89\%, respectively, while maintaining high fidelity for signal reconstruction with a correlation coefficient of 0.94, 0.94, and 0.97.

    \item We develop \textbf{Peak-Detector}, A two-stage SFT$\rightarrow$RL pipeline with a multi-objective reward jointly optimizes detection accuracy, temporal consistency, and concise, structured rationales, which is supported by a scalable data-generation pipeline that builds a self-explainable peak-detection dataset.

    \item Across ECG/PPG/BCG/BSG on six public datasets and one real-world cohort, Peak-Detector achieves best-or-tied-best detection under clinically relevant tolerance with competitive HR/HRV errors, while providing step-by-step rationales.
\end{itemize}

\section{RELATED WORK}
\label{sec:relatedwork}

\subsection{Peak Detection in Cardiac Physiological Signals}

The detection of fiducial points in cardiac physiological signals has primarily developed along two methodological lines: traditional signal-processing approaches and data-driven learning approaches.

\textbf{Signal-Processing Approaches.} Classical methods rely on modality-specific domain knowledge and handcrafted heuristics. For ECG R-peak detection, the Pan--Tompkins algorithm remains a foundational approach, using bandpass filtering, differentiation, squaring, and moving-window integration to isolate the QRS complex~\cite{fariha2020analysis,sathyapriya2014analysis,wu2020optimized}. Wavelet-based methods provide a multi-scale alternative for delineating salient peaks~\cite{banerjee2012delineation}. In PPG analysis, systolic peaks are commonly detected through local-maxima search or first-derivative analysis with adaptive thresholds~\cite{lerddararadsamee2012local,kelley2008automatic,li2010automatic,bruser2013robust,shin2008automatic}. BCG J-peak detection often requires more noise-tolerant strategies, such as template matching, autocorrelation, or motion-artifact reconstruction~\cite{alivar2019motion}. Similarly, BSG analysis frequently estimates cardiac periodicity by identifying dominant peaks in the autocorrelation function (ACF)~\cite{song2023engagement}. Although effective in controlled settings, these approaches are often brittle under changing signal morphology, sensor characteristics, or degraded signal quality, and they typically require substantial modality-specific tuning.

\textbf{Deep Learning Approaches.} To address the limitations of hand-engineered heuristics, deep learning architectures---particularly Convolutional Neural Networks (CNNs)---have been applied to peak detection in ECG, PPG, and BCG signals~\cite{schranz2024surrogate,kazemi2022robust,chen2023post}. These methods, such as RPNet, cast peak detection as a supervised prediction problem and learn salient features directly from data, often improving robustness and accuracy relative to traditional pipelines~\cite{vijayarangan2020rpnet}. However, their predictions are typically produced without explicit human-readable justification. As a result, although these models can be effective detectors, they often provide limited support for expert verification, audit, or failure analysis when signals are noisy or ambiguous.
This limitation motivates the exploration of explainable LLM-based approaches that aim to combine competitive detection performance with structured, interpretable rationales for peak selection.

\subsection{Large Language Models in Physiological Time-Series Analysis}

Large Language Models (LLMs) are increasingly applied to physiological time-series analysis, though research currently focuses on high-level semantic tasks like sleep stage captioning in OpenTSLM~\cite{langer2025opentslm} or arrhythmia diagnosis in ECG-LM~\cite{yang2025ecg}. Despite their power, these models often struggle with "numeric grounding"—the precise localization of events like a specific peak at a distinct time index—because they treat signals holistically rather than as sequences requiring fine-grained numeric inference.

This gap is largely due to the architectural limitations of auto-regressive models; as they generate tokens sequentially, cumulative errors can cause the model to "drift out of distribution" during long-form tasks~\cite{arbuzov2025beyond}. Consequently, numeric retrieval accuracy tends to degrade significantly as input length increases~\cite{fons2024evaluating}. To overcome this fundamental constraint, our approach distills physiological signals into a condensed Peak Representation, which shortens the input sequence for the LLM while preserving its essential informational content for precise temporal localization.

\subsection{Explainable AI for Physiological Signal Analysis}

Explainable AI (XAI) is increasingly vital in medicine due to ethical requirements for patient transparency and the practical need for expert confidence in automated decision-making~\cite{tjoa2020survey}. Traditional XAI research primarily focuses on post-hoc methods that seek to explain a model's decision by attributing importance to specific input features~\cite{baehrens2010explain, zeiler2014visualizing}. Techniques like Gradient SHAP (GS), for instance, explain predictions by computing the gradients of outputs with respect to points along an interpolation path from a baseline reference to the input~\cite{lundberg2017unified}. While these techniques identify influential data points, they often yield complex interpretations that fail to capture a model's underlying logic or provide a structured diagnostic rationale. To address these limitations, the Peak-Detector framework shifts from post-hoc attribution to inherent, self-explanatory design. By utilizing a custom Peak-Explanation Dataset and instruction-tuning, the model generates direct, intuitive justifications for its detections, positioning the framework as a transparent partner in expert review workflows where reasoning is as critical as numerical accuracy.

\section{METHOD}
\label{sec:system}
\begin{figure}[htbp]
    \centering
    \includegraphics[width=\textwidth]{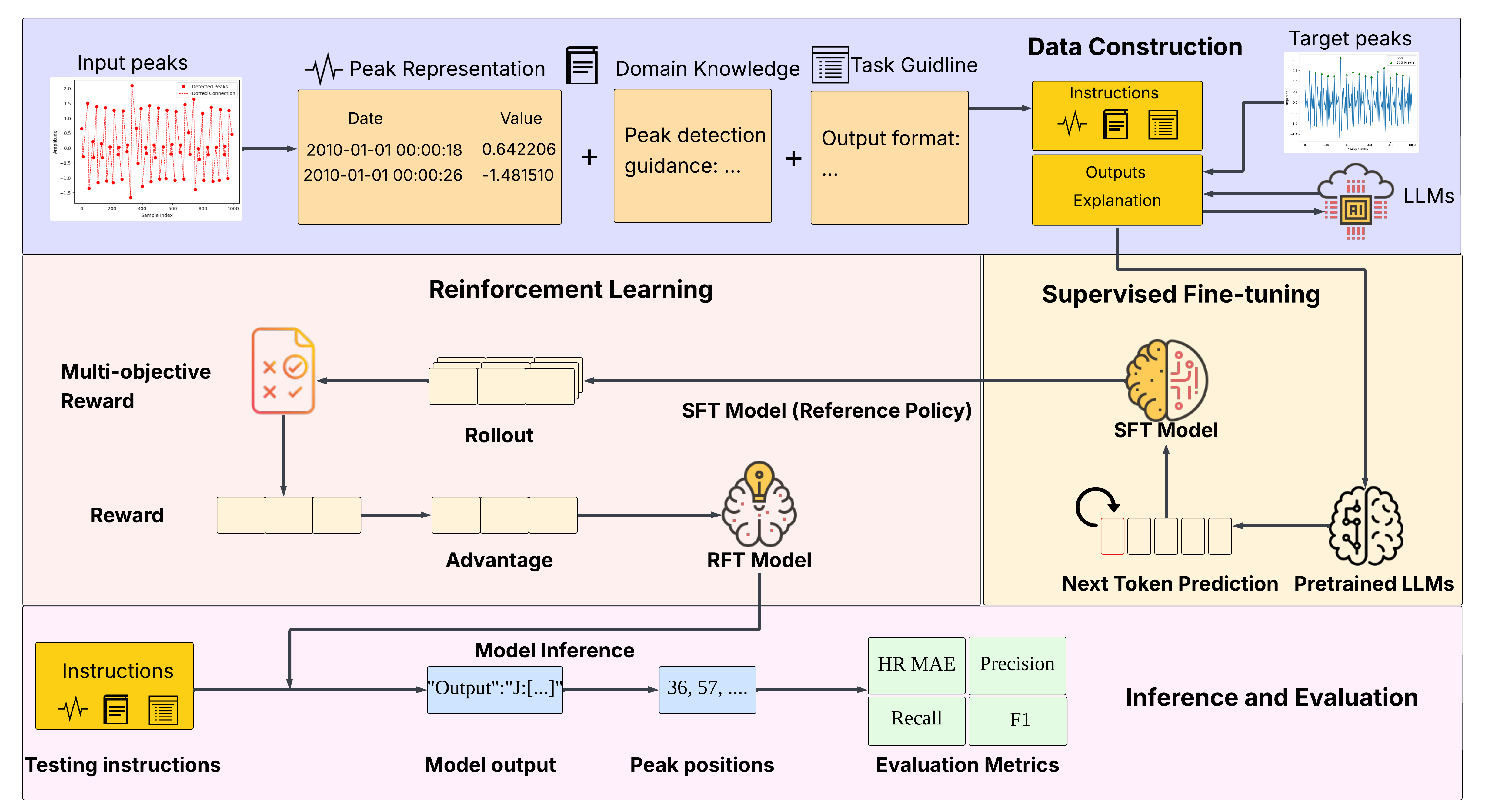}
    \Description{A diagram showing the Peak-Detector Framework.}
    \caption{Peak-Detector Framework}
    \label{fig:framework}
\end{figure}
\subsection{Peak Detector Framework}

This section details the design and implementation of Peak-Detector, our proposed framework for accurate and explainable cross-modal peak detection. As illustrated in Fig.~\ref{fig:framework}, the methodology is presented in a structured, block-by-block progression:

\begin{itemize}
    \item \textbf{Block 1: Data Construction and Representation.} We commence by introducing the foundational \textit{Peak Representation} (Section~\ref{sec:Peak Representation: From Signal to Sequence}), a technique that discretizes dense physiological time-series into a compact, text-based sequence optimized for LLM processing. Building on this, we delineate the construction of the \textit{Peak-Explanation Dataset} (Section~\ref{sec:Peak-Explanation Dataset Construction}), a novel instruction-tuning corpus synthesized via a semi-automated pipeline that pairs signal representations with human-aligned, LLM-generated explanations.
    
    \item \textbf{Block 2: Supervised Fine-Tuning (Stage 1).} As detailed in Section~\ref{sec:SFT}, the first phase of our \textit{Two-Stage Instruction Tuning Strategy} focuses on establishing a baseline capability. In this block, a base LLM undergoes supervised fine-tuning on the Peak-Explanation Dataset to internalize the specific signal syntax and explanatory format. This process yields the \textit{SFT Model}, which functions as the reference policy for the subsequent optimization stage.
    
    \item \textbf{Block 3: Reinforcement Learning (Stage 2).} Following the SFT phase, the framework transitions to the optimization block (Section~\ref{sec:RL}). Here, the SFT Model serves as the initialization for Group Relative Policy Optimization (GRPO). We optimize the model against a multi-objective reward function—encompassing detection accuracy ($F_1$) and physiological plausibility—to produce the fully refined \textit{Peak-Detector}.

    \item \textbf{Block 4: Inference and Evaluation.} The pipeline concludes with the deployment of the optimized Peak-Detector on unseen data. In this stage, the model processes raw signal representations to generate precise peak coordinates and diagnostic explanations, evaluated with standard peak-detection and cardiovascular error metrics.
\end{itemize}

\subsection{Peak Representation: From Signal to Sequence}
\label{sec:Peak Representation: From Signal to Sequence}

\begin{wrapfigure}{r}{0.38\textwidth}
    \centering
    \includegraphics[width=\linewidth]{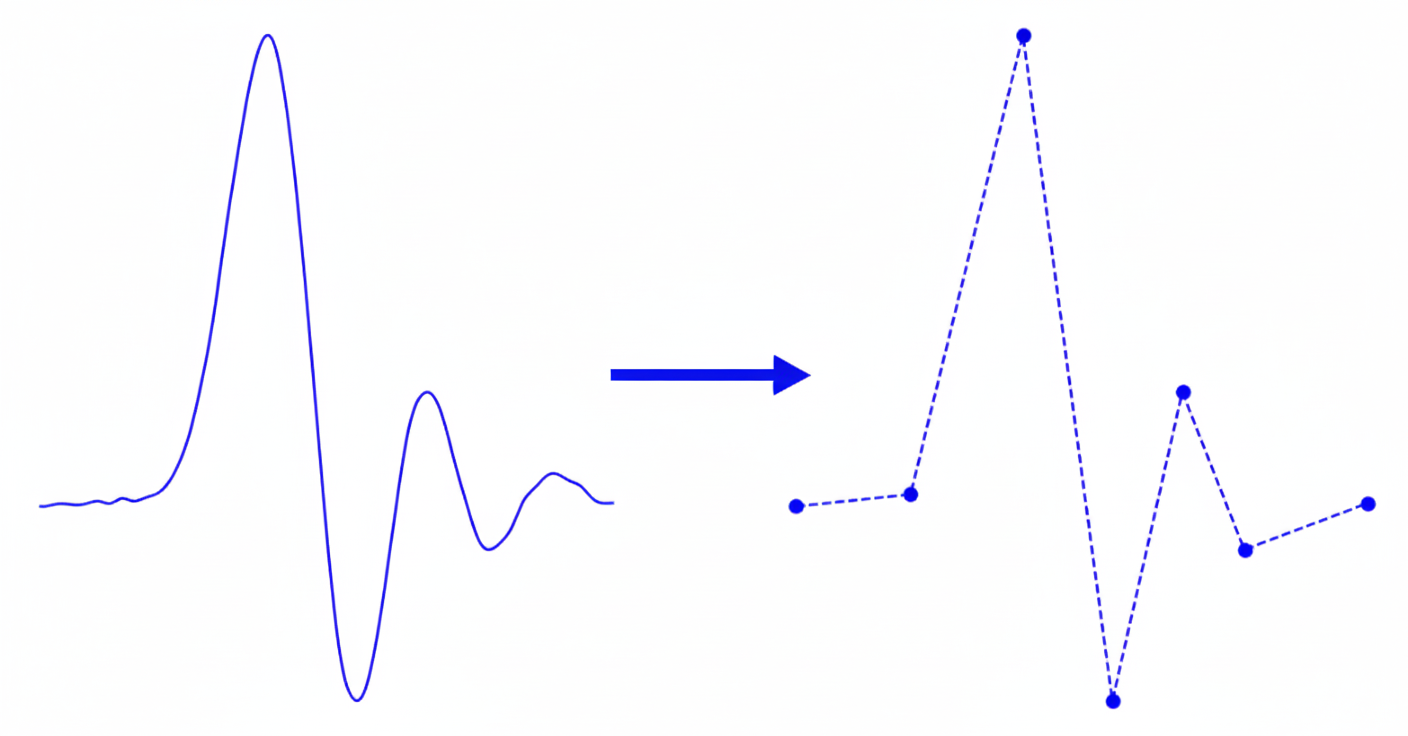}
    \caption{Signal approximation by interpolating peaks.}
    \Description{Original physiological signal approximated by interpolation between selected peaks.}
    \label{fig:approximation}
\end{wrapfigure}

Our approach assumes that the most informative content in a physiological time-series is primarily encoded around \textbf{local extrema} that delineate each cardiac cycle. Intermediate samples can be reconstructed via interpolation between these key points, as illustrated in Fig.~\ref{fig:approximation}. This observation motivates transforming a dense numeric signal into a compact, structured sequence that an LLM can reason over. The proposed \textit{Peak Representation} operationalizes this idea and serves as the input to \textit{Peak-Detector}.

\textit{Step 1: Inclusive Local-Extrema Extraction.}  
We perform a \textbf{lenient local-extrema search}~\cite{virtanen2020scipy} on each preprocessed segment $x = \{x_t\}_{t=1}^T$ to construct a comprehensive candidate set of peaks $\mathcal{C}=\{c_i\}$ by collecting a high-recall set of both maxima and minima. Each candidate $c_i$ stores its temporal index $t_i$ and normalized amplitude $a_i$. This ``inclusive-first'' strategy ensures that all potential fiducial points are preserved for downstream reasoning.

\textit{Step 2: Timestamp Transformation.}  
A key point of our representation is encoding each peak’s temporal position as a human-readable timestamp rather than a raw numeric index. Motivated by evidence that LLMs excel at calendrical and timestamp reasoning compared to long numeric sequences~\cite{fons2024evaluating}, each sample index $t_i$ is converted into a high-resolution \textbf{synthetic calendar timestamp}:
\begin{center}
$T_i(t_i) = \text{YYYY-MM-DD}~\text{HH:MM:SS}$
\end{center}

A fixed reference time $T_0=\texttt{2020-01-01 00:00:00}$ is used, and absolute indices are computed across segments to prevent window-local ambiguities. The index $t_i$ is transformed into elapsed seconds by dividing by the sampling frequency $F_s$ (i.e., $\text{calendar-second} = t_i / F_s$), then formatted into the ``HH:MM:SS'' string.  
For example, at $F_s=1$\,Hz, a peak at $t_i=97$ corresponds to 97\,s $\rightarrow$ ``00:01:37'', which is appended to the reference date to yield the final timestamp. Timestamps are reset at the beginning of each segment to maintain bounded token lengths while allowing the model to reason about relative temporal intervals.

\textit{Step 3: Serialization into a Compact Sequence.}  
Each detected peak is represented as a key–value pair and serialized into a structured textual sequence. All entries are arranged chronologically to preserve temporal coherence, forming a compact and information-dense representation well suited for LLM processing.

Overall, this design aims to:  
(1) compress long physiological waveforms into short, structure-preserving token sequences;  
(2) retain the fiducial information necessary for accurate peak detection across ECG, PPG, BCG, and BSG;  
(3) express inputs in a format that LLMs handle effectively (calendar/timestamp-like symbols rather than long raw numerics); and  
(4) remain modality-agnostic and reproducible.

\subsection{Peak-Explanation Dataset Construction}
\label{sec:Peak-Explanation Dataset Construction}

\begin{figure}[htbp]
    \centering
    \includegraphics[width=0.5\textwidth]{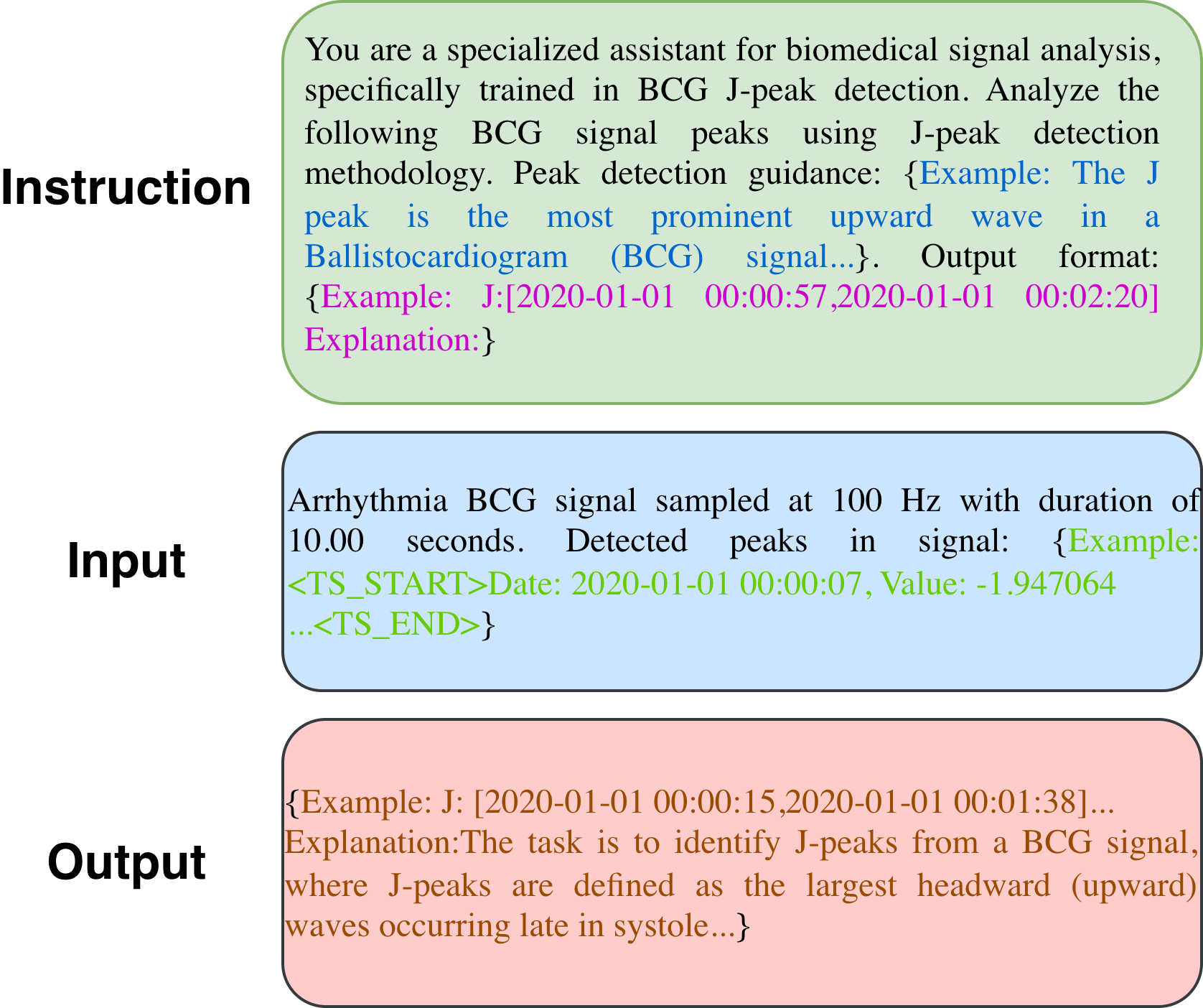}
    \caption{A representative sample from the Peak-Explanation Dataset (BCG Arrhythmia subset). The figure illustrates the instruction-tuning triplet structure, comprising the task instruction, the input peak representation, and the ground-truth output with explanatory reasoning.}
    \Description{A representative sample from the Peak-Explanation Dataset}
    \label{fig:dataset}
\end{figure}

To jointly optimize accurate peak detection and interpretable reasoning, we construct the \textbf{Peak-Explanation Dataset} as as \textit{(Instruction, Input, Output)} triplets (Fig.~\ref{fig:dataset}).  \textit{Instruction} specifies (i) the target peak type for the given modality (e.g., J-peaks in BCG/BSG, R-peaks in ECG, systolic peaks in PPG), (ii) a concise guideline describing the morphological/temporal characteristics of the target peaks, and (iii) the required output format (e.g., \texttt{\{J:[t1,t2,...] Explanation: ...\}}). \textit{Input} is a compact textual serialization of the signal segment produced by our Peak Representation, which lists candidate extrema as \texttt{(Date, Value)} pairs between \texttt{<TS\_START>} and \texttt{<TS\_END>}. \textit{Output} is a composite target: the ground-truth peak timestamps in the same representation, concatenated with an \textit{Explanation} describing why these peaks are selected and why common distractors are rejected.

\begin{figure}[htbp]
    \centering
    \includegraphics[width=0.8\textwidth]{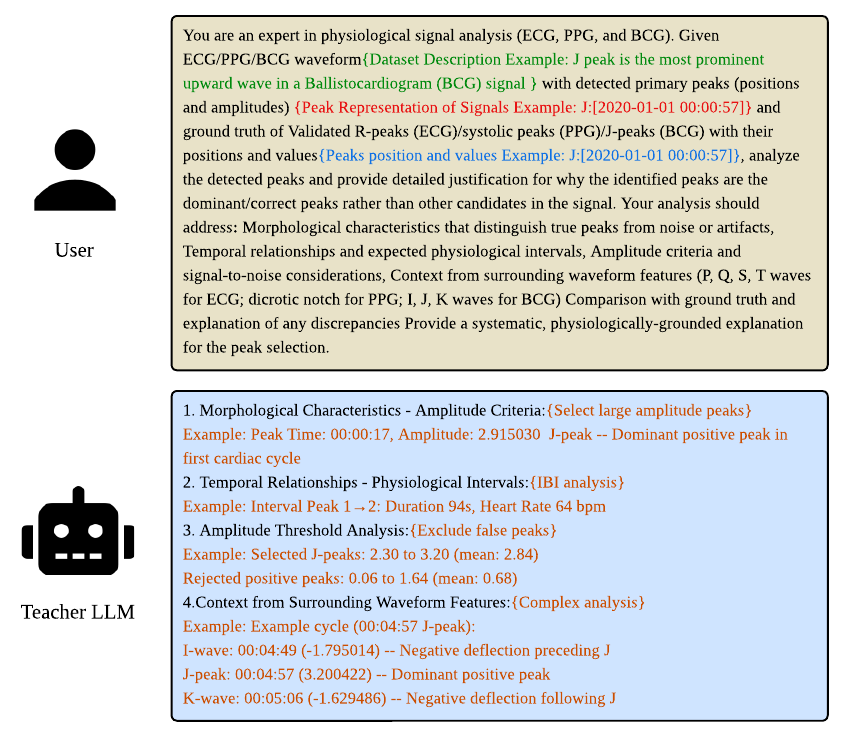}
    \caption{Teacher LLM-supervised data construction and structured response template.}
    \Description{Pipeline diagram showing teacher LLM generation of structured peak-detection explanations.}
    \label{fig:inquiry}
\end{figure}

\paragraph{Explanation generation.} We build this dataset using a semi-automated pipeline. First, raw signals and ground truth peaks are preprocessed and converted into the textual Input sequence as shown in section~\ref{sec:Peak Representation: From Signal to Sequence}. To generate the corresponding explanation in Output, we leverage a powerful "teacher" LLM (e.g., GPT-4o). As illustrated in Figure~\ref{fig:inquiry}, we prompt explicitly specifies the \emph{signal modality/dataset} (e.g., ECG/PPG/BCG) and the corresponding \emph{target-peak definition} (e.g., ECG R-peaks, PPG systolic peaks, BCG J-peaks), together with (1) the candidate peak list produced by our Peak Representation and (2) the ground-truth peaks (converted into the same timestamp/value format for consistency). Conditioned on these modality-specific instructions, the teacher LLM outputs a \emph{structured rational} to justify why the provided ground-truth target peaks correspond to physiologically meaningful peaks among the candidate extrema, and to explain why other candidates are rejected via: (1) morphology/amplitude evidence consistent with the modality, (2) temporal/IBI consistency, (3) amplitude threshold-based rejection of spurious candidates, and (4) surrounding-wave context (e.g., P/QRS/T for ECG, systolic upstroke and dicrotic notch for PPG, and I/J/K complexes for BCG.

\begin{figure}[ht]
    \centering
    \begin{subfigure}{0.7\textwidth}
        \centering
        \includegraphics[width=\textwidth]{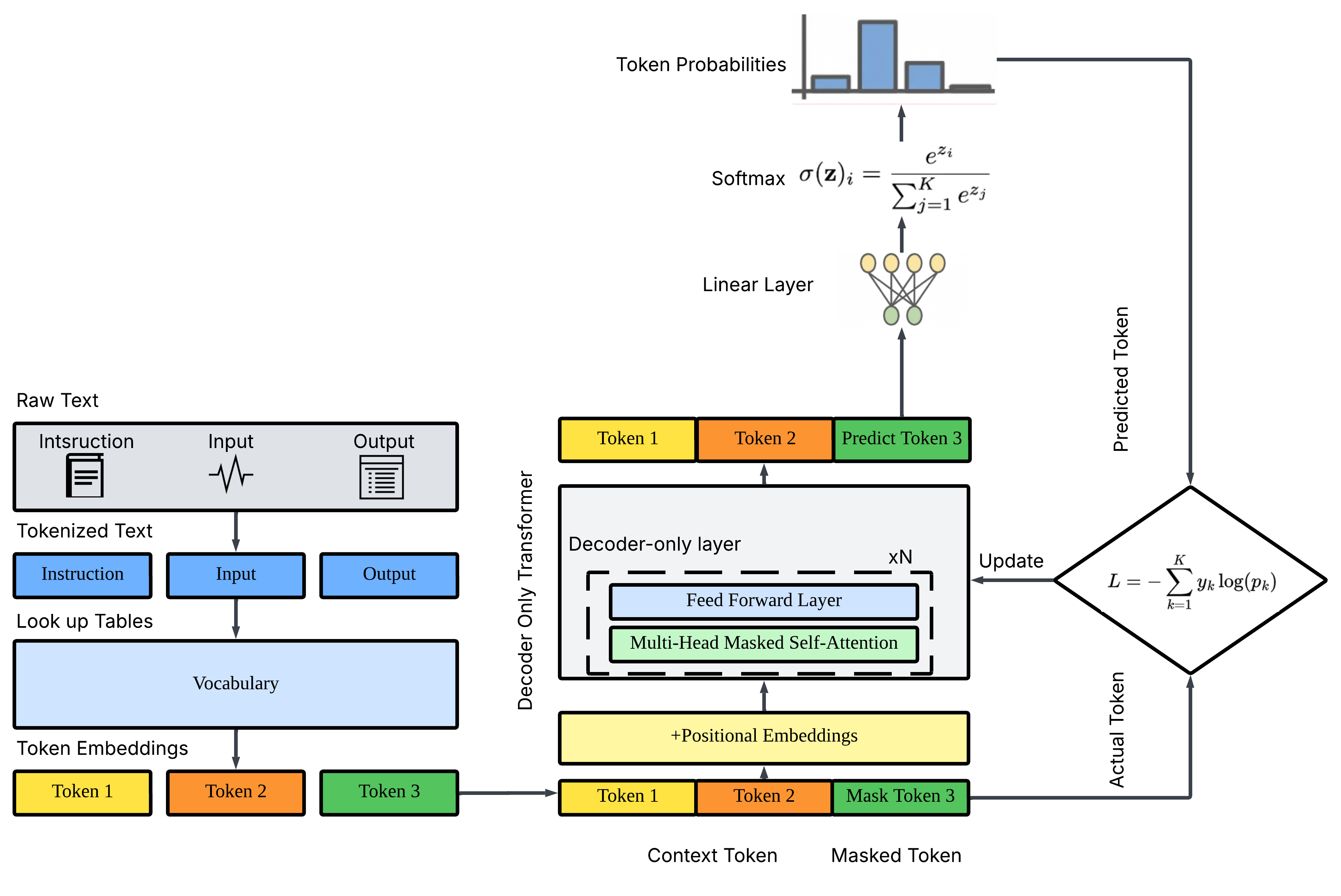}
        \caption{SFT illustration}
        \label{fig:sft}
    \end{subfigure}
    
    \begin{subfigure}{0.6\textwidth}
        \centering
        \includegraphics[width=\textwidth]{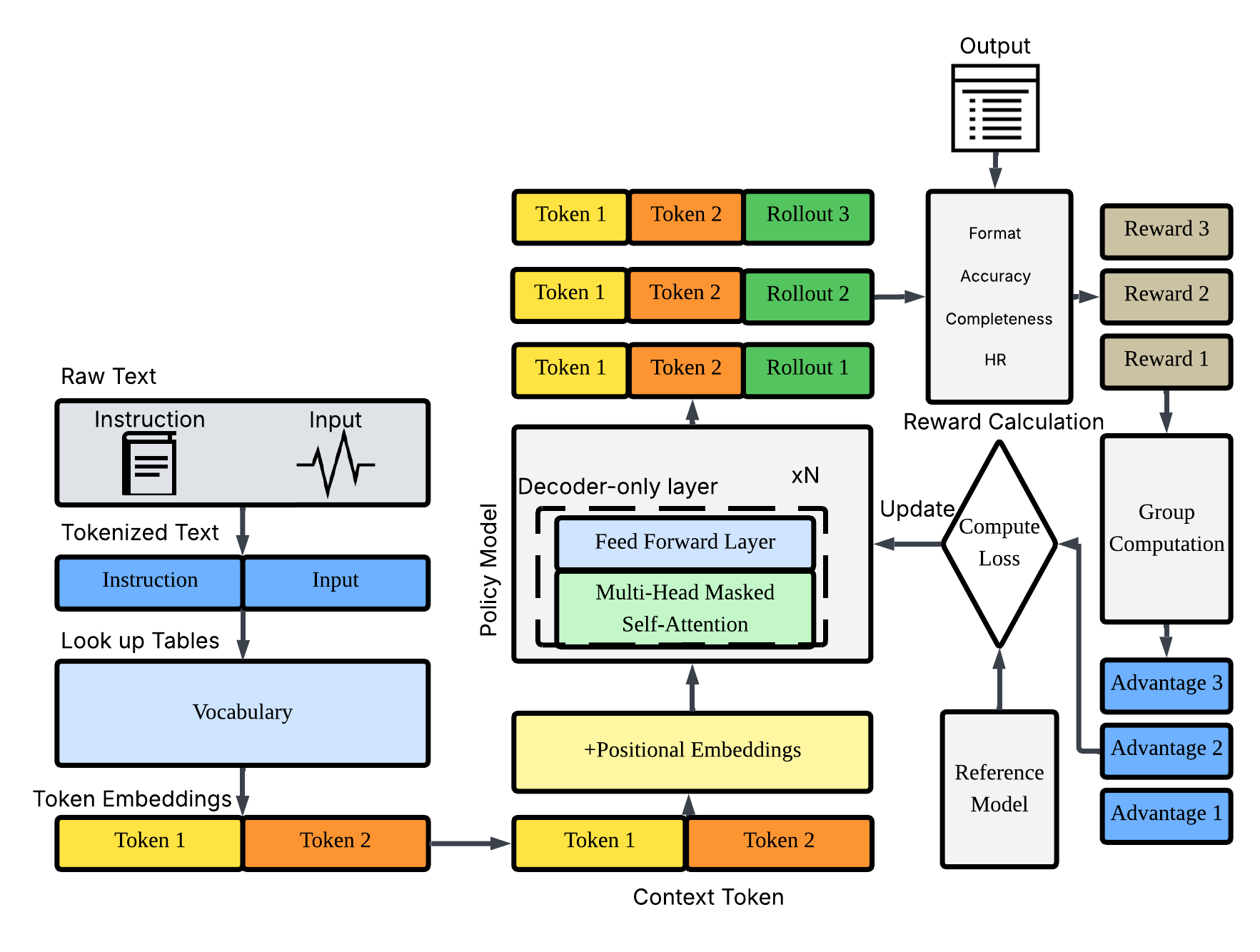}
        \Description{A brief description of the RL.png image.}
        \caption{GRPO illustration}
        \label{fig:grpo}
    \end{subfigure}
    \caption{Training strategies comparison}
    \Description{Comparison of supervised fine-tuning and GRPO reinforcement learning training workflows.}
    \label{fig:training-comparison}
\end{figure}

\subsection{Two-Stage Instruction Tuning Strategy}
\label{sec:tuning_strategy}

To effectively train Peak-Detector for both robust performance and explainability across diverse physiological signals, we employ a sophisticated two-stage instruction tuning strategy. This approach synergistically combines Supervised Fine-Tuning (SFT) with Reinforcement Learning (RL) optimization, leveraging the strengths of each paradigm. We utilize a pre-trained open-source LLM (specifically \texttt{Qwen2.5} for our experiments) as the base model, fine-tuning it with our meticulously constructed Peak-Explanation Dataset.

\subsubsection{Stage 1: Supervised Fine-Tuning (SFT)}
\label{sec:SFT}

This stage focuses on adapting the general-purpose base model to the specific syntax of physiological signals. As illustrated in the SFT schematic (Fig.~\ref{fig:sft}), the process operates through the following pipeline:

\textbf{Input Processing and Embedding.} The workflow begins with the \textit{Raw Text} inputs from Peak-Explanation Dataset—comprising the Instruction, Signal Input, and Ground Truth Output. These texts are tokenized and mapped to dense vectors via \textit{Look-up Tables} that access the model's Vocabulary. Crucially, \textit{Positional Embeddings} are added to these token embeddings to preserve the temporal order of the signal sequence, ensuring the model retains time-series awareness.

\textbf{Decoder-Only Architecture.} The combined embeddings are processed through a stack of $N$ \textit{Decoder-only layers}. Within each layer, \textit{Multi-Head Masked Self-Attention} allows the model to attend to relevant historical context (signal antecedents) while preventing information leakage from future tokens. This is followed by \textit{Feed Forward Layers} that process the attended features to capture non-linear morphological patterns.

\textbf{Next-Token Prediction.} The high-dimensional hidden states from the final transformer layer are projected into the vocabulary space via a \textit{Linear Layer}. A \textit{Softmax} function is then applied to generate a probability distribution over the vocabulary, determining the likelihood of the \textit{Predicted Token}.

\textbf{Optimization Objective.} The training objective is to minimize the discrepancy between the predicted distribution and the \textit{Actual Token}. We employ a standard cross-entropy loss function:
\begin{equation}
\label{eq:cross-entropy}
L = - \sum_{k=1}^{K} y_k \log(p_k)
\end{equation}
where $K$ is the vocabulary size, $y_k$ is the binary indicator for the ground-truth token, and $p_k$ is the predicted probability. The calculated loss drives the \textit{Update} step, adjusting the model parameters via backpropagation to maximize the likelihood of correct peak detections and valid explanations.

This SFT stage serves three primary objectives: (1) it stabilizes the LLM's learning dynamics for numerical data; (2) it enforces valid output formatting; and (3) it establishes a foundational capability for peak localization before the subsequent Reinforcement Learning stage.

\subsubsection{Stage 2: Reinforcement Learning Optimization.}
\label{sec:RL}
Following Supervised Fine-Tuning (SFT), the model undergoes a second stage of optimization using Reinforcement Learning (RL) with Group Relative Policy Optimization (GRPO)~\cite{shao2024deepseekmath}. As illustrated in the RL schematic (Fig.~\ref{fig:grpo}), this process refines the model's policy at the \textit{sequence level} through a structured, block-by-block progression:

\noindent \textbf{Rollout and Reference.}
The workflow initiates with the \textit{Policy Model} (initialized from the SFT model) sampling a group of $G=16$ distinct outputs, or ``Rollouts,'' for a single input instruction. Simultaneously, a frozen \textit{Reference Model} processes the same input. This parallel execution is critical for computing the KL-divergence constraint, which acts as a regularization anchor to prevent the policy from drifting too far from the initial linguistic distribution established during SFT.

\noindent \textbf{Reward Evaluation.}
Each generated rollout is evaluated against a multi-objective \textit{Reward Function} ($R_{\text{total}}$). Unlike standard RL which often uses a single scalar, our framework computes a composite score derived from four distinct criteria: Format Compliance, Detection Accuracy, Count Completeness, and Heart-Rate Consistency. This specific design ensures the model optimizes for both syntax validity and physiological fidelity.

\noindent \textbf{Group Advantage Estimation}
A core innovation of GRPO is the elimination of the separate ``Critic'' value model typically required in algorithms like PPO. Training a Critic for long-context physiological sequences is computationally expensive and prone to high variance. Instead, GRPO uses the \textit{group average} as a dynamic baseline.
The intuition is straightforward: for a given input query $q$, we compare the reward of a specific rollout $o_i$ against the average reward of its peers in the same batch. If $o_i$ performs better than the average, it yields a positive advantage ($A_{i,t} > 0$), signaling the model to increase the probability of that sequence.

The optimization objective maximizes this expected advantage while enforcing strict stability constraints:
\begin{equation}
\begin{split}
\mathcal{J}_{GRPO}(\theta) = & \mathbb{E}\left[q \sim P(Q), \{o_i\}_{i=1}^G \sim \pi_{\theta_{old}}(O|q)\right] \\
& \frac{1}{G}\sum_{i=1}^{G} \frac{1}{|o_i|} \sum_{t=1}^{|o_i|} \left\{ \underbrace{\min\left[ r_t(\theta) \hat{A}_{i,t}, \text{clip}\left(r_t(\theta), 1-\varepsilon, 1+\varepsilon\right) \hat{A}_{i,t}\right]}_{\text{Trust Region Constraint}} - \underbrace{\beta_{KL} \mathbb{D}_{KL}\left[\pi_\theta || \pi_{ref}\right]}_{\text{Language Regularization}} \right\}
\end{split}
\label{eq:grpo}
\end{equation}
Equation~\ref{eq:grpo} incorporates three critical mechanisms designed to balance the \textit{stability-plasticity dilemma}:
\begin{enumerate}
    \item \textbf{Probability Ratio $r_t(\theta)$:} Measures how much more likely an action is under the new policy compared to the old, tracking the magnitude of the update.
    \item \textbf{Trust Region Clipping ($\varepsilon=0.2$):} This prevents the model from making distinctively large updates based on a single batch. We selected $\varepsilon=0.2$ as a conservative bound to avoid ``catastrophic forgetting,'' where a single noisy batch could otherwise destroy the model's learned representations.
    \item \textbf{KL-Penalty ($\beta_{KL}=0.04$):} Physiological signal processing requires the model to strictly adhere to the syntax learned during SFT. The term $\beta_{KL}$ acts as a ``syntax anchor.'' We empirically set $\beta_{KL}=0.04$, which is sufficiently high to preserve linguistic coherence but low enough to allow the model to shift its \textit{reasoning} logic toward better peak detection.
\end{enumerate}

\noindent \textbf{Update} The model uses these advantages to Compute Loss and perform a policy Update, increasing the probability of rollouts that outperform their peers while ensuring the policy does not drift too far from the original SFT distribution.

\noindent \textbf{Multi-Objective Reward Design}
Defining ``success'' in physiological peak detection is multifaceted. To guide the optimization through this complex landscape, we derive a composite reward function $R_{\text{total}}$ structured as a \textit{hierarchy of clinical needs}:
\begin{equation}
    R_{\text{total}} = \alpha \cdot R_{\text{format}} + \beta \cdot R_{\text{detection}} + \gamma \cdot R_{\text{complete}} + \delta \cdot R_{\text{HR}}
\label{eq:reward}
\end{equation}
The coefficients are set to $\alpha=0.1$, $\beta=0.6$, $\gamma=0.15$, and $\delta=0.15$. These values enforce a specific learning priority: syntax is a baseline constraint ($\alpha$), detection accuracy is the primary clinical objective ($\beta$), while count and heart-rate metrics ($\gamma, \delta$) serve as physiological regularizers.

The individual components are defined as follows:
\begin{itemize}
    \item \textbf{Format Compliance ($R_{\text{format}}$):} This is a binary reward where $R_{\text{format}}=1$ if the output adheres to the specified syntax (Fig.~\ref{fig:dataset}) and $0$ otherwise. We assign a low weight ($\alpha=0.1$) because this acts as a gating mechanism. Once the model learns the syntax (typically early in training), this reward saturates. A higher weight would risk the model optimizing for valid but empty structures.

    \item \textbf{Detection Accuracy ($R_{\text{detection}}$):} As the primary objective, this is quantified by the $F_1$-score between predicted and ground-truth peaks using a strict 30\,ms tolerance window. The $F_1$-score is chosen over simple accuracy to robustly handle class imbalance (sparse peaks vs. abundant background signal).

    \item \textbf{Count Completeness ($R_{\text{complete}}$):} To penalize missed or hallucinated peaks, we employ an exponential decay function based on the total peak count difference:
    \begin{equation}
        R_{\text{complete}} = \exp\left(-|N_{\text{pred}} - N_{\text{gt}}|\right)
    \end{equation}
    We selected an exponential form rather than a linear penalty (e.g., Mean Absolute Error) to strictly bound the reward to $(0,1]$. This design prevents gradient explosions in early training stages where the model might predict wildly incorrect counts (e.g., 0 or 100 peaks), which would otherwise destabilize the policy update.

    \item \textbf{Heart-Rate Consistency ($R_{\text{HR}}$):} Physiological plausibility is assessed via the normalized error in derived heart rate:
    \begin{equation}
        R_{\text{HR}} = \exp\left(-2 \cdot \frac{|HR_{\text{pred}} - HR_{\text{gt}}|}{HR_{\text{gt}}}\right)
    \end{equation}
    The scaling factor of $-2$ was empirically chosen to sharpen the reward curve, ensuring that only high-precision estimates (within $\approx 5\%$ error) yield significant positive reinforcement. This forces the model to refine its peak localization to meet clinical-grade heart rate standards.
\end{itemize}

\section{EXPERIMENT SETUP AND EVALUATION}
\label{sec:experiment}

\subsection{Datasets and Preprocessing}

\textbf{Datasets}.
To thoroughly evaluate the cross-modal and explanatory capabilities of Peak-Detector, we conducted experiments on a comprehensive suite of seven publicly available physiological signal datasets. In this experiment, we focus on detecting the prominent peak, such as R-peak for ECG, systolic peak for PPG, and J-peak for BCG/BSG. The ECG datasets used were the MIT-BIH Arrhythmia Database~\cite{moody1992bih} and the Incart Arrhythmia Database~\citep{tihonenko2007st}. For PPG analysis, we utilized the BIDMC Database~\cite{pimentel2016toward} and the Capnobase Database~\cite{karlen2013multiparameter}. Then, the BCG experiments were performed on the Kansas Database~\cite{carlson2020bed}, the BCG Arrhythmia Database~\cite{zhan2025multi}. Finally, for BSG experiments, we self-collected ICU database.
Detailed descriptions of each dataset are provided in Appendix~\ref{appendix:Dataset details}.

\textbf{Signal Preprocessing}.
Prior to analysis, all physiological signals undergo a standardized preprocessing pipeline to ensure consistency and mitigate noise. Each continuous recording is first segmented into fixed-length windows of 1000 samples. A fourth-order Butterworth bandpass filter is then applied, with a passband frequency ranging from 0.6\,Hz to 15\,Hz. This step effectively removes baseline wander and high-frequency noise while preserving the characteristic morphology of cardiac events. Following filtration, a z-score normalization is performed on each segment, transforming the signal to have a mean of zero and a standard deviation of one. This standardization mitigates amplitude variations across different subjects and recording sessions, thereby enhancing the robustness of the peak detection algorithms.

\textbf{Subject-Independent Cross-Validation.}
To assess the robustness and generalizability of our model, we implemented a 4-fold subject-independent cross-validation scheme. For each dataset, subjects were randomly partitioned into four disjoint subsets of approximately equal size. The experimental procedure was conducted iteratively: in each iteration, three folds constituted the training set, while the remaining fold served as the unseen testing set for the learning-based models. Signal-processing baselines, which do not require training, were evaluated directly on the corresponding test folds to ensure comparable evaluation conditions. To demonstrate model stability, final performance metrics are reported as the mean $\pm$ standard deviation across the four folds. Furthermore, to ascertain statistical significance, we performed a two-tailed Welch’s $t$-test~\cite{west2021best} comparing the proposed Peak-Detector against the best-performing baseline model (defined by the lowest MAE).

\subsection{Evaluation Metrics}
\label{sec:Evaluation Metrics}

To rigorously assess the performance of Peak-Detector, we employ a comprehensive set of evaluation metrics that capture both the absolute accuracy of peak localization and the physiological consistency of the derived cardiac rhythm. The primary metric for evaluating accuracy is the F1-score, Precision, and Recall. Following established conventions in physiological signal processing~\cite{reiss2019deep,zuo2025tau}, a predicted peak is considered a True Positive (TP) if it falls within a specified tolerance radius of $\pm 30$\,ms around a ground-truth peak, which is smaller than common used threshold~\cite{zuo2025tau,reiss2019deep,elgendi2013systolic}. This strict window ensures that only highly precise detections contribute positively to the F1-score. In addition to this, we also evaluate the model's ability to accurately derive key parameters: heart rate (HR) and heart rate variability (HRV). These are calculated from the sequence of detected peak-to-peak intervals (e.g., R-R, systolic-systolic, J-J), and We report the Mean Absolute Error (MAE) for both HR and HRV compared to ground-truth values in beats per minute (bpm) and milliseconds (ms), respectively. To mitigate the bias of fixed temporal constraints, we implemented an adaptive tolerance mechanism (Appendix~\ref{sec:relative_tolerance}) using a window of $\pm 5\%$ of the local Inter-Beat Interval (IBI). 

\subsection{Baselines}
\label{sec:Baselines}

To comprehensively benchmark the performance of Peak-Detector, we compared our model against a diverse set of nine state-of-the-art baselines. This selection encompasses both modality-specific signal-processing approaches and generalizable deep learning models. Detailed descriptions of each methodology are provided in Appendix~\ref{appendix:Baseline details}. Comprehensive details regarding the experimental setup for Peak-Detector are documented in Appendix~\ref{app:experimental_protocol}. All baseline methods were evaluated using identical preprocessing pipelines and matching tolerance criteria ($\pm 30$~ms fixed window) to ensure fair comparison.

The signal-processing baselines leverage domain-specific heuristics and include the \textbf{Pan-Tompkins} algorithm~\cite{fariha2020analysis} and \textbf{Nabian}'s method~\cite{nabian2018open} for ECG; \textbf{Elgendi}'s algorithm~\cite{elgendi2012analysis} and the multi-scale approach by \textbf{Bishop}~\cite{bishop2018multi} for PPG; and \textbf{Pino}'s method~\cite{pino2017bcg} alongside the segmentation-based algorithm by \textbf{Choi}~\cite{choi2009slow} for BCG and BSG. For these methods, the algorithms were applied directly to the preprocessed signals, and the detected peaks were compared against the ground truth to compute evaluation metrics.

The deep learning baselines represent advanced data-driven approaches, including \textbf{CNN-SWT}~\cite{yun2022robust} for ECG, \textbf{1D-UNet++}~\cite{zhou20211d} for BCG, and \textbf{FR-Net}~\cite{chen2023sample}, a specialized CNN-Transformer hybrid network developed for fetal R-peak detection. In contrast to the heuristic methods, these deep learning models output a probability distribution of peak locations rather than discrete coordinates. Consequently, we employed a local extrema search~\cite{virtanen2020scipy} to extract the final peak positions from the generated probability maps.

\subsection{Results}
\label{sec:Results}

\begin{table}[t]
\centering

\caption{Peak Detection Performance Comparison across Signal Modalities and Baselines. Lower values are better for MAE and MAPE (HR(bpm), HRV(ms)), higher values are better for F1, Pre, and Rec. Best performance in each metric (excluding Pre/Rec) is \textbf{bold}, second best is \underline{underlined}. Standard deviations are shown in a smaller font. Superscripts on Peak-Detector indicate significance against the second-best method: \textsuperscript{*} $p<0.05$, \textsuperscript{**} $p<0.01$. The $p$-value column is retained for reference.}
\label{tab:results_combined}
\resizebox{\textwidth}{!}{%
\begin{tabular}{cc|l|ccccccccccc} 
\toprule
& & \textbf{Metric} & \textbf{Pan-T} & \textbf{Nabian} & \textbf{Elg} & \textbf{Bishop} & \textbf{Pino} & \textbf{Choi} & \textbf{CNN-SWT} & \textbf{1D-UNet++} & \textbf{FR-Net} & \textbf{Peak-Detector} & \textbf{$p$-value} \\ 
\midrule
\multirow{12}{*}{\rotatebox{90}{\scriptsize\textbf{ECG}}} & \multirow{6}{*}{\rotatebox{90}{\scriptsize\textbf{MIT-BIH}}}
& HR MAE(bpm) & 1.06 &4.23&3.81&79.75&3.59&7.01& \val{1.19}{0.16} & \underline{\val{0.57}{0.07}} & \textbf{\val{0.48}{0.01}} & \valsig{0.86}{0.08}{**} & $2.23e-03$ \\
& & HRV MAE(ms) & 8.48 &27.13&39.10&68.57&26.00&62.94& \val{18.50}{0.57} & \underline{\val{6.50}{0.68}} & \textbf{\val{5.02}{0.43}} & \valsig{9.97}{0.17}{**} & $3.33e-05$ \\
& & HR MAPE (\%) & 1.48 &5.44&5.63&105.98&4.45&9.61& \val{1.48}{0.16} & \underline{\val{0.79}{0.07}} & \textbf{\val{0.59}{0.01}} & \valsig{1.24}{0.15}{**} & $3.14e-03$ \\
& & HRV MAPE (\%) & 33.09 &43.66&157.56&268.89&78.28&240.54& \val{75.08}{6.93} & \underline{\val{27.84}{4.71}} & \textbf{\val{15.10}{0.75}} & \valsig{61.10}{0.91}{**} & $5.68e-10$ \\
&   & F1 & 0.975 &0.775&0.107&0.541&0.886&0.717& \underline{\val{0.988}{0.001}} & \val{0.980}{0.001} & \textbf{\val{0.991}{0.001}} & \valsig{0.973}{0.002}{**} & $1.40e-04$ \\
&   & Pre & 0.980 &0.934&0.109&0.424&0.911&0.744& \val{0.990}{0.001} & \val{0.981}{0.001} & \val{0.994}{0.001} & \valsig{0.971}{0.003}{**} & $2.03e-04$ \\
&   & Rec & 0.970 &0.662&0.104&0.748&0.863&0.691& \val{0.985}{0.002} & \val{0.980}{0.001} & \val{0.989}{0.001} & \valsig{0.975}{0.002}{**} & $3.32e-04$ \\
\cmidrule(l){2-14} 
& \multirow{6}{*}{\rotatebox{90}{\scriptsize\textbf{Incart}}}
& HR MAE(bpm) & 3.21 &4.88&2.72&51.42&4.40&10.35& \val{0.59}{0.04} & \textbf{\val{0.34}{0.06}} & \val{0.46}{0.11} & \underline{\val{0.42}{0.04}} & $7.49e-02$ \\
& & HRV MAE(ms) & 45.22 &47.70&40.00&73.26&46.16&87.01& \val{8.82}{0.35} & \textbf{\val{5.70}{0.59}} & \underline{\val{5.90}{1.18}} & \valsig{18.79}{0.58}{**} & $6.64e-08$ \\
& & HR MAPE (\%) & 4.54 &5.69&3.79&66.97&5.04&12.07& \val{0.72}{0.05} & \textbf{\val{0.42}{0.08}} & \underline{\val{0.50}{0.09}} & \valsig{1.05}{0.08}{**} & $3.12e-05$ \\
& & HRV MAPE (\%) & 250.36 &149.26&189.22&263.30&201.51&395.78& \val{39.97}{1.80} & \textbf{\val{22.40}{2.19}} & \underline{\val{25.76}{8.49}} & \valsig{72.07}{6.41}{**} & $2.09e-04$ \\
& & F1 & 0.915 &0.778&0.164&0.482&0.840&0.588& \underline{\val{0.993}{0.001}} & \val{0.992}{0.001} & \textbf{\val{0.993}{0.001}} & \valsig{0.960}{0.001}{**} & $6.13e-08$ \\
& & Pre & 0.919 &0.893&0.167&0.413&0.867&0.639& \val{0.992}{0.001} & \val{0.993}{0.001} & \val{0.996}{0.001} & \valsig{0.960}{0.002}{**} & $9.94e-08$ \\
& & Rec & 0.910 &0.690&0.161&0.580&0.814&0.543& \val{0.994}{0.001} & \val{0.992}{0.000} & \val{0.990}{0.002} & \valsig{0.960}{0.001}{**} & $6.27e-07$ \\
\midrule
\multirow{12}{*}{\rotatebox{90}{\scriptsize\textbf{PPG}}} & \multirow{6}{*}{\rotatebox{90}{\scriptsize\textbf{BIDMC}}}
& HR MAE(bpm) & 8.92 &3.08&1.48&3.06&3.21&5.29& \underline{\val{0.71}{0.08}} & \val{0.68}{0.45} & \val{1.68}{0.54} & \textbf{\valsig{0.35}{0.13}{**}} & $5.29e-03$ \\
& & HRV MAE(ms) & 172.34 &28.05&14.72&17.11&29.67&42.07& \underline{\val{9.28}{0.76}} & \val{10.51}{6.45} & \val{27.71}{10.86} & \textbf{\valsig{5.57}{2.34}{*}} & $4.46e-02$ \\
& & HR MAPE (\%) & 10.26 &3.28&1.57&3.46&3.38&4.98& \val{0.80}{0.08} & \underline{\val{0.75}{0.50}} & \val{1.84}{0.64} & \textbf{\valsig{0.40}{0.16}{**}} & $8.82e-03$ \\
& & HRV MAPE (\%) & 1118.19 &61.21&34.57&31.67&94.50&230.03& \textbf{\val{27.51}{2.59}} & \val{34.44}{8.28} & \val{73.19}{41.87} & \val{45.39}{59.81} & $5.92e-01$ \\
& & F1 & 0.432 &0.923&0.952&0.929&0.930&0.922& \underline{\val{0.986}{0.006}} & \val{0.983}{0.009} & \val{0.969}{0.010} & \textbf{\val{0.991}{0.005}} & $2.22e-01$ \\
& & Pre & 0.438 &0.991&0.986&0.976&0.960&0.966& \val{0.985}{0.005} & \val{0.984}{0.007} & \val{0.985}{0.007} & \val{0.988}{0.005} & $3.46e-01$ \\
& & Rec & 0.427 &0.863&0.921&0.886&0.903&0.882& \val{0.987}{0.006} & \val{0.982}{0.011} & \val{0.954}{0.012} & \val{0.993}{0.004} & $1.66e-01$ \\
\cmidrule(l){2-14} 
& \multirow{6}{*}{\rotatebox{90}{\scriptsize\textbf{Capno}}}
& HR MAE(bpm) & 4.48 &2.99&0.69&1.12&8.52&6.26& \val{0.72}{0.12} & \underline{\val{0.41}{0.19}} & \val{0.47}{0.16} & \textbf{\val{0.35}{0.28}} & $7.37e-01$ \\
& & HRV MAE(ms) & 67.80 &15.17&8.60&9.90&111.56&39.71& \val{11.80}{2.10} & \val{7.14}{2.33} & \underline{\val{6.73}{1.26}} & \textbf{\val{6.19}{4.11}} & $7.05e-01$ \\
& & HR MAPE (\%) & 6.41 &2.57&0.83&1.45&8.32&5.54& \val{0.86}{0.14} & \underline{\val{0.53}{0.21}} & \val{0.54}{0.19} & \textbf{\val{0.44}{0.33}} & $6.64e-01$ \\
& & HRV MAPE (\%) & 493.32 &96.87&47.75&59.04&905.87&304.46& \val{85.38}{15.52} & \underline{\val{37.57}{9.20}} & \val{41.96}{7.04} & \textbf{\val{34.58}{16.14}} & $7.61e-01$ \\
& & F1 & 0.467 &0.838&0.928&0.876&0.554&0.902& \val{0.978}{0.019} & \val{0.990}{0.005} & \underline{\val{0.992}{0.001}} & \textbf{\val{0.992}{0.005}} & $5.89e-01$ \\
& & Pre & 0.465 &0.996&0.990&0.987&0.587&0.977& \val{0.980}{0.019} & \val{0.987}{0.007} & \val{0.996}{0.001} & \val{0.993}{0.007} & $2.89e-01$ \\
& & Rec & 0.468 &0.724&0.874&0.788&0.527&0.842& \val{0.976}{0.019} & \val{0.993}{0.005} & \val{0.987}{0.002} & \val{0.991}{0.004} & $6.90e-01$ \\
\midrule
\multirow{12}{*}{\rotatebox{90}{\scriptsize\textbf{BCG}}} & \multirow{6}{*}{\rotatebox{90}{\scriptsize\textbf{Kansas}}}
& HR MAE(bpm)& 30.57 &4.90&21.33&142.28&2.91&4.77& \underline{\val{2.74}{0.44}} & \val{2.71}{1.67} & \val{3.81}{0.56} & \textbf{\val{2.39}{1.92}} & $8.10e-01$ \\
& & HRV MAE(ms) & 310.85 &60.10&110.03&92.73&65.59&77.69& \underline{\val{51.95}{8.97}} & \val{55.31}{34.69} & \val{71.33}{15.50} & \textbf{\val{44.38}{30.27}} & $6.52e-01$ \\
& & HR MAPE (\%) & 50.51 &7.81&34.80&223.55&4.34&7.52& \val{4.06}{0.66} & \val{4.42}{2.77} & \val{5.39}{0.77} & \textbf{\val{3.94}{2.51}} & $8.06e-01$ \\
& & HRV MAPE (\%) & 481.80 &61.21&230.22&86.49&68.92&125.44& \val{31.97}{2.51} & \underline{\val{30.55}{17.19}} & \val{45.56}{7.61} & \textbf{\val{25.80}{21.45}} & $7.42e-01$ \\
& & F1 & 0.218 &0.926&0.732&0.445&0.867&0.867& \val{0.939}{0.013} & \underline{\val{0.948}{0.034}} & \val{0.940}{0.011} & \textbf{\val{0.957}{0.033}} & $6.99e-01$ \\
& & Pre & 0.177 &0.892&0.620&0.289&0.814&0.825& \val{0.936}{0.011} & \val{0.935}{0.038} & \val{0.961}{0.008} & \val{0.956}{0.026} & $4.08e-01$ \\
& & Rec & 0.285 &0.964&0.895&0.982&0.927&0.913& \val{0.942}{0.015} & \val{0.961}{0.037} & \val{0.920}{0.016} & \val{0.959}{0.041} & $9.37e-01$ \\
\cmidrule(l){2-14} 
& \multirow{6}{*}{\rotatebox{90}{\scriptsize\textbf{Arrhy}}}
& HR MAE(bpm) & 27.17 &3.97&19.42&148.79&2.71&3.76& \val{5.77}{0.81} & \underline{\val{3.34}{2.30}} & \val{5.72}{1.51} & \textbf{\val{1.43}{0.95}} & $6.85e-02$ \\
& & HRV MAE(ms) & 338.98 &57.44&128.91&67.36&75.96&68.12& \val{92.44}{6.01} & \val{84.52}{32.98} & \underline{\val{130.43}{36.23}} & \textbf{\valsig{36.85}{22.71}{*}} & $3.23e-02$ \\
& & HR MAPE (\%) & 41.00 &6.40&30.93&233.37&3.94&5.88& \val{8.97}{1.32} & \underline{\val{4.89}{2.73}} & \val{7.77}{2.07} & \textbf{\val{2.14}{1.43}} & $8.09e-02$ \\
& & HRV MAPE (\%) & 1742.69 &209.44&696.06&133.11&330.90&334.95& \val{455.37}{95.15} & \underline{\val{193.74}{51.64}} & \val{524.88}{200.78} & \textbf{\valsig{98.05}{64.46}{**}} & $2.33e-03$ \\
& & F1 & 0.583 &0.899&0.801&0.466&0.914&0.897& \val{0.934}{0.005} & \val{0.930}{0.029} & \underline{\val{0.938}{0.013}} & \textbf{\val{0.947}{0.034}} & $1.54e-01$ \\
& & Pre & 0.517 &0.926&0.731&0.310&0.921&0.917& \val{0.899}{0.010} & \val{0.947}{0.019} & \val{0.975}{0.004} & \val{0.940}{0.029} & $3.17e-01$ \\
& & Rec & 0.669 &0.874&0.886&0.939&0.907&0.878& \val{0.971}{0.001} & \val{0.914}{0.045} & \val{0.904}{0.021} & \val{0.954}{0.039} & $9.34e-02$ \\
\midrule
\multirow{6}{*}{\rotatebox{90}{\scriptsize\textbf{BSG}}} & \multirow{6}{*}{\rotatebox{90}{\scriptsize\textbf{ICU}}}
& HR MAE(bpm) & 29.51 &11.35&10.79&155.76&13.67&14.17& \underline{\val{8.43}{1.64}} & \val{10.57}{1.25} & \val{16.25}{1.25} & \textbf{\val{8.17}{0.80}} & $7.89e-01$ \\
& & HRV MAE(ms) & 138.93 &\textbf{95.80}&117.23&143.78&111.37&105.65& \textbf{\val{67.26}{8.88}} & \val{84.52}{4.59} & \val{126.07}{6.24} & \underline{\valsig{83.67}{9.85}{*}} & $4.86e-02$ \\
& & HR MAPE (\%) & 43.07 &\textbf{12.68}&13.79&164.92&14.70&15.25& \textbf{\val{10.81}{2.51}} & \val{16.77}{2.16} & \val{18.25}{0.85} & \underline{\val{13.06}{1.63}} & $1.91e-01$ \\
& & HRV MAPE (\%) & 65.94 &\textbf{57.38}&103.04&86.42&99.11&77.86& \val{39.76}{2.67} & \textbf{\val{35.79}{6.26}} & \val{114.67}{26.67} & \underline{\val{39.04}{4.41}} & $7.91e-01$ \\
& & F1 & 0.297 &\underline{0.834}&0.029&0.511&0.728&0.744& \textbf{\val{0.859}{0.024}} & \val{0.792}{0.030} & \val{0.830}{0.017} & \val{0.833}{0.016} & $1.24e-01$ \\
& & Pre & 0.244 &0.863&0.027&0.359&0.744&0.777& \val{0.837}{0.037} & \val{0.745}{0.037} & \val{0.953}{0.021} & \val{0.783}{0.029} & $6.49e-02$ \\
& & Rec & 0.379 &0.808&0.031&0.936&0.716&0.716& \val{0.883}{0.021} & \val{0.846}{0.024} & \val{0.735}{0.017} & \val{0.890}{0.015} & $5.94e-01$ \\
\bottomrule
\end{tabular}%
}
\end{table}

This section presents a comprehensive evaluation of Peak-Detector's performance against established baselines across six diverse physiological datasets. The quantitative results, encompassing peak detection accuracy (F1-score, Precision, Recall) and physiological consistency (HR MAE, HRV MAE), are summarized in Table~\ref{tab:results_combined}. Comprehensive ablation studies—evaluating the influence of specific training stages and providing performance benchmarks against state-of-the-art online LLMs—are detailed in Appendix~\ref{sec:ablation_study}.

Traditional signal-processing algorithms demonstrate highly variable performance, often exhibiting significant degradation when applied outside their target modality. For instance, Elgendi's algorithm, while effective for PPG, fails to generalize to BCG datasets, confirming the limitations of methods reliant on hand-crafted, modality-specific features. Conversely, deep learning baselines (e.g., FR-Net, 1D-UNet++) demonstrate stronger generalization; however, they remain opaque ``black-boxes,'' lacking the interpretability required for clinical trust.

Our proposed Peak-Detector addresses these challenges, demonstrating performance that is either state-of-the-art or statistically comparable to specialized deep learning models across modalities:

\textbf{Electrocardiography (ECG):} On the MIT-BIH and Incart datasets, Peak-Detector maintains robust detection capabilities. While specialized deep learning baselines like FR-Net achieved statistically lower error rates ($p < 0.001$) on these cleaner signals, Peak-Detector's performance remains highly competitive for clinical utility. This suggests that while specialized architectures may offer marginal gains in low-noise environments, our generative approach provides a viable, high-performance alternative without requiring architecture-specific tuning. We provide a further analysis of failure cases for this dataset in Appendix \ref{sec:failure case analysis}.

\textbf{Photoplethysmography (PPG):} In the optical domain, Peak-Detector demonstrates exceptional efficacy. Notably, on the BIDMC dataset, it achieves state-of-the-art performance with a statistically significant reduction in HR MAE ($0.35$ bpm vs. $0.71$ bpm, $p < 0.01$) compared to the best baseline. On the Capnobase dataset, our model similarly achieves the lowest mean HR MAE ($0.35$ bpm) and highest F1-score ($0.9920$). Although the difference on Capnobase did not reach statistical significance ($p > 0.05$), the results confirm that Peak-Detector is, at a minimum, statistically equivalent to the top-performing specialized models in the PPG domain.

\textbf{Ballistocardiography (BCG) and Seismocardiography (BSG):} On the challenging mechanical signals (Kansas, Arrhy, BSG ICU), Peak-Detector demonstrates its strongest advantage. For the Arrhythmia dataset, our model achieved a statistically significant improvement in HRV MAE ($36.85$ ms vs. $57.44$ ms, $p < 0.05$) against the best baseline, highlighting its superior ability to track heart rate variability in irregular rhythms. On the large-scale BSG ICU dataset, Peak-Detector achieved the lowest mean HR MAE ($8.17$ bpm), numerically outperforming the top deep learning baseline (CNN-SWT: $8.43$ bpm). While this specific difference was not statistically significant ($p > 0.05$), the consistent top-tier performance across these mechanical modalities confirms that Peak-Detector matches the stability of specialized architectures in high-noise environments while offering the distinct advantage of explainability.

To evaluate the generalization capabilities and robustness of our framework beyond single-source data, we performed a series of cross-dataset experiments. Detailed results are documented in Appendix~\ref{sec:Cross_Modality}.

\subsection{Qualitative Analysis}
\label{sec:Interpretation}

\begin{figure*}[htbp]
    \centering

    \begin{subfigure}[b]{0.48\textwidth}
        \centering
        \includegraphics[width=\textwidth]{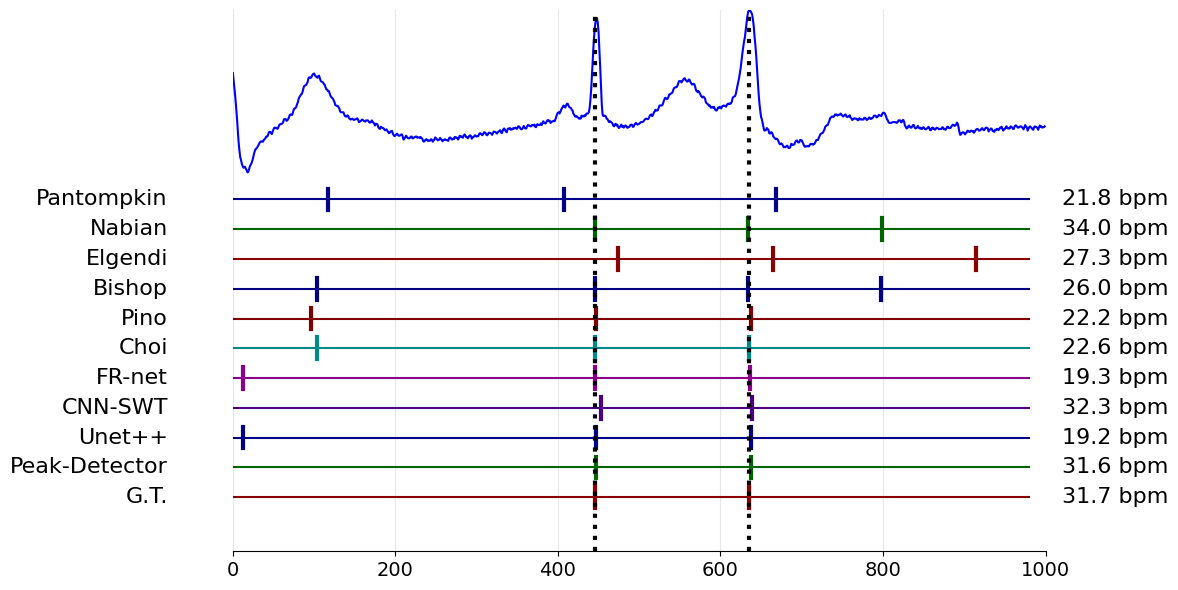}
        \caption{ECG R-Peak Detection}
        \label{fig:ECG_R_peak_visualization}
    \end{subfigure}
    \hfill
    \begin{subfigure}[b]{0.48\textwidth}
        \centering
        \includegraphics[width=\textwidth]{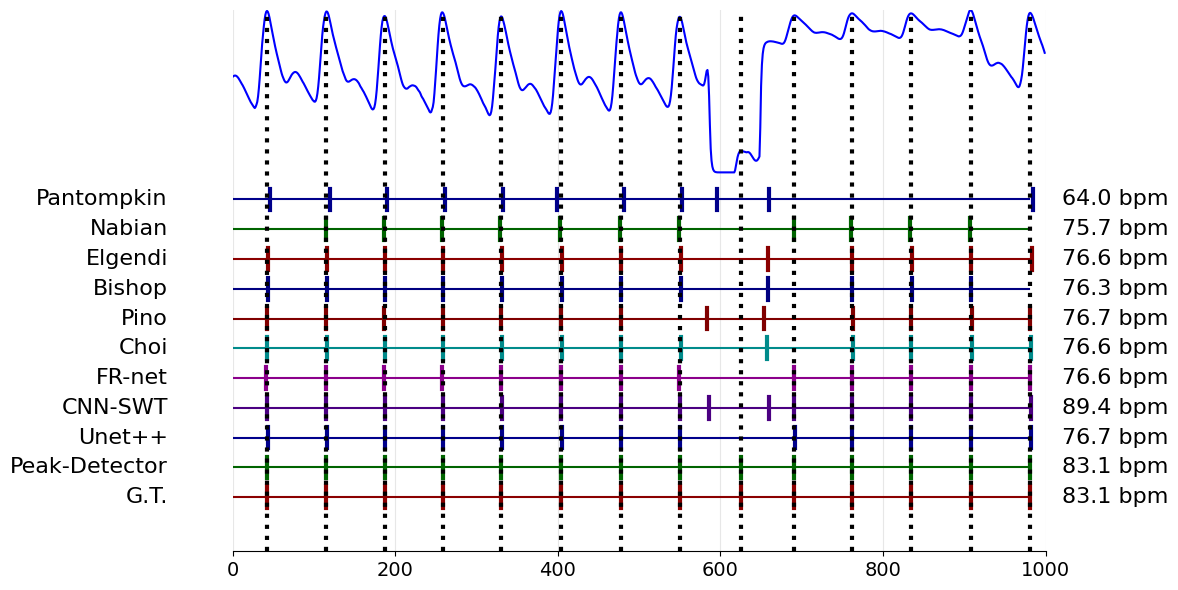}
        \caption{PPG Systolic Peak Detection}
        \label{fig:PPG_peak_visualization}
    \end{subfigure}
    
    \par\medskip
    
    \begin{subfigure}[b]{0.48\textwidth}
        \centering
        \includegraphics[width=\textwidth]{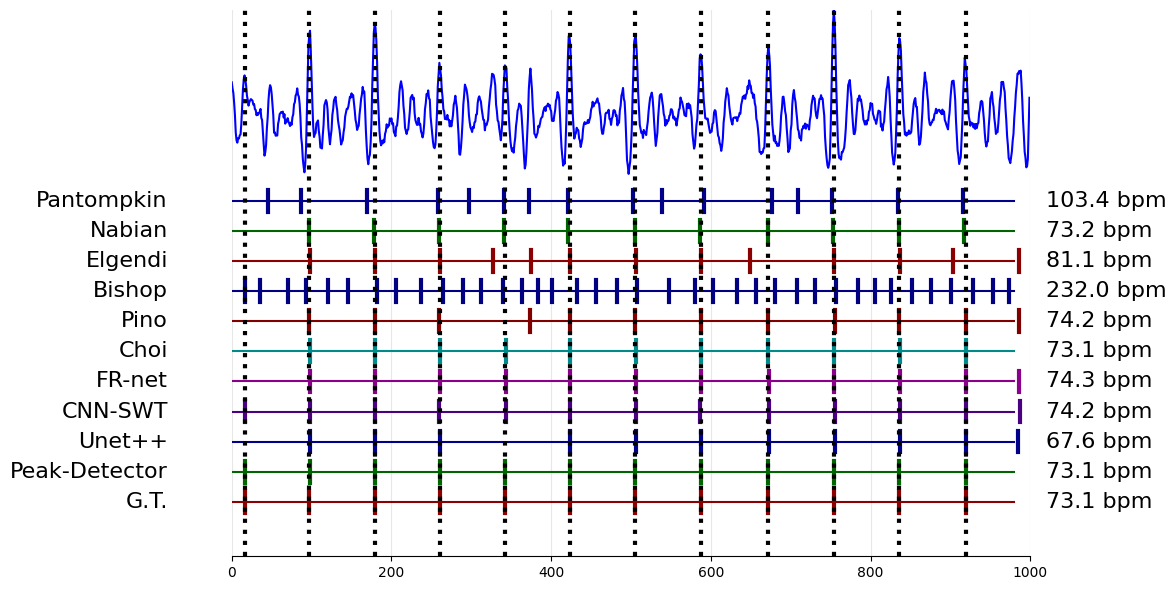}
        \caption{BCG J-Peak Detection}
        \label{fig:BCG_J_peak_visualization}
    \end{subfigure}
    \hfill
    \begin{subfigure}[b]{0.48\textwidth}
        \centering
        \includegraphics[width=\textwidth]{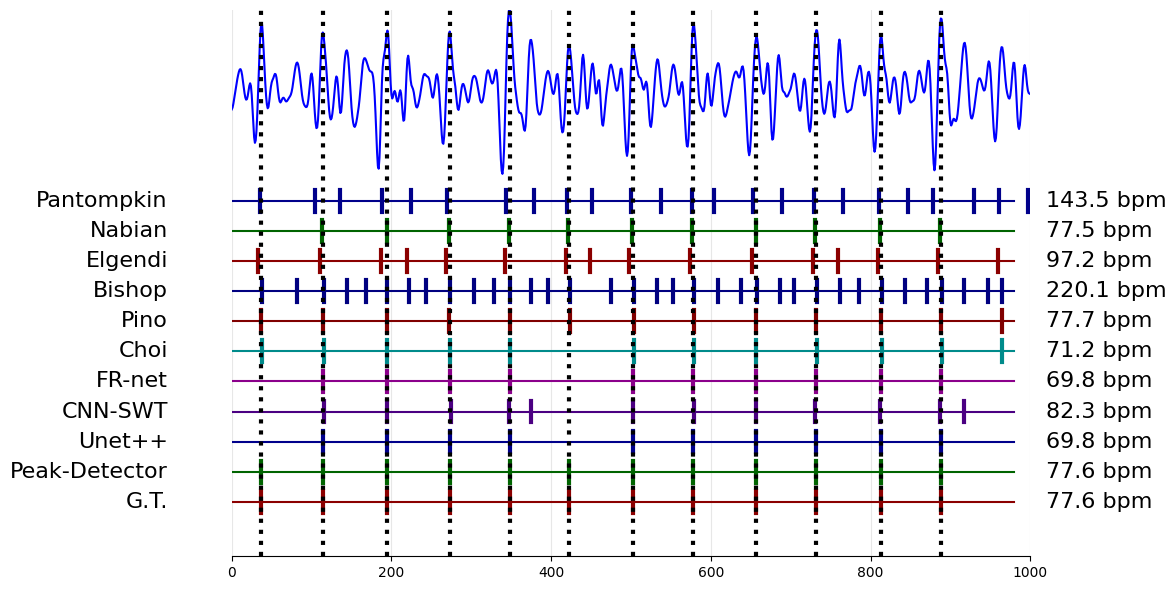}
        \caption{BSG J-Peak Detection}
        \label{fig:BSG_J_peak_visualization}
    \end{subfigure}
    \caption{Qualitative visualization of peak detection performance across challenging segments of (a) ECG with T-wave interference, (b) PPG with noise artifacts, and (c) BCG with arrhythmia (d) BSG in ICU.}
    \Description{Four qualitative signal plots comparing peak detection performance across ECG, PPG, BCG, and BSG examples.}
    \label{fig:Peak_visualization}
\end{figure*}

To provide a qualitative assessment of performance in challenging scenarios, Figure~\ref{fig:Peak_visualization} illustrates the comparative detection results on representative segments of ECG, PPG, BCG, and BSG signals. This visual analysis highlights the practical failure modes of baseline algorithms and underscores the robustness of our proposed method.

The ECG example (Figure~\ref{fig:ECG_R_peak_visualization}) features a prominent, extended T-wave, a common source of error for heuristic-based methods. As observed, several signal-processing algorithms (Pan-Tompkins, Bishop, Pino, Choi) erroneously classify this T-wave as a true R-peak due to its significant amplitude. Other methods, like Elgendi, correctly identify the R-peak's general location but exhibit notable positional inaccuracies. While the deep learning baselines (FR-Net, CNN-SWT) are more robust to the morphological interference, they still show slight deviations from the precise R-peak location, a challenge that Peak-Detector overcomes.

The PPG segment (Figure~\ref{fig:PPG_peak_visualization}) demonstrates the impact of signal quality degradation. In the initial clean portion of the signal, all methods perform reasonably well. However, as the signal becomes corrupted by noise in the latter half, the performance of signal-processing methods deteriorates significantly; Pan-Tompkins fails to detect subsequent peaks entirely, and most other heuristic methods miss the crucial peak around the 700-sample mark. While the data-driven approaches maintain better performance in the noisy region, Peak-Detector is the only method to correctly identify the systolic peak within a morphologically distorted area around the 600-sample mark, showcasing its superior resilience to artifacts.

Then, the BCG arrhythmia example (Figure~\ref{fig:BCG_J_peak_visualization}) highlights the difficulty of analyzing signals with irregular timing and morphology. Traditional algorithms, relying on fixed assumptions about rhythm and shape, struggle significantly; Pan-Tompkins and Elgendi exhibit considerable over-detection, while Bishop and Pino show incorrect localization. In contrast, data-driven approaches, and particularly Peak-Detector, successfully identify the correct J-peak positions even amidst arrhythmia and reduced signal clarity. These qualitative observations across diverse modalities reinforce the quantitative findings, visually confirming that Peak-Detector's contextual reasoning provides superior robustness against morphological anomalies, noise, and rhythm disturbances.

Finally, the BSG ICU example (Figure~\ref{fig:BSG_J_peak_visualization}) illustrates performance on a highly challenging BSG signal, characterized by the elevated noise levels typical of an ICU environment. As observed, traditional algorithms are prone to over-detection, while deep learning models tend towards under-detection in this scenario. Peak-Detector, however, correctly identifies all ground-truth peaks, demonstrating that its contextual reasoning is highly effective at distinguishing true cardiac events from artifacts, even under conditions of both high noise and arrhythmic variability.

\subsection{Explanation Evaluation}

\begin{wrapfigure}{r}{0.32\textwidth}
    \centering
    \includegraphics[width=\linewidth]{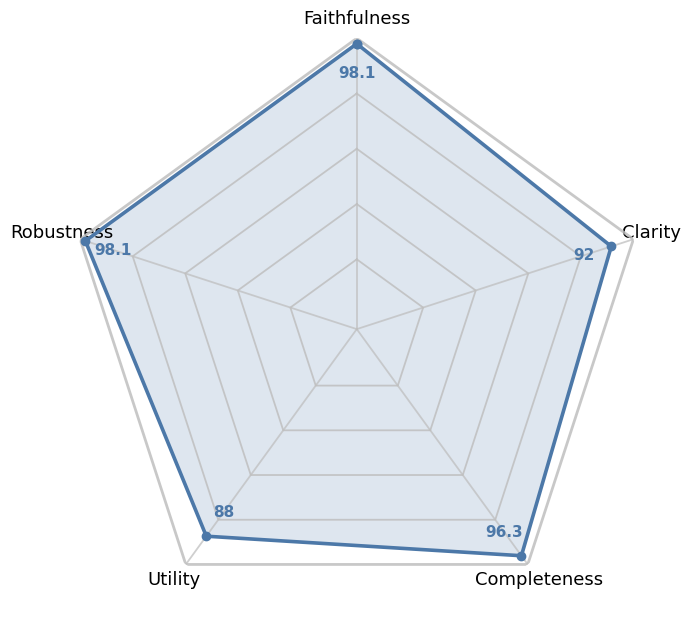}
    \caption{Radar Plot of Explanation Evaluation}
    \Description{Radar chart showing explanation evaluation scores across five interpretability dimensions.}
    \label{fig:Explanation Evaluation}
\end{wrapfigure}

Evaluating the quality of machine-generated explanations is a nuanced challenge. To provide a holistic assessment, we designed a multi-faceted evaluation framework that assesses the generated explanations across five key dimensions. The results of this evaluation, summarized in the radar plot in Figure~\ref{fig:Explanation Evaluation}, demonstrate the high quality and reliability of Peak-Detector's explanatory capabilities.

The evaluation framework assesses the Peak-Detector’s interpretability across five key dimensions, demonstrating high quality and clinical readiness. \textbf{Faithfulness} and \textbf{Robustness} quantify the alignment between qualitative explanations and quantitative signal features, specifically verifying assertions against input data and testing stability under perturbations like clipping or noise; on both metrics, the model achieved a score of 98.1, proving its reasoning is strictly data-grounded and resilient to signal variations. \textbf{Clarity} and \textbf{Utility} were evaluated via user studies to measure intelligibility and practical efficacy, with both receiving scores of 88, confirming the framework's accessibility and its ability to bolster clinician confidence. Finally, \textbf{Completeness}—measured by the inclusion of morphological, temporal, amplitude, and contextual analyses—scored 96.3, highlighting the model's robust instruction-following capabilities and its ability to generate comprehensive, multi-faceted rationales.

\textbf{Analysis of Generated Explanations.}
Table~\ref{tab: explanation} presents a side-by-side comparison between an explanation generated by the "teacher" LLM and one produced by our fine-tuned Peak-Detector, revealing several key insights into our model's interpretability. The analysis confirms that Peak-Detector demonstrates a strong adherence to the predefined explanatory format, consistently integrating multiple analytical dimensions into its reasoning. These dimensions include morphological characteristics (peak shape), temporal relationships (physiological intervals), signal quality criteria (signal-to-noise ratio and amplitude), and waveform context. Furthermore, the model's textual rationale is demonstrably faithful to the underlying signal data, as the provided justification directly supports the final peak selections, showing a strong alignment between its analysis and prediction. Finally, a notable trade-off between conciseness and verbosity is observed; the explanation from Peak-Detector is more succinct than that of the larger teacher model, while still effectively communicating the core logic for its decisions.

\begin{figure}[htbp]
    \centering
    \begin{subfigure}{0.48\textwidth}
        \centering
        \includegraphics[width=\textwidth]{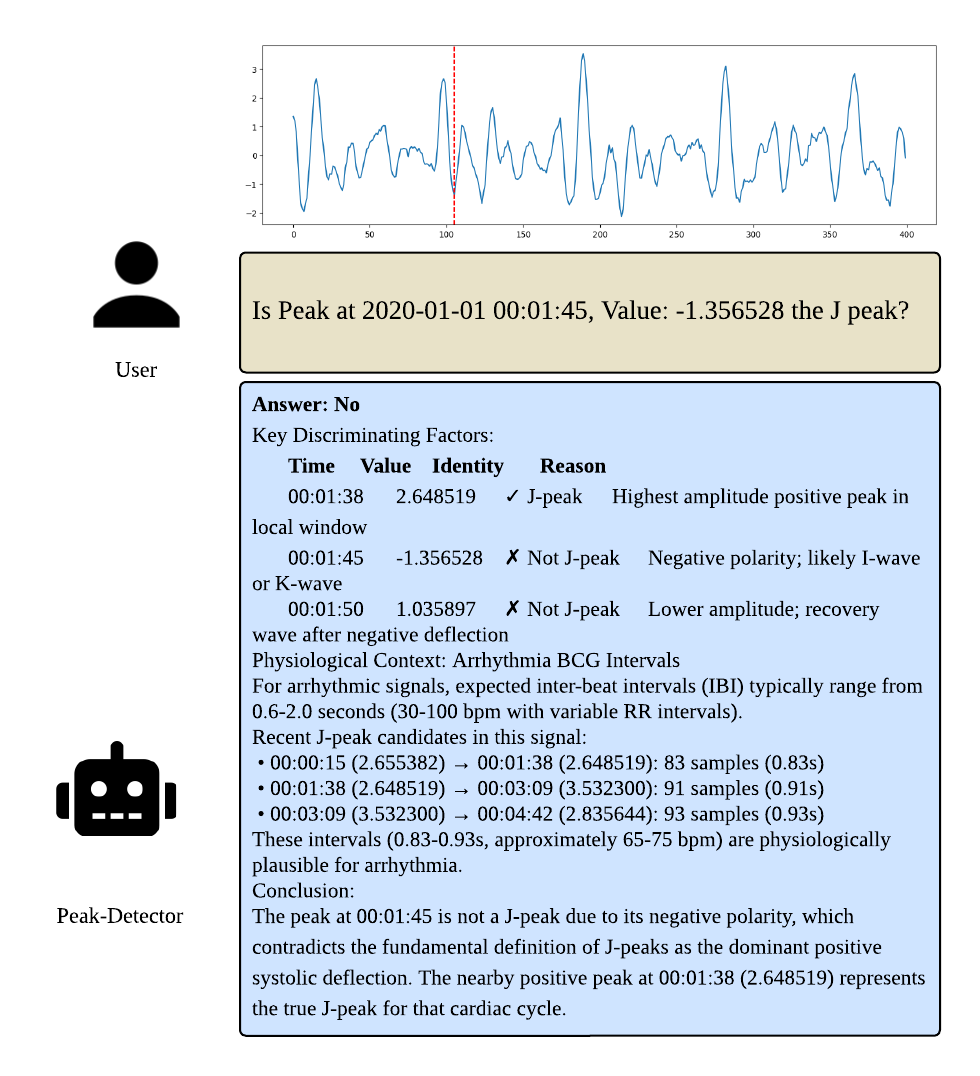}
        \caption{Zero-Shot Peak Classification}
        \label{fig:interaction1}
    \end{subfigure}
    \hfill
    \begin{subfigure}{0.48\textwidth}
        \centering
        \includegraphics[width=\textwidth]{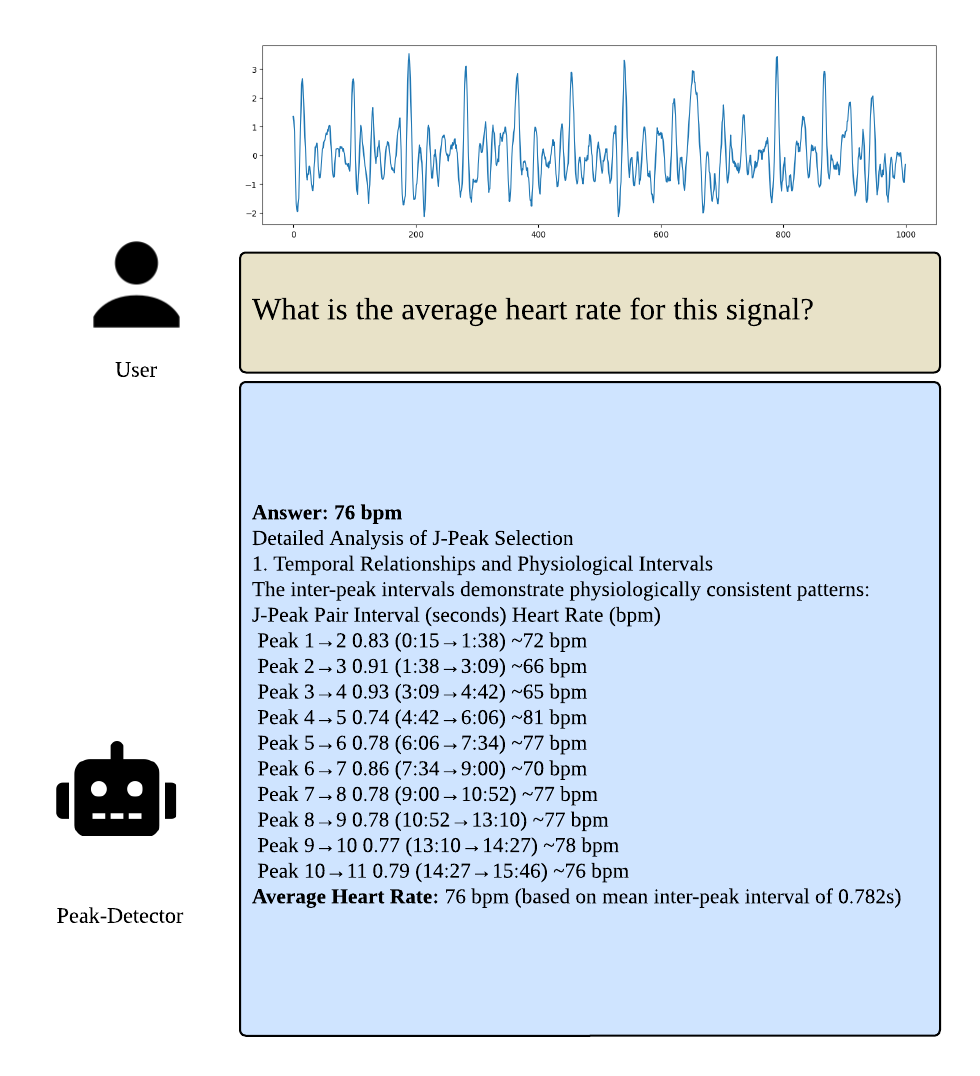}
        \caption{Zero-Shot HR Calculation}
        \label{fig:interaction2}
    \end{subfigure}
    \caption{Demonstration of Peak-Detector's emergent analytical capabilities on zero-shot tasks. (a) The model correctly classifies a non-J-peak and hypothesizes its identity. (b) The model executes a multi-step procedure to accurately calculate heart rate from the detected peaks.}
    \Description{Two zero-shot task examples showing peak classification and heart-rate calculation interactions.}
    \label{fig:interactions_mode}
\end{figure}

\textbf{Emergent Analytical Capabilities.}
A significant advantage of leveraging a pre-trained LLM is its potential to generalize its learned knowledge beyond the explicit fine-tuning task. To explore this, we evaluated Peak-Detector on two zero-shot tasks not present in the training data: fine-grained peak classification and procedural heart rate calculation. The results, shown in Figure~\ref{fig:interactions_mode}, demonstrate that our training cultivates a model with a foundational understanding of the underlying signal, rather than just a narrow peak detection function.

As shown in Figure~\ref{fig:interaction1}, when queried to classify a specific point in the waveform, the model correctly identifies it as a non-J-peak. Critically, it goes further by hypothesizing that the point may correspond to a K-wave. This demonstrates a latent understanding of the complete IJK complex morphology for Peak-Detector. Furthermore, Figure~\ref{fig:interaction2} showcases the model's capacity for multi-step procedural reasoning. When tasked with calculating the heart rate, Peak-Detector spontaneously executes a correct analytical workflow: It first identifies all J-peaks, then calculates the individual peak-to-peak intervals, and finally averages these intervals to derive the mean heart rate.

These emergent capabilities highlight that our framework does not merely produce a static peak detector. Instead, it cultivates an interactive analytical tool with a deeper, more flexible understanding of cardiac signals, showcasing the broader potential for LLMs to serve as versatile partners in physiological data analysis.

\section{ANALYSIS}
\label{analysis}

\subsection{Validation of Peak Representation}
\label{sec:compression_ratio}

\begin{figure}[ht]
\centering

\begin{subfigure}[b]{0.32\textwidth}
    \centering
    \includegraphics[width=\textwidth]{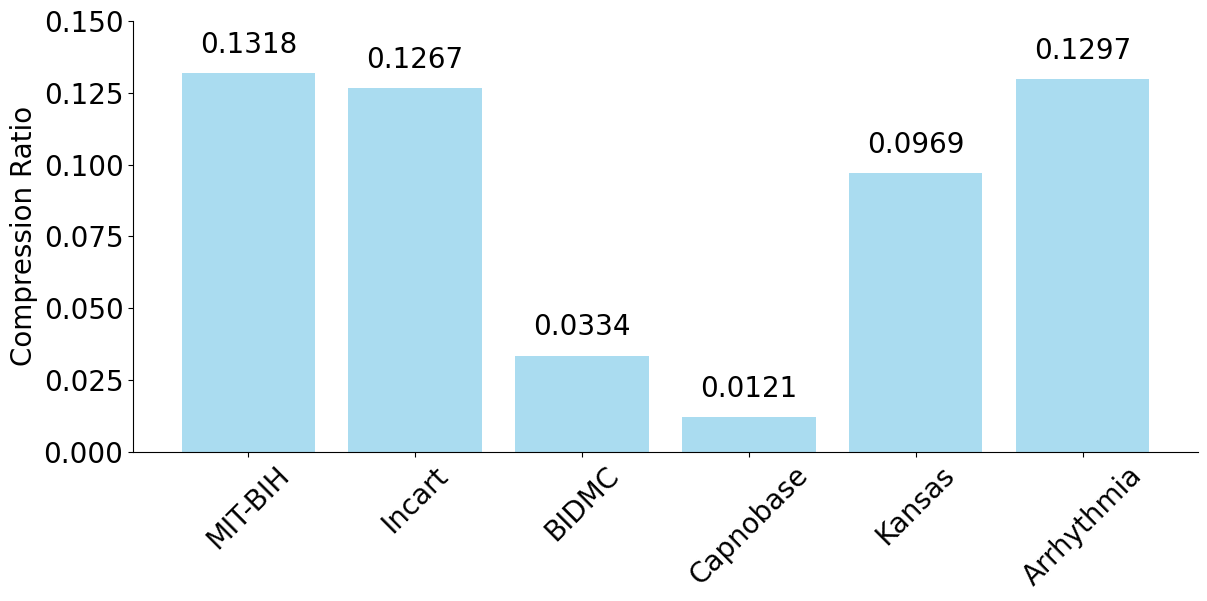}
    \caption{Compression Ratio}
    \label{fig:compression_ratio}
\end{subfigure}
\hfill
\begin{subfigure}[b]{0.32\textwidth}
    \centering
    \includegraphics[width=\textwidth]{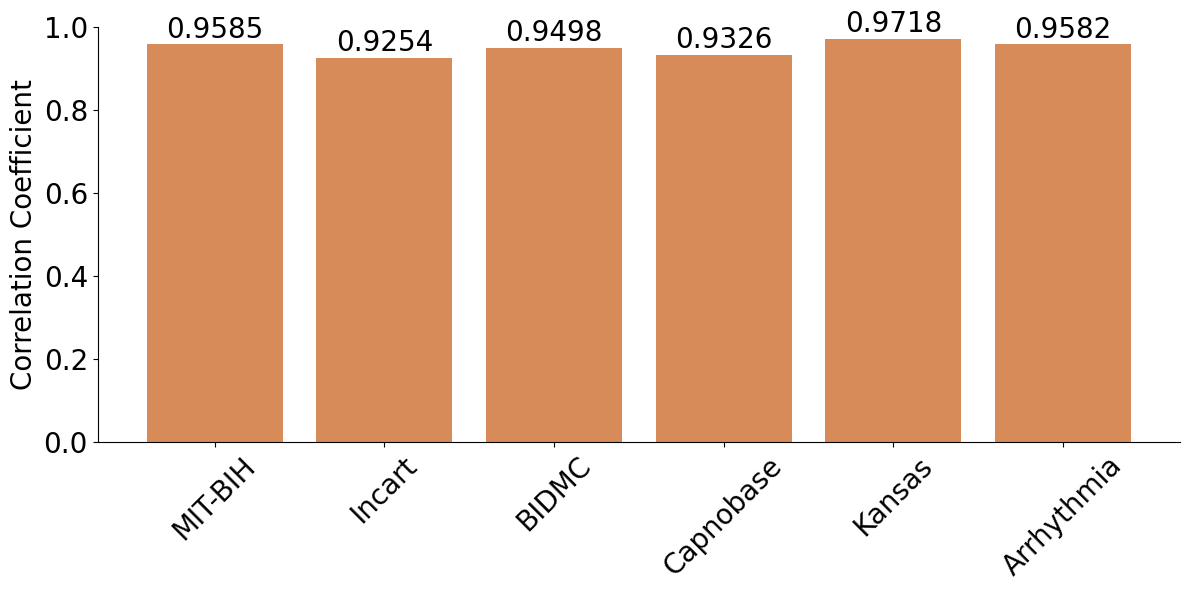}
    \caption{Correlation Coefficient}
    \label{fig:correlation_coefficient}
\end{subfigure}
\hfill
\begin{subfigure}[b]{0.32\textwidth}
    \centering
    \includegraphics[width=\textwidth]{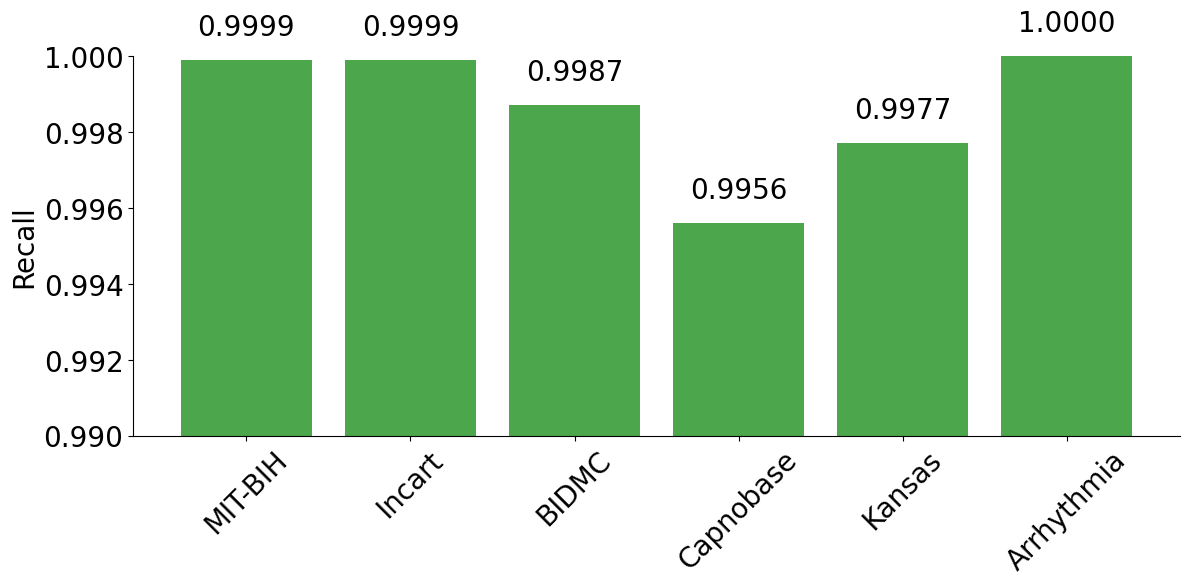}
    \caption{Preliminary Peak Recall}
    \label{fig:recall}
\end{subfigure}
\caption{Validation of the Peak Representation. (a) Data Retention Ratio, showing the percentage of data points remaining after transformation. (b) Pearson correlation between original and reconstructed signals, demonstrating high signal fidelity. (c) Recall of prominent peaks in the initial candidate set, confirming completeness.}
\Description{Three plots showing compression ratio, reconstruction correlation, and preliminary peak recall for the peak representation.}
\label{fig:compression_and_correlation}
\end{figure}

To validate the Peak Representation framework, we evaluated its performance across three critical metrics: data compression efficiency, signal reconstruction fidelity, and peak set completeness. Efficiency was quantified via a \textbf{Data Retention Ratio}, defined as the cardinality of the Peak Representation relative to the original signal length. As illustrated in Figure~\ref{fig:compression_and_correlation}(a), the framework achieves substantial compression across modalities, with mean retention ratios of 13.2\% for ECG and 11.3\% for BCG; notably, the smoother PPG waveforms required as little as 1.2\% of the original data.
To ensure information preservation, we assessed reconstruction fidelity using cubic spline interpolation. The resulting \textbf{Pearson correlation coefficients} (Figure~\ref{fig:compression_and_correlation}(b)) consistently exceeded 0.90 across all datasets, confirming the retention of vital temporal dynamics. Finally, we verified the completeness of the initial extraction via \textbf{Prominent Peak Recall}. As shown in Figure~\ref{fig:compression_and_correlation}(c), the framework achieved a near-perfect recall (minimum 0.9956), ensuring that all ground-truth peaks are captured. This transforms the detection task into a high-confidence filtering and reasoning challenge for the LLM. Detailed tabulations are provided in Section~\ref{appendix:Peak Representation Analysis}.

\begin{figure*}[!t]
\centering

\begin{subfigure}[b]{0.32\textwidth}
    \includegraphics[width=\textwidth]{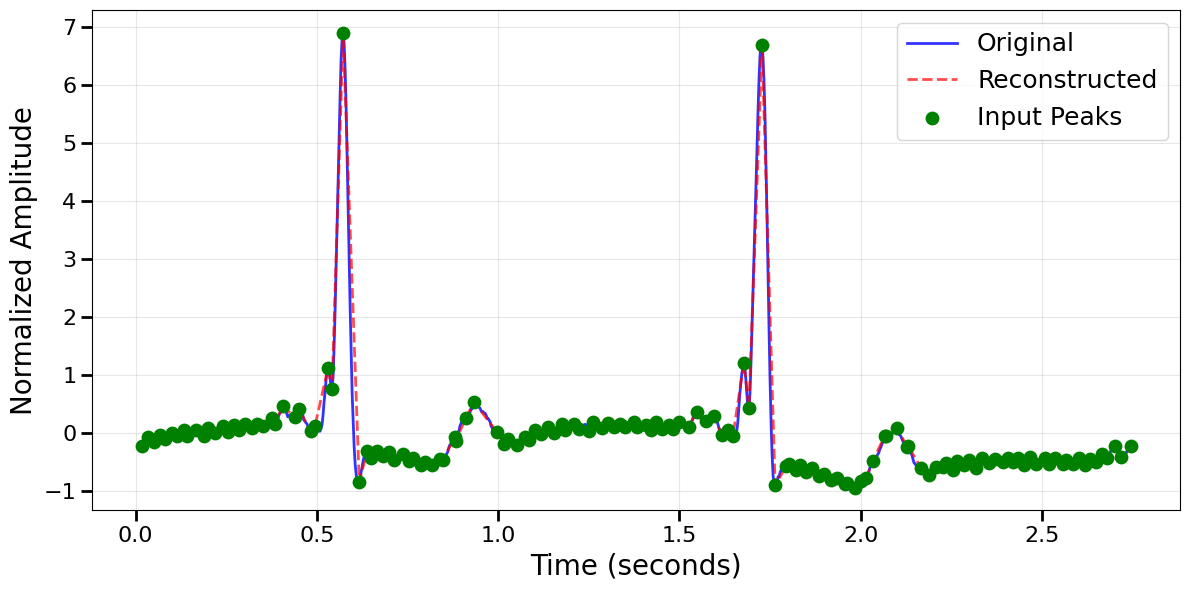}
    \caption{ECG Reconstruction}
    \label{fig:ECG_reconstruction}
\end{subfigure}
\hfill
\begin{subfigure}[b]{0.32\textwidth}
    \includegraphics[width=\textwidth]{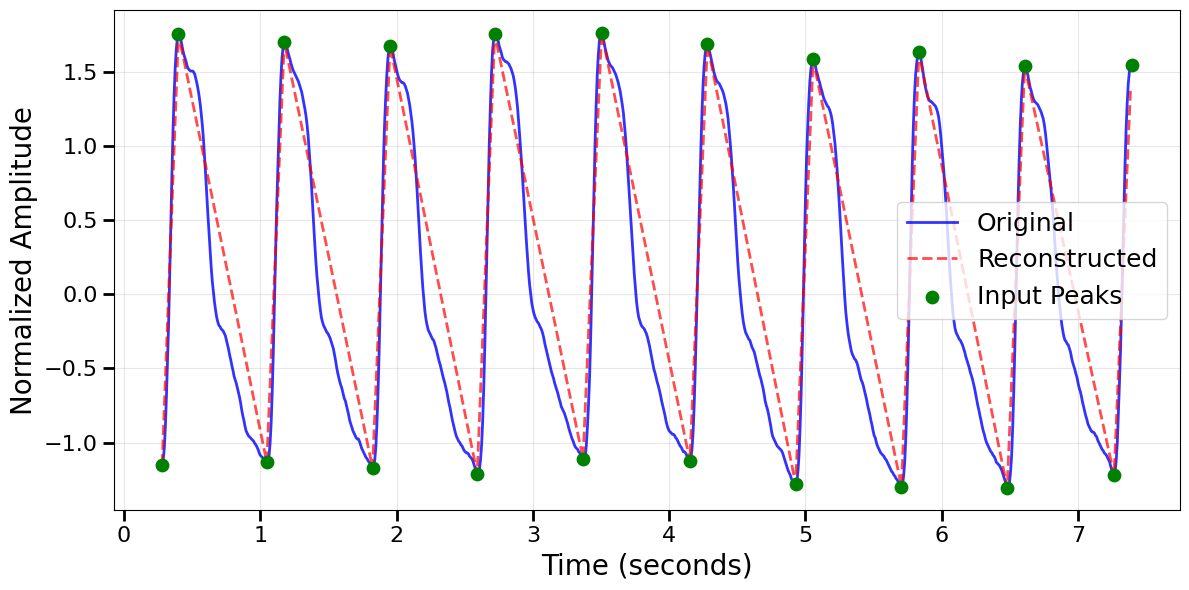}
    \caption{PPG Reconstruction}
    \label{fig:PPG_reconstruction}
\end{subfigure}
\hfill
\begin{subfigure}[b]{0.32\textwidth}
    \includegraphics[width=\textwidth]{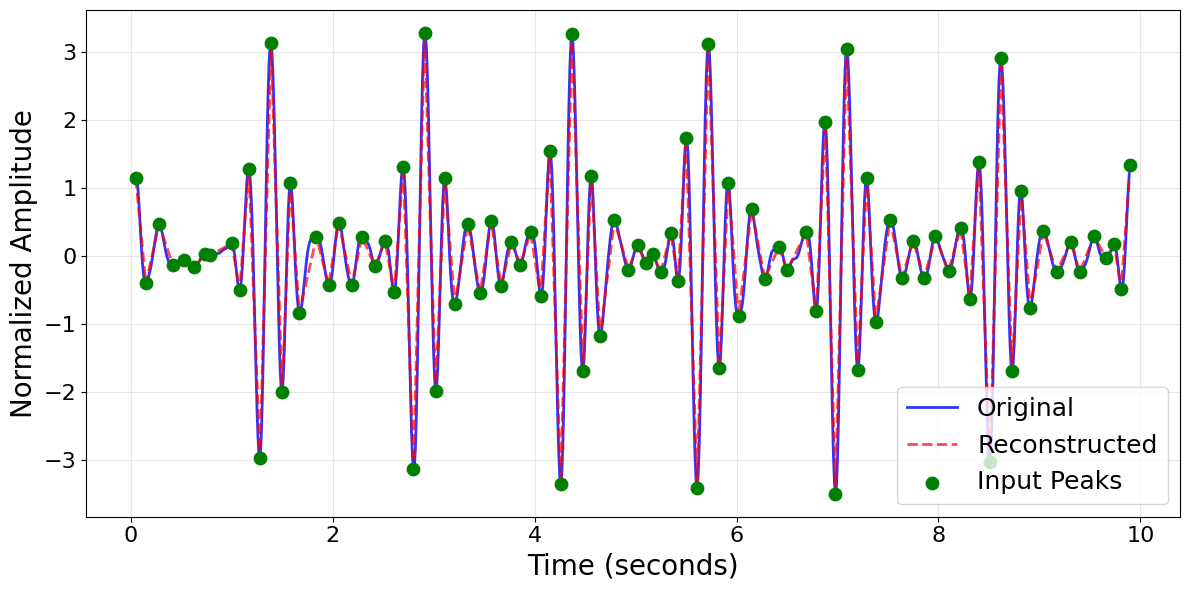}
    \caption{BCG Reconstruction}
    \label{fig:BCG_reconstruction}
\end{subfigure}

\caption{Qualitative comparison of original signals (blue) and their reconstructions (orange) from the sparse Peak Representation. The visualizations demonstrate high fidelity across modalities: (a) An ECG segment from MIT-BIH, where high-frequency components are captured. (b) A PPG segment from BIDMC, showing excellent approximation from only major peaks and troughs. (c) A BCG segment from the Kansas dataset, where the complex multi-phasic waveform is effectively outlined.}
\Description{Original and reconstructed ECG, PPG, and BCG waveforms overlaid to compare reconstruction fidelity.}
\label{fig:Reconstruction}
\end{figure*}

Figure~\ref{fig:Reconstruction} qualitatively illustrates the Peak Representation by comparing original physiological signals with reconstructions derived from sparse peak data. For ECG signals (Fig~\ref{fig:ECG_reconstruction}), the reconstruction accurately tracks the high-frequency QRS complex while capturing subtle baseline fluctuations, accounting for the modality's moderate compression ratio. Conversely, the PPG waveform (Fig~\ref{fig:PPG_reconstruction}), characterized by smoother morphology and information density concentrated at systolic and diastolic extrema, achieves a superior compression ratio with high visual fidelity. The BCG signal (Fig~\ref{fig:BCG_reconstruction}) represents an intermediate complexity; the representation effectively preserves the multi-component inflection points (e.g., I, J, K waves) essential to its contour. These visualizations confirm that the framework successfully retains critical morphological characteristics across diverse signal types, establishing a robust foundation for LLM-based contextual analysis.

To isolate the contribution of the LLM’s sequential reasoning capabilities, we performed a comparative analysis against established benchmarks. Using the raw features from the Peak Representation (e.g., peak index and amplitude), we trained several traditional machine learning classifiers—Random Forest \cite{rigatti2017random}, Logistic Regression \cite{lavalley2008logistic}, and XGBoost \cite{chen2016xgboost}. These models were tasked with the same binary classification objective: distinguishing true prominent peaks from candidate noise or artifacts. This experimental design ensures that any performance gains observed in our framework can be specifically attributed to the LLM's contextual processing rather than the input features themselves.

\begin{figure}[ht]
    \centering
    \includegraphics[width=0.65\textwidth]{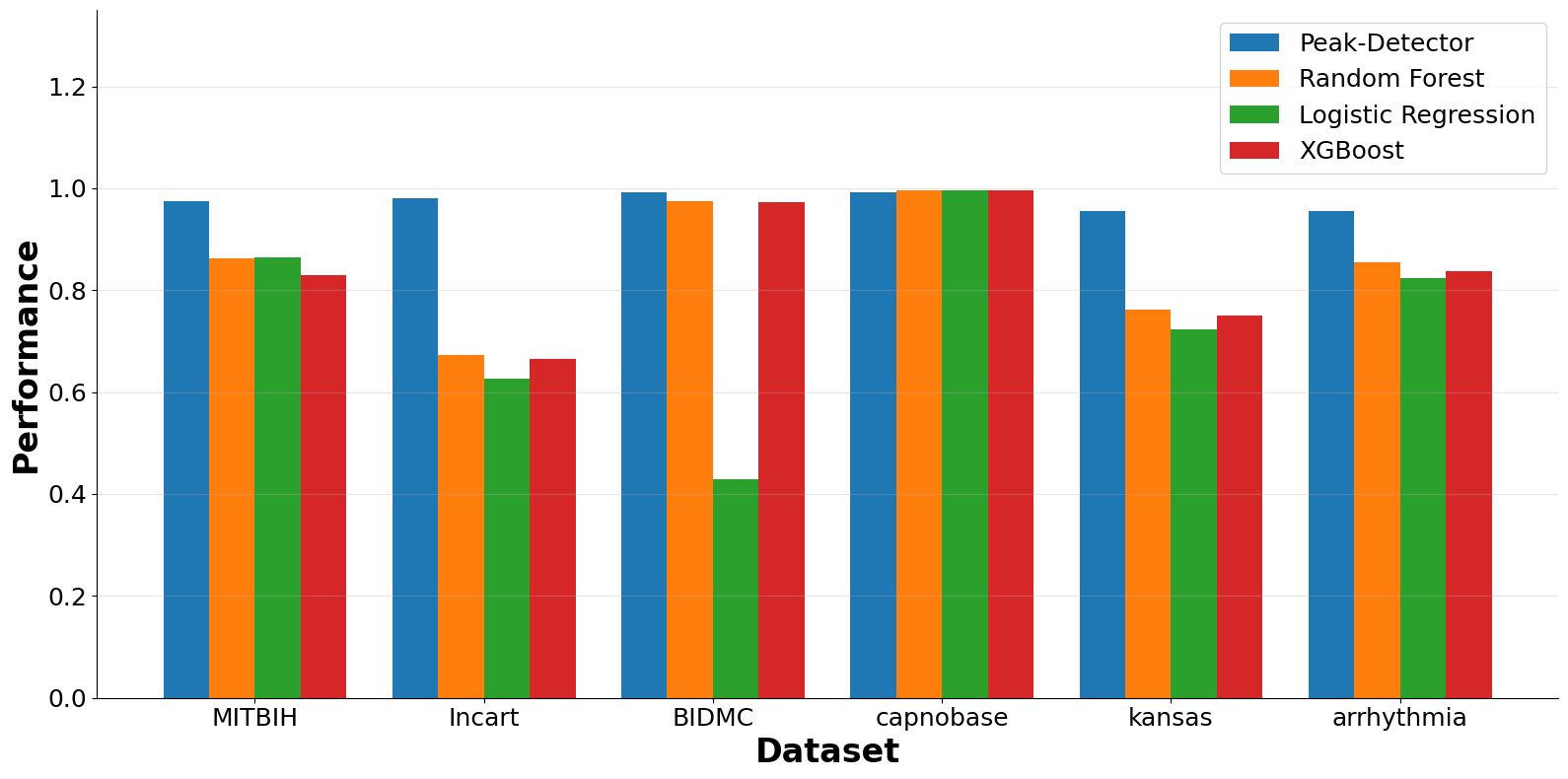}
    \caption{F1-score comparison between Peak-Detector and traditional machine learning classifiers trained on features from the Peak Representation. The performance gap is minimal for simpler PPG signals but substantial for morphologically complex ECG and BCG signals, highlighting the value of the LLM's sequential reasoning.}
    \Description{Bar chart comparing F1 scores for Peak-Detector and traditional classifiers across physiological datasets.}
    \label{fig:comparison}
\end{figure}

Results in Figure~\ref{fig:comparison} illustrate a performance gap dictated by signal modality. For morphologically stable PPG signals (BIDMC, Capnobase), traditional classifiers perform competitively, suggesting that local features within the Peak Representation suffice for systolic peak identification. Conversely, Peak-Detector significantly outperforms baselines on complex ECG and BCG signals. This divergence underscores that resolving R-peaks from T-waves, or J-peaks within an IJK complex, requires a holistic understanding of sequential patterns and morphological context. Ultimately, these findings confirm that the framework’s primary advantage is the LLM’s unique capacity for contextual reasoning, which is indispensable for interpreting high-complexity physiological waveforms.

\subsection{Sensitivity Analysis of Peak Representation Parameters}
\label{appendix:Peak Representation Analysis}

\begin{figure*}[htbp]
    \centering
    \begin{subfigure}[b]{0.32\textwidth}
        \includegraphics[width=\textwidth]{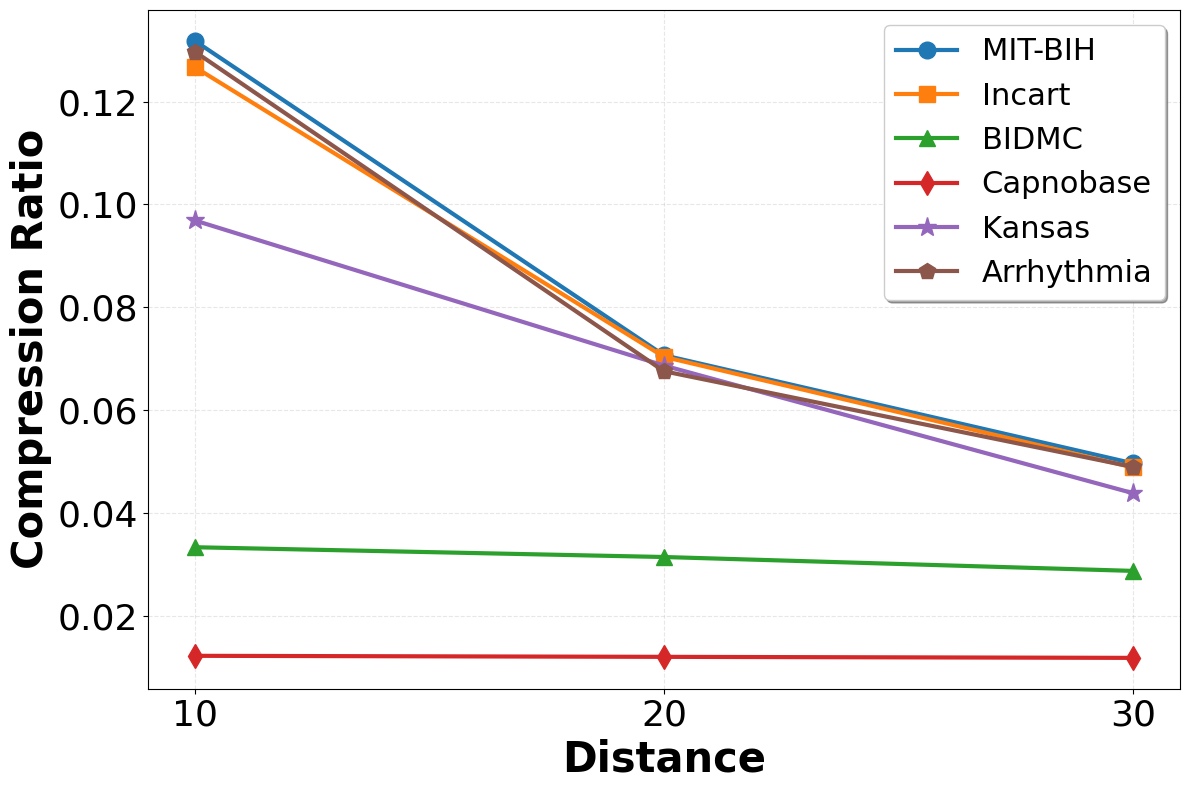}
        \caption{Data Retention Ratio}
        \label{fig:compression_ratio_vs_distance}
    \end{subfigure}
    \hfill
    \begin{subfigure}[b]{0.32\textwidth}
        \includegraphics[width=\textwidth]{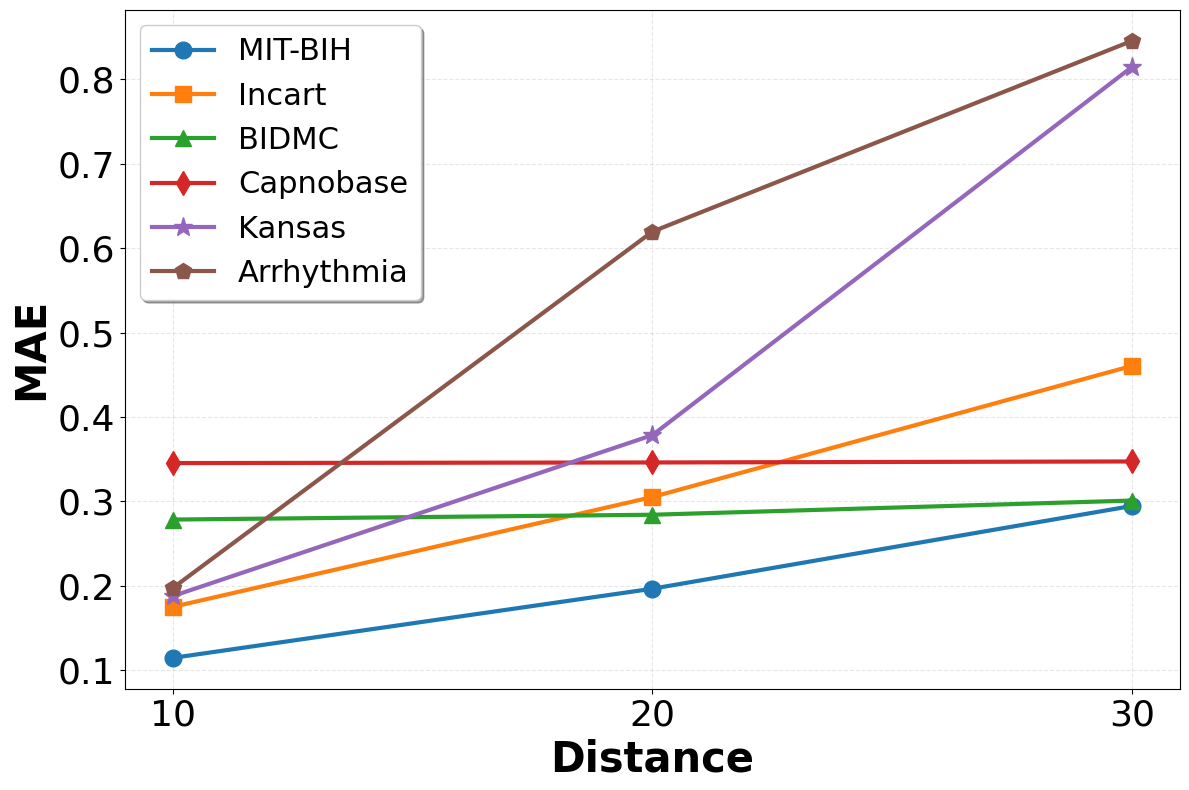}
        \caption{MAE}
        \label{fig:mae_vs_distance}
    \end{subfigure}
    \hfill
    \begin{subfigure}[b]{0.32\textwidth}
        \includegraphics[width=\textwidth]{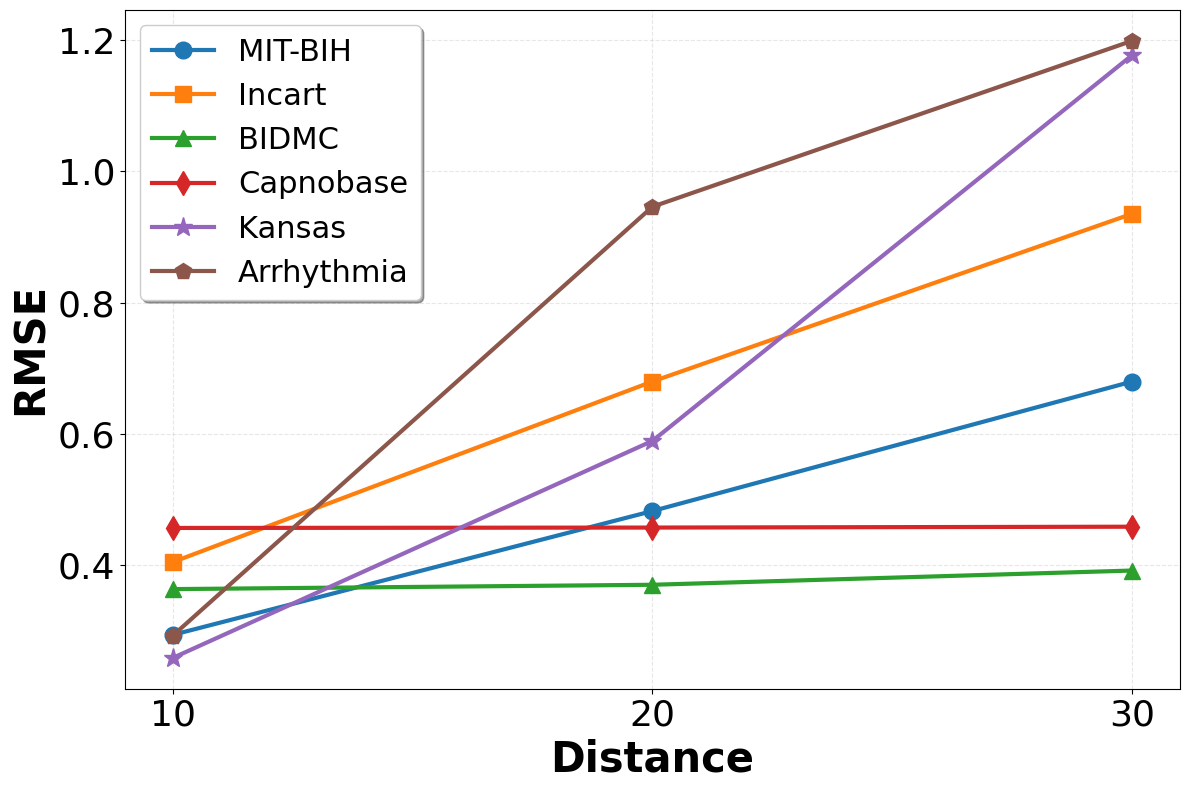}
        \caption{RMSE}
        \label{fig:rmse_vs_distance}
    \end{subfigure}
    
    \par\medskip
    
    \hspace*{\fill}
    \begin{subfigure}[b]{0.32\textwidth}
        \includegraphics[width=\textwidth]{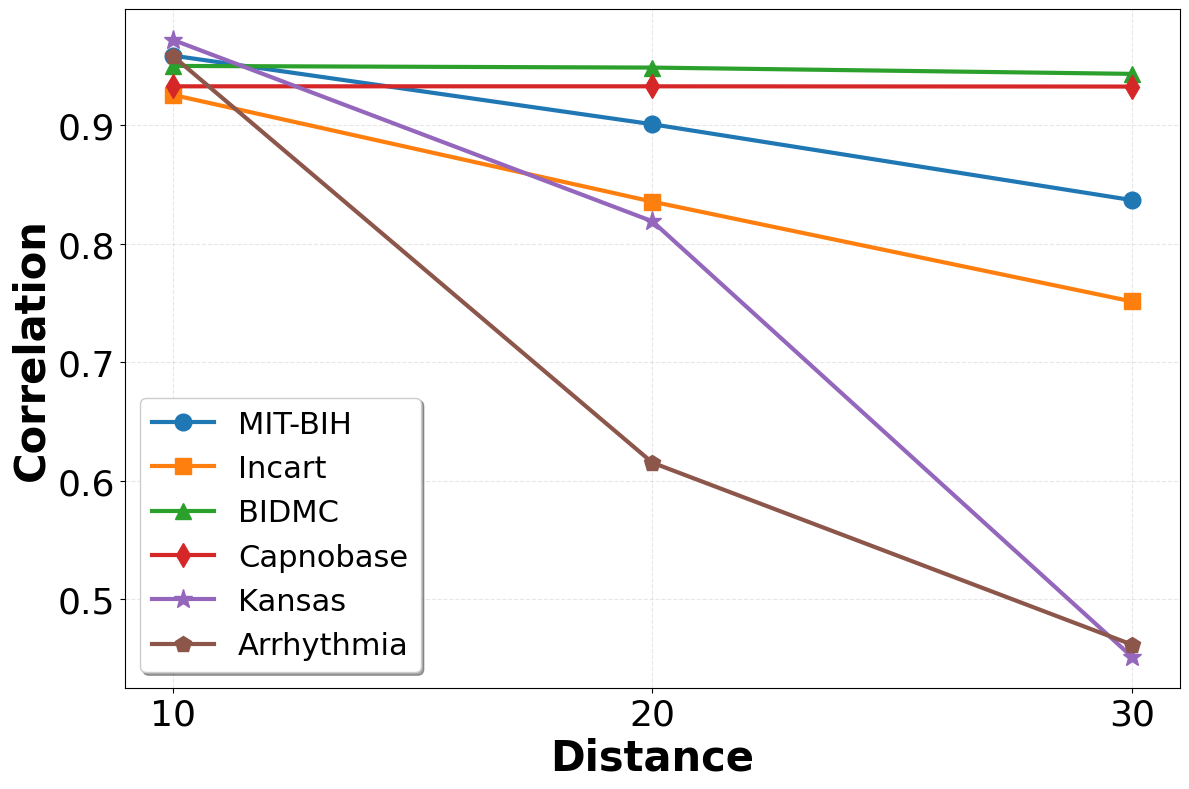}
        \caption{Correlation}
        \label{fig:correlation_vs_distance}
    \end{subfigure}
    \hspace{0.02\textwidth}
    \begin{subfigure}[b]{0.32\textwidth}
        \includegraphics[width=\textwidth]{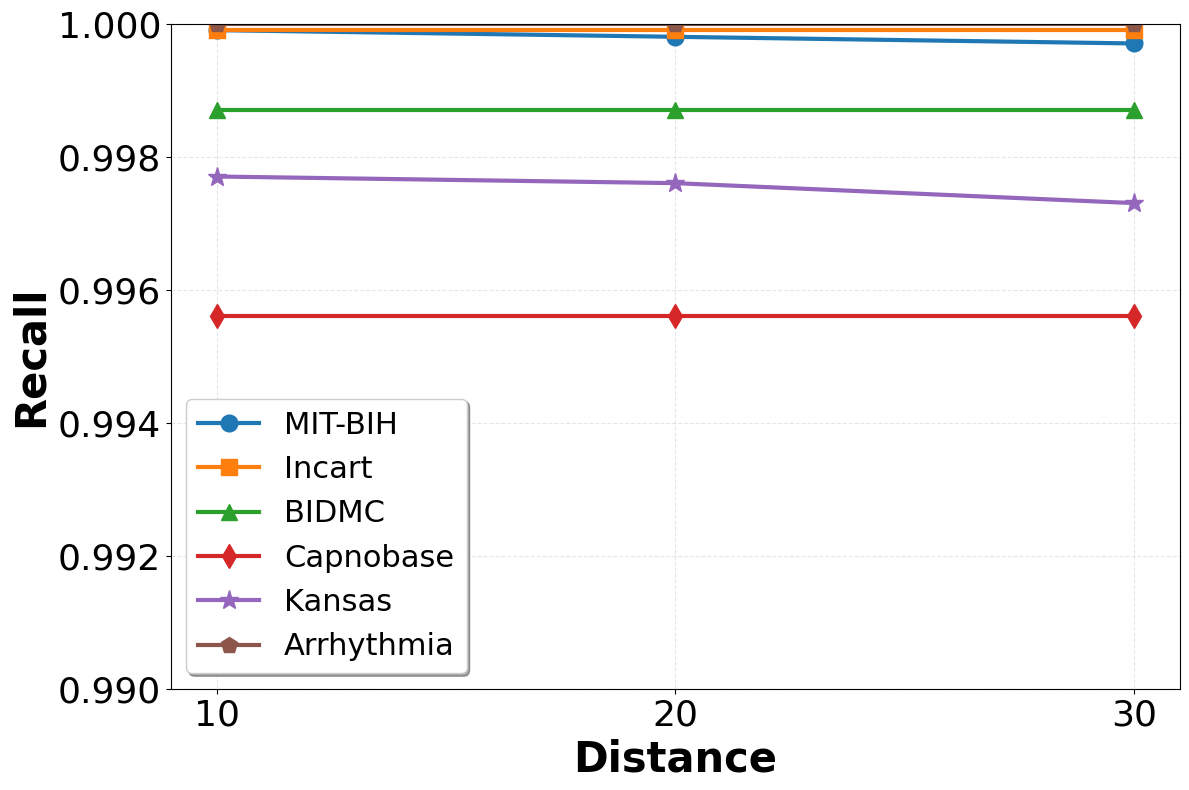}
        \caption{Peak Recall}
        \label{fig:recall_vs_distance}
    \end{subfigure}
    \hspace*{\fill}
    \caption{Sensitivity analysis of Peak Representation metrics as a function of the minimum horizontal distance parameter in peak extraction. As distance increases, data retention decreases, while MAE, RMSE, correlation and prominent peak recall vary, demonstrating the trade-off between compression and signal fidelity across different physiological modalities.}
    \Description{Sensitivity plots showing how compression, error, correlation, and peak recall change with the peak-distance parameter.}
    \label{fig:metrics_vs_distance}
\end{figure*}

The effectiveness of our Peak Representation is inherently linked to the minimum horizontal distance parameter used in the initial peak extraction. To understand this trade-off between data compression and signal fidelity, we conducted a sensitivity analysis by varying this distance, with the results presented in Figure~\ref{fig:metrics_vs_distance}. The analysis reveals a clear inverse relationship: increasing the distance improves compression (i.e., lowers the data retention ratio) but degrades signal reconstruction fidelity, as evidenced by rising MAE and RMSE values and a corresponding drop in correlation. Crucially, the prominent peak recall remains consistently high, confirming that the essential target peaks are robustly captured regardless of the compression level. The impact of this trade-off varies significantly by modality; the smoother PPG signals are least affected, while the complex, multi-phasic BCG signals are highly sensitive, showing a sharp drop in correlation from 0.9582 to 0.4616. Based on this analysis, we selected a final parameter value that ensures reconstruction correlation remains above 0.90, striking an effective balance between substantial data reduction and the preservation of critical information for the LLM.

\begin{figure}[htbp]
    \centering
    \includegraphics[width=0.4\linewidth]{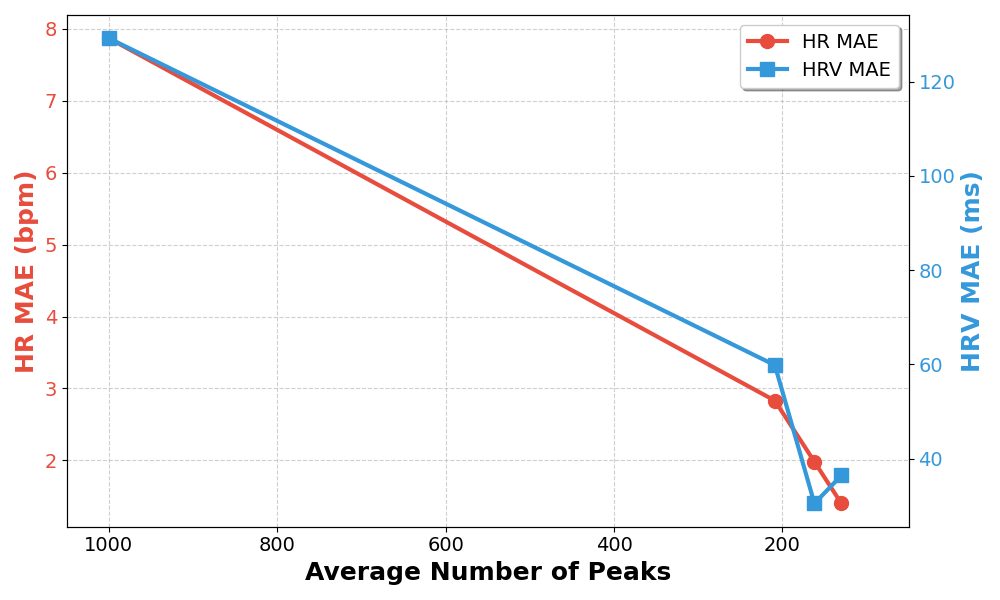}
    
    \caption{Impact of the distance parameter on detection error. The plot demonstrates the trade-off between the average number of peaks retained and the resulting MAE for HR and HRV.}
    \Description{Line plot showing how retained peak count affects HR and HRV mean absolute error.}
    \label{fig:scaling_peaks}
\end{figure}

To evaluate the impact of the Peak Representation Parameter on the performance of the Peak Detector, we controlled the minimum distance parameter across the set $\{0, 2, 5, 10\}$ using the Arrhythmia Dataset. This modulation resulted in average peak counts of $1,000$, $209$, $162$, and $130$, respectively. 
We evaluated the detection accuracy using Heart Rate (HR) Mean Absolute Error (MAE) and Heart Rate Variability (HRV) MAE, as shown in Fig.~\ref{fig:scaling_peaks}. The results indicate that performance consistently improves as the average number of detected peaks decreases. This trend highlights the benefit of primarily excluding noisy peaks to enhance signal fidelity.

\subsection{Impact of Model Scale}
\label{sec:scaling}

To investigate the relationship between model capacity and task performance, we conducted a scaling analysis by fine-tuning several variants of our base LLM architecture, with parameter counts ranging from 0.5 billion to 7 billion. The performance of each model variant was evaluated on the challenging BCG Arrhythmia Dataset.

\begin{figure*}[!t]
\centering

\begin{subfigure}[b]{0.32\textwidth}
    \includegraphics[width=\textwidth]{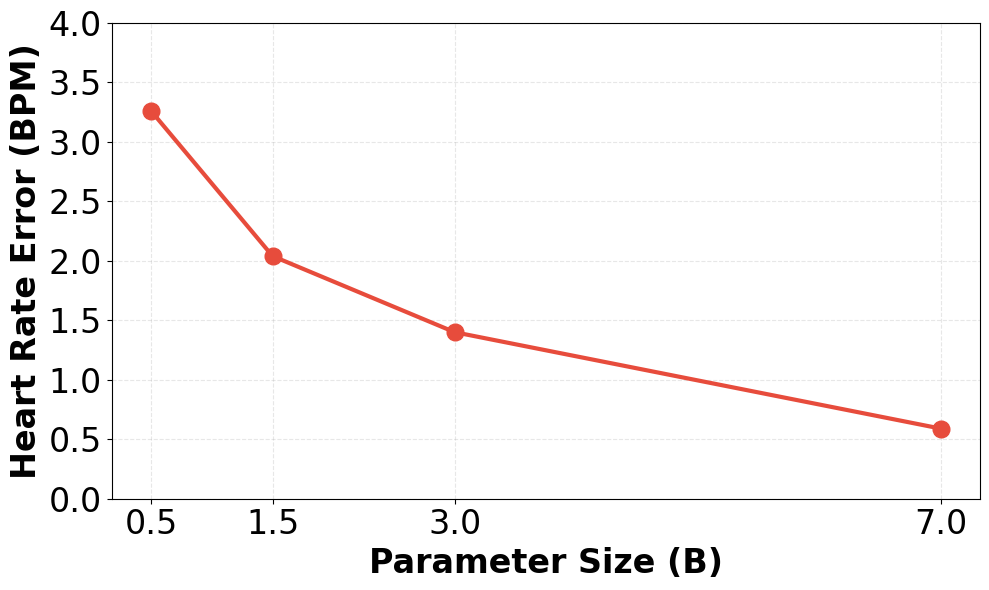}
    \caption{HR MAE}
    \label{fig:scale_HR}
\end{subfigure}
\hfill
\begin{subfigure}[b]{0.32\textwidth}
    \includegraphics[width=\textwidth]{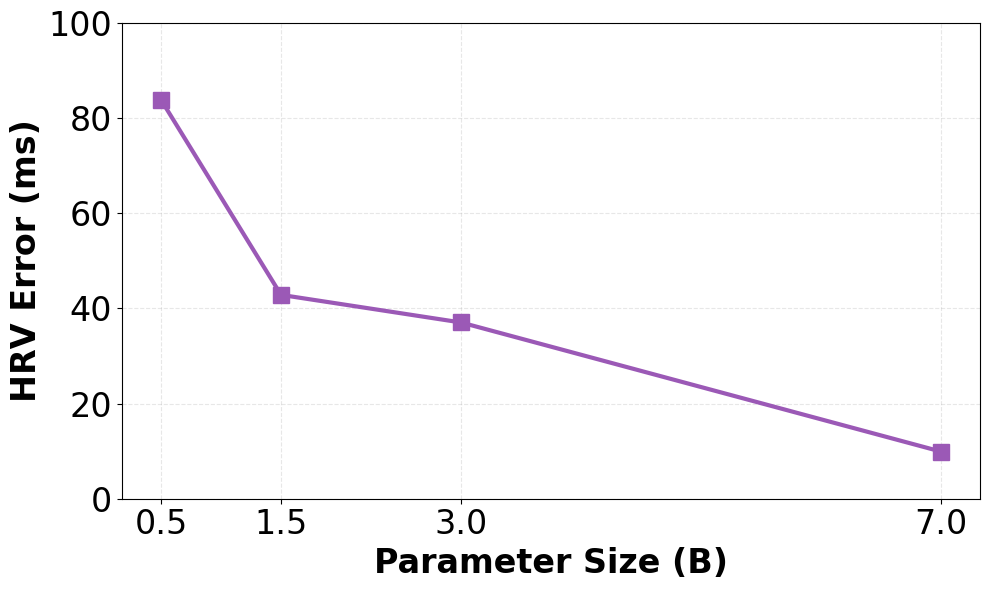}
    \caption{HRV MAE}
    \label{fig:scale_HRV}
\end{subfigure}
\hfill
\begin{subfigure}[b]{0.32\textwidth}
    \includegraphics[width=\textwidth]{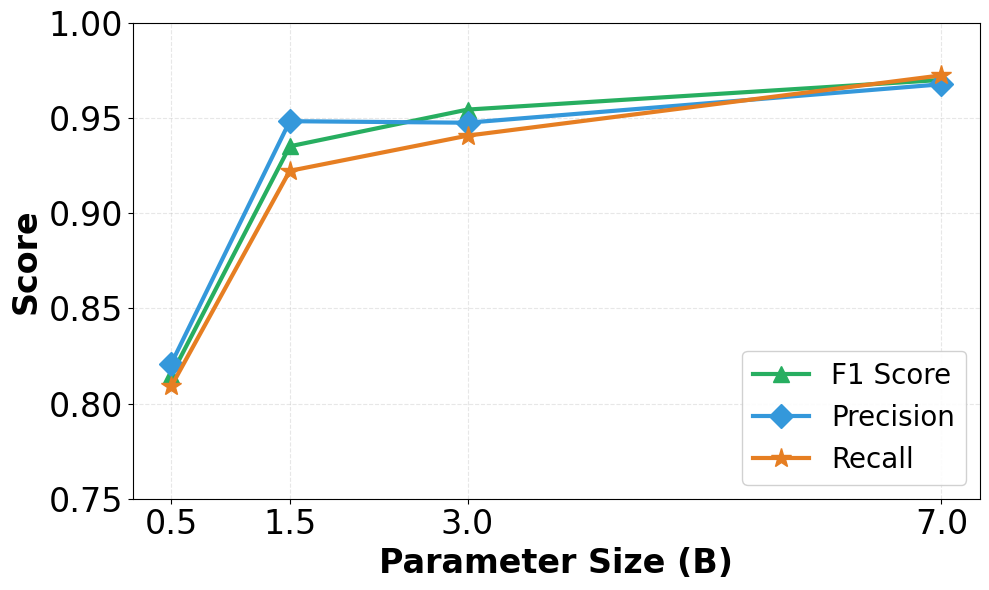}
    \caption{F1, Precision, and Recall}
    \label{fig:scale_F1}
\end{subfigure}
\caption{The impact of model scaling on Peak-Detector performance. As model size increases from 0.5B to 7B parameters, we observe consistent improvements in (a) HR MAE, (b) HRV MAE, and (c) F1/precision/recall under the same protocol. This demonstrates meaningful headroom and enables a practical compute--accuracy trade-off while retaining the same Peak-Detector interface (peak outputs + rationales).}
\Description{Three plots showing model-size scaling effects on heart-rate error, heart-rate-variability error, and detection metrics.}
\label{fig:scale}
\end{figure*}

As depicted in Figure~\ref{fig:scale}, the results demonstrate a clear and consistent scaling law: performance across all metrics improves monotonically with the number of model parameters. Specifically, as the model size increased from 0.5B to 7B, the HR MAE decreased substantially from 3.26 to 0.59, and the HRV MAE dropped from 83.85 to 9.91. Concurrently, the F1-score rose from 0.8150 to 0.9701, driven by marked improvements in both precision (from 0.8210 to 0.9678) and recall (from 0.8091 to 0.9725).

This strong positive correlation suggests that larger models possess a greater capacity to learn the complex, non-linear patterns and subtle morphological cues inherent in arrhythmic physiological signals. The concurrent improvement in both detection accuracy (F1-score) and physiological consistency (HR/HRV MAE) indicates that increased model scale enhances not just pattern matching but also a deeper, more context-aware understanding of the underlying cardiac rhythm. This scaling trend suggests that the performance of the Peak-Detector framework is not yet saturated and could be further enhanced by leveraging even larger foundational models, affirming that the principles of scaling laws extend effectively to this specialized domain of time-series analysis. In contrast, the deep neural networks under evaluation fail to demonstrate performance gains commensurate with model scaling (see Appendix~\ref{sec:scale} for a detailed scaling analysis).

\subsection{Computational Complexity Analysis}
\label{sec:complexity}

\begin{table}[ht]
 \caption{Computational complexity and throughput analysis. Throughput is measured in segments per second (SPS). The analysis highlights the trade-off: Peak-Detector is slower than lightweight numerical algorithms but orders of magnitude faster than general-purpose LLMs and human interpretation.}
 \label{tab:complexity}
 \centering
 \resizebox{0.6\columnwidth}{!}{
 \begin{tabular}{llcc}
    \toprule
    \textbf{Model/Algorithm} & \textbf{Parameters} & \textbf{Training (SPS)} & \textbf{Inference (SPS)} \\
    \midrule
    \multicolumn{4}{l}{\textit{Large Language Models}} \\
    \textbf{Peak-Detector (Ours)} & \textbf{$\sim$3 Billion} & \textbf{0.397} & \textbf{3.571} \\
    Claude-Sonnet-4.5 & N/A & N/A & 0.070 \\
    GPT-5 & N/A & N/A & 0.070 \\
    Gemini-2.5-pro & N/A & N/A & 0.070 \\
    \midrule
    \multicolumn{4}{l}{\textit{Deep Learning Models}} \\
    1D-UNet++ & 1.79 M & 3,414 & 12,379 \\
    CNN-SWT & 0.69 M & 2,904 & 14,059 \\
    FR-Net & 6.69 M & 1,061 & 4,037 \\
    \midrule
    \multicolumn{4}{l}{\textit{Signal Processing Algorithms}} \\
    Pino & N/A & N/A & 2,110 \\
    Choi & N/A & N/A & 5,278 \\
    Bishop & N/A & N/A & 272 \\
    Elgendi & N/A & N/A & 578 \\
    Nabian & N/A & N/A & 624 \\
    Pan-Tompkins & N/A & N/A & 1,550 \\
    \midrule
    \multicolumn{4}{l}{\textit{Human Benchmark}} \\
    Physician (Manual) & N/A & N/A & 0.074~\cite{winters2022time} \\
    \bottomrule
 \end{tabular}}
\end{table}

Building on the scaling findings, a comprehensive complexity analysis was conducted to evaluate the computational viability of our model within expert review workflows. Table~\ref{tab:complexity} contrasts Peak-Detector ($\sim$3B parameters) against a spectrum of baselines by quantifying throughput in signal segments per second (SPS). The results reveal a distinct performance hierarchy: lightweight numerical algorithms, such as Pan-Tompkins, and specialized deep learning models like CNN-SWT achieve extremely high throughput ($>1,000$ SPS) by prioritizing processing speed over semantic depth. In contrast, general-purpose proprietary LLMs prove prohibitively slow ($\approx 0.07$ SPS) for large-scale clinical analysis. \textbf{Peak-Detector} occupies a critical middle ground; although its computational overhead results in a lower throughput (3.571 SPS) than pure signal processing methods, the Peak Representation strategy allows it to operate over $50\times$ faster than general-purpose LLMs. Furthermore, it outperforms the human benchmark of 0.074 SPS~\cite{winters2022time} by approximately $48\times$. While this model scale may limit real-time edge processing, the system remains highly efficient for asynchronous tasks—such as automated reporting and retrospective analysis—providing a unique combination of accuracy and explainability that lighter models cannot match.

\section{DISCUSSIONS AND LIMITATIONS}
\label{sec:discussion}

\subsection{Real-world Impact for Cardiovascular Metrics in Ubiquitous Sensing}

Peak detection is not an end goal in mobile and ubiquitous health; it is a prerequisite for reliable cardiovascular metrics such as IBI/HR/HRV and downstream inferences (e.g., stress monitoring, sleep assessment, arrhythmia screening, and longitudinal health tracking) as shown in Appendix~\ref{sec:implication}. In real-world deployments, the primary difficulty is not only average accuracy on curated benchmarks, but \emph{robust, auditable operation under non-ideal conditions}: heterogeneous sensors and modalities, motion artifacts and context-dependent noise, variable contact quality, intermittent missing data, and user/context shifts over time. Improved statistics are informative, but a complete picture for field use requires understanding how the system supports trustworthy metric extraction and practical deployment choices.

Peak-Detector advances field use through three system-relevant capabilities that directly support ubiquitous cardiovascular metric extraction. First, \textit{modality-agnostic generalization} reduces per-device calibration and engineering overhead when integrating new wearable/contactless sensors (ECG/PPG/BCG/BSG) into ubiquitous pipelines. 
Second, \textit{explainable, audit-ready outputs} provide structured rationales that enable rapid expert verification and error triage when signals degrade (e.g., motion/artifacts or arrhythmia), addressing silent failure modes that can otherwise propagate into downstream HR/HRV estimates. 
Third, its \textit{compute--accuracy trade-off} suggests a tiered deployment model in which lightweight front-end methods perform continuous low-power screening, while Peak-Detector is used selectively for flagged windows, uncertain segments, noisy intervals, or retrospective summaries that require higher-fidelity peak localization and interpretable justification.

Under this interpretation, we do not view Peak-Detector as a strict real-time, always-on model for edge execution on battery-constrained wearables. Rather, we consider its most appropriate role to be that of a selectively invoked second-stage analyzer within broader ubiquitous cardiovascular sensing workflows, particularly in settings where interpretability, auditability, and higher-confidence peak localization are important. The present results therefore support Peak-Detector primarily as a system-oriented algorithmic component for trustworthy cardiovascular metric extraction. At the same time, long-duration free-living validation and integration into fully end-to-end ubiquitous health systems remain important directions for future work.

\subsection{Generalizability Across Physiological Modalities}
\label{subsec:implications}

The development of a generalizable and explainable peak-detection framework addresses an important systems challenge in mobile and ubiquitous health. In many cardiovascular sensing pipelines, peak detection is not the final objective, but a prerequisite for deriving reliable intermediate measures such as inter-beat interval (IBI), heart rate (HR), and heart rate variability (HRV), which in turn support higher-level applications including rhythm monitoring, sleep-related analysis, stress assessment, and longitudinal health tracking. From this perspective, the practical value of \textit{Peak-Detector} lies less in any single downstream task than in its potential to provide a common, interpretable front end for trustworthy cardiovascular metric extraction across heterogeneous sensing modalities.

This role is particularly relevant because different sensing modalities have traditionally required separate peak-detection strategies tailored to their own signal morphology and noise characteristics. ECG, PPG, BCG, and BSG each expose distinct fiducial structures and are often deployed in different sensing contexts, which has historically led to fragmented algorithm design and repeated modality-specific tuning. Our results suggest that a modality-agnostic framework such as \textit{Peak-Detector} can reduce this fragmentation by supporting a shared analysis interface across heterogeneous signals, thereby lowering the engineering overhead associated with selecting, adapting, and calibrating separate algorithms for each device or modality.

Under this interpretation, the contribution of \textit{Peak-Detector} is not that it directly solves all downstream clinical or wellness tasks, but that it provides a more unified and auditable foundation for the upstream peak-analysis step on which such tasks depend. This property is especially valuable for ubiquitous sensing systems that must incorporate new sensors, operate across multiple modalities, or balance accuracy with interpretability in broader cardiovascular monitoring workflows.

\subsection{Explainability and Clinical Trust}
Many deep learning baselines~\cite{zuo2025tau,reiss2019deep,biswas2019cornet,schranz2024surrogate} for peak detection produce only discrete peak locations (or a probability map) without an explicit, human-readable justification of \emph{why} those peaks are selected.
While post-hoc interpretation tools (e.g., saliency maps / attention visualizations) can highlight influential regions, they typically do not provide structured, physiologically grounded reasoning that supports rapid expert auditing.

In contrast, Peak-Detector is designed to output not only peak timestamps but also concise rationales aligned with domain criteria. Concretely, Peak-Detector summarizes evidence across four complementary dimensions: morphology (shape consistency), temporal consistency (interval constraints), amplitude dominance (relative prominence), and waveform context (local neighborhood patterns) as shown in Appendix~\ref{app: visualization}. These cues align with how practitioners reason about fiducial points, making model outputs easier to audit and validate than predictions that provide no rationale.

Under this framing, the value of explainability is not merely descriptive, but operational: structured rationales can assist audit, failure analysis, and rapid verification when signal quality degrades or when candidate peaks are ambiguous. We further support this process through the visualization-assisted audit interface described in Appendix~I, which provides a practical mechanism for human-in-the-loop inspection in the evaluated settings. Taken together, these design choices position explainability in \textit{Peak-Detector} as a tool for improving transparency and trustworthiness in upstream peak analysis, rather than as a post-hoc add-on to otherwise opaque predictions.

\subsection{LLMs Hallucination}
Large Language Model (LLM) hallucination \cite{Huang_2025} describes the generation of content that remains semantically plausible while being factually ungrounded. Within the domain of physiological signal analysis, hallucination risks are primarily localized to the Peak-Explanation text and typically manifest in two forms: (a) Objective Inconsistencies, where generated timestamps, amplitudes, or intervals contradict the input metadata; and (b) Semantic Inaccuracies, characterized by logically unsupported diagnostic rationale. To quantify these risks, we conducted a rigorous verification process on the Peak-Explanation dataset (see Appendix~\ref{sec:hallucination}). Our empirical results demonstrate zero factual errors and a negligible incidence of ambiguous expressions, suggesting that the dataset is highly resistant to hallucinatory outputs.

\subsection{The Signal-to-Sequence Paradigm}
The success of our approach is fundamentally rooted in the \textit{Peak Representation}, which serves as a powerful abstraction layer between raw numerical data and the LLM. While this transformation is intuitive for peak detection—where salient information is locally concentrated—its implications extend far beyond this specific task. This signal-to-sequence paradigm represents a transformative approach for a wide range of time-series analysis tasks where information is sparse or event-driven, including anomaly detection, EEG seizure detection, and sleep stage classification. By converting irregular time-series data into a symbolic, token-based format, this method naturally aligns with the architectural strengths of LLMs, potentially unlocking their sophisticated reasoning capabilities for problems previously dominated by specialized numerical models. Furthermore, this strategy effectively distills critical temporal information, offering a robust alternative to traditional frequency-domain analyses and mitigating the impact of missing data in volatile time-series environments~\cite{chen2023contiformer,zhang2025diffode}. From a ubiquitous computing perspective, this event-centric representation is well matched to resource-constrained, event-driven sensing pipelines, where only sparse salient events need to be transmitted, reviewed, or fused across devices.

{
\subsection{Limitations and Future Directions}
\label{subsec:limitations_future}

\textbf{Real-time and on-device constraints.}
While Peak-Detector-3B is practical for offline/batch analysis on workstation/server hardware (Sec.~5.4; Table~2), it remains slower than lightweight CNN baselines (e.g., FR-Net) and may be unsuitable for strict real-time, always-on, on-device processing in unconstrained field settings. Appendix~H and ~K quantify the memory--accuracy and device feasibility trade-offs; developing more resource-aware variants that preserve explanation quality remains an important direction. Our intended use case is therefore not strict always-on edge execution, but selective second-stage analysis for windows that have already been surfaced by lightweight front-end detectors or downstream review pipelines.

\textbf{Scope of target peaks.}
This study focuses on the dominant fiducial point per modality (ECG R, PPG systolic, BCG/BSG J), which is sufficient for the primary metrics emphasized here (IBI/HR/HRV). Extending the framework to multi-fiducial detection (e.g., ECG P/T waves) requires datasets with consistent, high-quality multi-wave annotations and is left for future work.

\textbf{Generalization beyond the benchmark.}
Although we evaluate across multiple datasets and report cross-dataset analyses (Appendix~J), fully in-the-wild deployments introduce additional sources of domain shift (device diversity, long-term drift, context transitions). Future work will include longitudinal field evaluations and end-to-end integration into ubiquitous health workflows. Although our robustness analyses under controlled noise in Appendix~\ref{sec:Robustness aginst noise} provide supporting evidence, we do not claim that they substitute for fully free-living validation, where motion artifacts, posture transitions, contact variability, and long-term domain drift remain open challenges.

\textbf{Explanation quality at scale.}
We provide verification analyses for teacher-generated rationales (Appendix~I). While objective consistency checks are applied at scale, semantic quality is assessed via sampling due to the cost of exhaustive manual review. Improving automatic semantic validation and studying how explanations impact expert confidence and decision-making in real workflows are promising directions.

}
\section{CONCLUSION}
\label{sec:conclusion}

Peak-Detector presents a specialized framework that reformulates cardiac peak detection as a language-guided reasoning task. By leveraging Large Language Models, it moves beyond rigid handcrafted heuristics and opaque deep learning pipelines toward a more transparent and auditable approach to peak analysis. The framework combines a compact Peak Representation, a two-stage instruction-tuning strategy, and a dedicated Peak-Explanation Dataset to jointly support accurate peak localization and structured self-explanations. Across seven benchmarks spanning ECG, PPG, BCG, and BSG, the results show that Peak-Detector achieves strong cross-modal performance while offering interpretable rationales that can support verification and error analysis.

Rather than positioning Peak-Detector as a strict real-time, always-on model for battery-constrained wearable deployment, we view its current strength as a selectively invoked, higher-fidelity second-stage analyzer within broader ubiquitous cardiovascular sensing workflows. In this role, Peak-Detector is particularly well suited for flagged windows, uncertain segments, and retrospective summaries where explainability, auditability, and higher-confidence peak localization are especially important. Overall, this work provides a system-oriented foundation for trustworthy cardiovascular metric extraction across heterogeneous sensing modalities, while long-duration free-living validation and more resource-aware edge deployment remain important directions for future work.

\section{ACKNOWLEDGMENT}
Portions of the text in this manuscript were refined using OpenAI ChatGPT to improve clarity, phrasing, and grammar. The tool was not used to generate ideas, analyses, figures, tables, or any scientific content. The authors take full responsibility for the integrity and accuracy of all content in this manuscript.
\bibliographystyle{ACM-Reference-Format}
\bibliography{ref}
\newpage
\appendix
\appendix
\section{DATASET DETAILS}
\label{appendix:Dataset details}

To thoroughly evaluate the cross-modal and explanatory capabilities of Peak-Detector, we conducted experiments on a comprehensive suite of six publicly available physiological signal datasets. This collection includes two datasets each for Electrocardiogram (ECG), Photoplethysmogram (PPG), and Ballistocardiogram (BCG) signals, ensuring broad coverage of signal modalities and varying data characteristics. A summary of these datasets is provided below:

\begin{itemize}
    \item \textbf{MIT-BIH Arrhythmia Database}~\cite{moody1992bih}: This seminal dataset comprises 48 half-hour, two-channel ambulatory ECG recordings, acquired from 47 subjects between 1975 and 1979 by the BIH Arrhythmia Laboratory. ECG signals are sampled at 360\,Hz. The dataset includes human annotations for various arrhythmia types. For our purposes, both normal and arrhythmic QRS complexes are considered as target R-peaks.

    \item \textbf{Incart Arrhythmia Database}~\citep{tihonenko2007st}: This ECG dataset was collected from 32 patients (17 men, 15 women; aged 18-80, mean age: 58 years) undergoing tests for coronary artery disease. Signals are sampled at 257\,Hz. Similar to MIT-BIH, the dataset provides annotations for arrhythmia types, and we include all detected QRS complexes (normal and arrhythmic) as target R-peaks.

    \item \textbf{BIDMC Database}~\cite{pimentel2016toward}: This dataset features synchronized ECG and PPG signals collected from 53 Intensive Care Unit (ICU) patients (32 females, 21 males; median age: 64.8 years). Both signal modalities are sampled at 125\,Hz. Notably, the BIDMC dataset lacks pre-existing systolic peak annotations for PPG signals. \textit{Therefore, we employed a semi-automatic annotation strategy: initial systolic peak detection was performed using Elgendi's algorithm~\cite{elgendi2013systolic}, followed by meticulous manual correction by annotators based on signal morphology and temporal alignment with corresponding ECG R-peaks.}

    \item \textbf{Capnobase Database}~\cite{karlen2013multiparameter}: This dataset contains PPG signals recorded from 29 pediatric patients (median age: 8.7 years) and 13 adult patients (median age: 52.4 years) during medication examinations. PPG signals were acquired in transmission mode via fingertip pulse oximeters at a sampling frequency of 300\,Hz. The dataset consists of 42 recordings, each approximately 8 minutes in duration. Crucially, this dataset provides high-quality systolic peak annotations that have been validated by medical experts.

    \item \textbf{Kansas Database}~\cite{carlson2020bed}: Developed by Kansas State University, this open-source dataset offers synchronized multimodal physiological signals, including BCG, ECG, PPG, and Arterial Blood Pressure (ABP) waveforms. BCG signals were captured using four electromechanical film (EMFi) sensors placed under the mattress and four load cells under the bed frame. Data were collected from 40 subjects (17 males, ages 18-65 years), with four subjects presenting cardiovascular-related conditions. The raw BCG signal is sampled at 1000\,Hz; for our analysis, it was downsampled to 100\,Hz to optimize for computational efficiency while retaining sufficient peak information within each segment. \textit{As with BIDMC, this dataset lacked pre-provided BCG J-peak labels, necessitating the adoption of a similar semi-automatic annotation procedure involving algorithmic detection using Elgendi's algorithm and annotators review.}

    \item \textbf{BCG Arrhythmia Database}~\cite{zhan2025multi}: This dataset includes BCG recordings from 85 participants, encompassing individuals with sinus rhythm, heart failure (HF), and various cardiac arrhythmias such as Atrial Fibrillation (AF), Premature Ventricular Contractions (PVCs), and Premature Atrial Contractions (PACs). Signals are sampled at 100\,Hz. \textit{BCG J-peaks in this database were labeled semi-automatically, combining Elgendi’s algorithmic detection with subsequent manual verification by annotators.}

    \item \textbf{BSG ICU Database}: 
    This clinical dataset was acquired from Hospital under an IRB-approved protocol with informed consent. Ethical Approval and Data Collection is shown in Appendix.~\ref{sec:ethical}.  It comprises 1,120 hours of continuous BSG and ECG recordings from 44 ICU patients. The cohort encompasses a wide demographic range, with patient ages spanning from 6 to 86 years (pediatric to geriatric). Both signals were sampled at a frequency of 100\,Hz. Following the quality control protocols established in~\cite{song2023engagement}, we retained a curated subset of 6,534 segments from 15 subjects for subsequent analysis.
    \textit{The BCG J-peaks in this database were annotated using a rigorous semi-automatic protocol. This process involved initial algorithmic detection using Nabian's algorithm~\cite{nabian2018open} to generate candidate peaks, followed by meticulous manual verification and correction by annotators.}
\end{itemize}

\textbf{Human Annotation and Data Statistics.} We use official peak annotations when available; otherwise we create human-verified peak labels. The ground truth for the BIDMC, Kansas, BCG Arrhythmia, and BSG ICU datasets was established using a rigorous semi-automated protocol, comprising initial algorithmic detection followed by meticulous manual verification. We invited three annotators with prior experience in peak annotation,  and trained them using a standardized guideline that defines modality-specific fiducial characteristics (ECG R, PPG systolic, BCG/BSG J). To ensure label integrity, we enforced a strict consensus mechanism: only peak positions independently identified by all three annotators were retained. Any samples exhibiting more than two disagreement peaks were excluded from the dataset. The resulting dataset statistics are presented in Table~\ref{tab:dataset_statistics}. To demonstrate the physiological diversity of the corpus, we report the total data volume (number of segments and peaks) alongside key hemodynamic metrics: Inter-Beat Interval (IBI), Heart Rate (HR), and Heart Rate Variability (HRV). This rigorous annotation protocol ensures that the ground truth labels achieve the highest possible fidelity. Although the preliminary annotation algorithm may be identical to the baseline, the evaluation phase maintains experimental integrity; unlike the manual annotation process, the baseline is assessed in a fully autonomous manner to ensure an equitable comparison across all models.

\textbf{Inter-Rater Agreement Analysis:} To evaluate the reliability of our ground truth labels, we conducted a rigorous inter-rater agreement analysis, as detailed in Table~\ref{tab:inter_rater_agreement}. We define \textit{Initial Agreement} as the percentage of peak candidates unanimously identified by all three independent annotators prior to any reconciliation or consensus-building phases. To further quantify the statistical reliability of these annotations, we utilized Cohen’s Kappa ($\kappa$)~\cite{sun2011meta} to assess the level of consensus while accounting for the probability of agreement occurring by chance.

As evidenced by the results, the PPG-based BIDMC dataset exhibits the highest degree of agreement. This is primarily attributable to the high signal-to-noise ratio and distinct morphological clarity of PPG signals, which allow for consistent peak localization. Among the mechanical modalities, the Kansas BCG dataset demonstrates superior agreement compared to other BCG and BSG signals, likely due to the standardized and controlled laboratory conditions under which the data were acquired. 

Conversely, the BCG Arrhythmia and BSG ICU datasets show relatively lower agreement metrics. This trend reflects the inherent challenges of annotating signals captured in clinical environments, where pathological arrhythmias, baseline wander, and varied patient movements introduce significant morphological ambiguity.

\begin{table}[htbp]
    \centering
    \caption{Statistical summary of the utilized datasets including subject counts. Values for IBI, HR, and HRV are presented as Mean $\pm$ Standard Deviation.}
    \label{tab:dataset_statistics}
    \resizebox{0.7\textwidth}{!}{
    \begin{tabular}{lcccccc}
        \toprule
        \textbf{Dataset} & \textbf{\# Subjects} & \textbf{\# Segments} & \textbf{\# Peaks} & \textbf{IBI (ms)} & \textbf{HR (bpm)} & \textbf{HRV (ms)} \\
        \midrule
        \textbf{MIT-BIH} & 48 & 29\,837 & 104\,878 & $764.1 \pm 200.8$ & $77.3 \pm 17.8$ & $68.8 \pm 106.0$ \\
        \midrule
        \textbf{Incart} & 75 & 34\,650 & 175\,360 & $749.0 \pm 233.7$ & $78.9 \pm 20.5$ & $99.6 \pm 129.0$ \\
        \midrule
        \textbf{BIDMC} & 50 & 3\,178 & 37\,581 & $671.9 \pm 122.0$ & $89.1 \pm 13.3$ & $44.6 \pm 64.4$ \\
        \midrule
        \textbf{Capnobase} & 42 & 6\,048 & 27\,936 & $707.6 \pm 174.0$ & $83.5 \pm 21.1$ & $19.3 \pm 36.3$ \\
        \midrule
        \textbf{Kansas} & 40 & 1\,654 & 16\,158 & $887.9 \pm 233.6$ & $67.1 \pm 12.0$ & $131.8 \pm 138.2$ \\
        \midrule
        \textbf{BCG Arrhythmia} & 85 & 238 & 2\,750 & $693.6 \pm 149.4$ & $86.3 \pm 13.8$ & $71.4 \pm 85.2$ \\
        \midrule
        \textbf{BSG ICU} & 15 & 6\,534 & 19\,862 & $640.0 \pm 218.2$ & $93.3 \pm 18.9$ & $180.1 \pm 83.3$ \\
        \bottomrule
    \end{tabular}%
    }
\end{table}

\begin{table}[htbp]
\centering
\caption{Inter-Rater Agreement Metrics for Semi-Annotated Datasets}
\label{tab:inter_rater_agreement}
\resizebox{0.8\columnwidth}{!}{
\begin{tabular}{lcccc}
\toprule
\textbf{Dataset (Modality)} & \textbf{Total Segments} & \textbf{Removed Segments} & \textbf{Initial Agreement (\%)} & \textbf{Cohen's Kappa ($\kappa$)} \\ \midrule
BIDMC (PPG)           & 3,178 & 5   & 99.4\% & 0.99 \\
Kansas (BCG)          & 1,654 & 12  & 97.2\% & 0.95 \\
BCG Arrhythmia (BCG)  & 238   & 4   & 94.8\% & 0.91 \\
BSG ICU (BSG)         & 6,534 & 132 & 91.2\% & 0.84 \\ \bottomrule
\end{tabular}}
\end{table}

\section{BASELINE DETAILS}
\label{appendix:Baseline details}
\subsection{Signal-Processing Baselines}
These methods leverage domain-specific heuristics and mathematical transformations tailored to individual signal modalities. To simplify the replication, we adopt public implementation from neurokit2~\cite{makowski2021neurokit2}.

\begin{itemize}
    \item \textbf{Pan-Tompkins}~\cite{fariha2020analysis}: A widely adopted algorithm for ECG R-peak detection, it utilizes a series of signal processing steps including filtering, differentiation, squaring, and moving-window integration to extract slope and energy information, followed by adaptive thresholding to identify R-peaks.
    \item \textbf{Nabian}~\cite{nabian2018open}: This ECG R-peak detection method employs a sliding window technique. It identifies the maximum amplitude within each window, designating it as a potential R-peak, and subsequently refines these detections.
    \item \textbf{Elgendi}~\cite{elgendi2012analysis}: Designed for PPG systolic peak detection, this algorithm defines regions of interest by calculating two moving averages with distinct window sizes. Peaks are then identified as local maxima within these predefined regions.
    \item \textbf{Bishop}~\cite{bishop2018multi}: A multi-scale approach for PPG, it constructs a Local Maxima Scalogram (LMS) by analyzing the signal at various levels of smoothing. Peaks are then robustly identified by detecting common local maxima across these different scales.
    \item \textbf{Pino}~\cite{pino2017bcg}: This method focuses on BCG J-peak detection, employing techniques such as wavelet transformations, template matching, and signal envelopes to enhance and isolate the characteristic J-peak morphology within noisy BCG signals.
    \item \textbf{Choi}~\cite{choi2009slow}: This BCG J-peak detection algorithm segments the signal based on an estimated mean heartbeat interval. It then identifies local maxima within each segment and eliminates false positives through analysis of peak-to-peak intervals, enhancing robustness to noise.
\end{itemize}

\subsection{Deep Learning Baselines}
These models represent advanced data-driven approaches, designed to learn complex features directly from the physiological signals.

\begin{itemize}
    \item \textbf{CNN-SWT}~\cite{yun2022robust}: Originally proposed for robust ECG R-peak detection, this model is a Convolutional Neural Network (CNN) architecture that leverages Stationary Wavelet Transform (SWT) coefficients as input, combined with separable convolutions to enhance feature extraction.
    \item \textbf{1D-UNet++}~\cite{zhou20211d}: An extension of the U-Net architecture, this model utilizes nested dense skip connections between its encoder and decoder paths. Although initially designed for medical image segmentation, its 1D adaptation has demonstrated strong performance in time-series analysis, including BCG J-peak detection.
    \item \textbf{FR-Net}~\cite{chen2023sample}: This is a specialized CNN-Transformer based network primarily developed for fetal R-peak detection in challenging ECG signals. Its architecture combines convolutional layers for local feature extraction with transformer blocks to capture long-range dependencies, making it suitable for complex peak detection tasks.
\end{itemize}

\section{DETAILED EXPERIMENT PROTOCOL}
\label{app:experimental_protocol}

\textbf{Supervised Fine-Tuning Configuration}
Peak-Detector was fine-tuned using Qwen2.5-3B-Instruct as the base model with a maximum sequence length of 2048 tokens and a dropout rate of 0.1. The model was trained for 5 epoachs using the AdamW optimizer with hyperparameters $\beta_1=0.9$, $\beta_2=0.999$, and $\epsilon=10^{-8}$. The learning rate was set to $2 \times 10^{-5}$ with a cosine decay schedule following 500 warmup steps. To prevent overfitting, we applied weight decay of 0.01 and gradient clipping with a maximum norm of 1.0. The training process utilized a per-device batch size of 32 with 4 gradient accumulation steps, yielding an effective batch size of 128. Mixed precision training with BF16 was employed to improve computational efficiency. The model was trained on 4$\times$ NVIDIA A6000 GPUs (80GB each).

\textbf{GRPO Configuration}
The model was further optimized using Group Relative Policy Optimization (GRPO). The actor model was initialized from a supervised fine-tuned checkpoint with a learning rate of $1 \times 10^{-6}$. Maximum prompt and response lengths were set to 4000 and 2000 tokens, respectively. To control policy divergence, KL divergence loss was enabled with a coefficient of 0.001 using the low-variance formulation. The rollout generation utilized vLLM with tensor model parallelism and generated 16 samples per prompt. Training was conducted on 4 GPUs for a single epoch with a training batch size of 24, PPO mini-batch size of 12, and micro-batch size of 2 per GPU.

\textbf{Code Availability.}
The source code for the Peak-Detector framework and the generated Peak-Explanation Dataset is available on \href{https://github.com/jimmylihui/Peak-Detector}{GitHub}.

\section{ETHICAL APPROVAL AND DATA COLLECTION FOR BSG ICU DATASET}
\label{sec:ethical}
The data collection for this study was conducted in the hospital ICU under an Institutional Review Board (IRB)-approved protocol. The patient consent process adhered strictly to ethical guidelines: data from patients, or their legal representatives, who declined consent—either during their ICU stay or after recovery—were excluded from the analysis.

The sensing devices were permanently installed beneath the hospital beds, seamlessly integrated as part of the bed infrastructure. The data acquisition process was entirely passive (i.e., involving no emission of active radiation) and contactless (i.e., without any physical contact with patients). As such, the system posed no risk to human subjects or interference with existing medical equipment. Importantly, the installation and operation of these devices did not disrupt standard ICU monitoring or treatment procedures, nor did they require any intervention from patients or clinical staff during routine care.

To minimize any potential disruption, device setup was meticulously planned and executed. Prior to installation, all devices were fully configured and validated for network connectivity and operational reliability. Installation was performed under the supervision of hospital personnel, ensuring a rapid, safe, and non-intrusive process. Once deployed, the devices operated autonomously, requiring no further human intervention. To further reduce maintenance demands within the ICU, devices were powered by a direct connection to the hospital mains, obviating the need for battery changes. In addition, the system software was engineered with robust fault-tolerance features, including automatic recovery from network or system errors, to maximize operational stability.

All data collection activities were performed solely by hospital staff. Data science researchers had access exclusively to de-identified data, as provided by the hospital. This strict separation ensured the protection of patient privacy and eliminated the risk of any information leakage.

\section{CONTROLLED ADDITIVE-NOISE STRESS TEST}
\label{sec:Robustness aginst noise}

\begin{figure*}[t]
     \centering
     \begin{subfigure}[b]{0.3\textwidth}
         \centering
         \includegraphics[width=\textwidth]{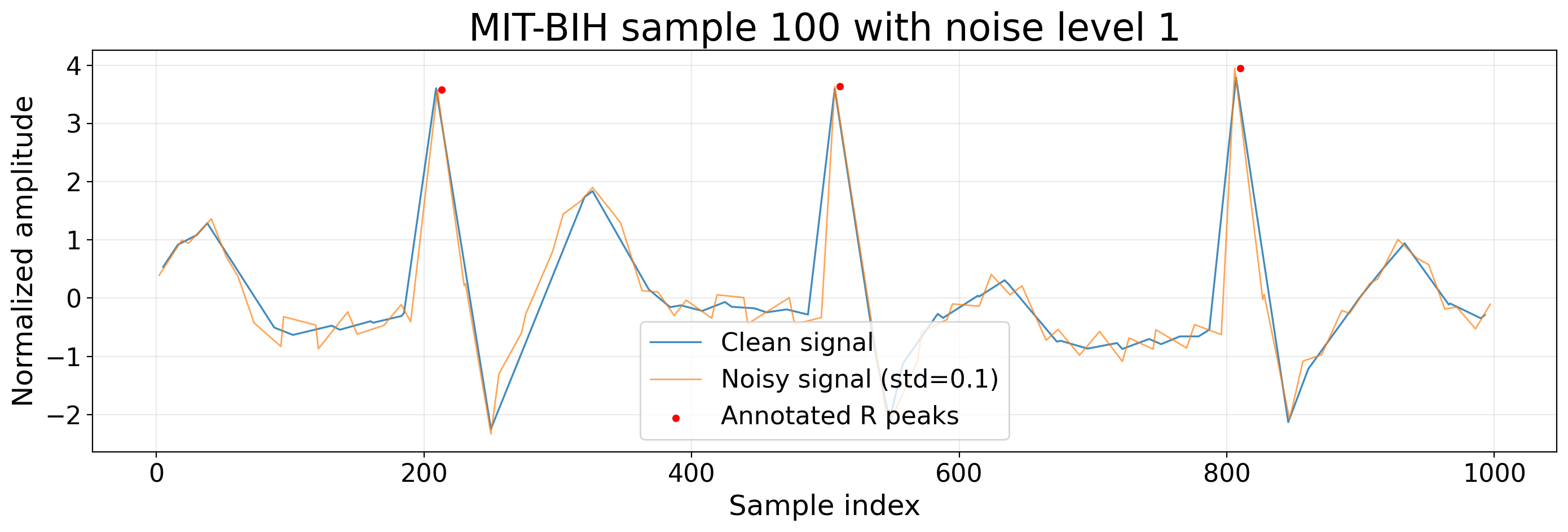}
         \caption{Noise level $\sigma=0.1$}
         \label{fig:sub1}
     \end{subfigure}
     \hfill
     \begin{subfigure}[b]{0.3\textwidth}
         \centering
         \includegraphics[width=\textwidth]{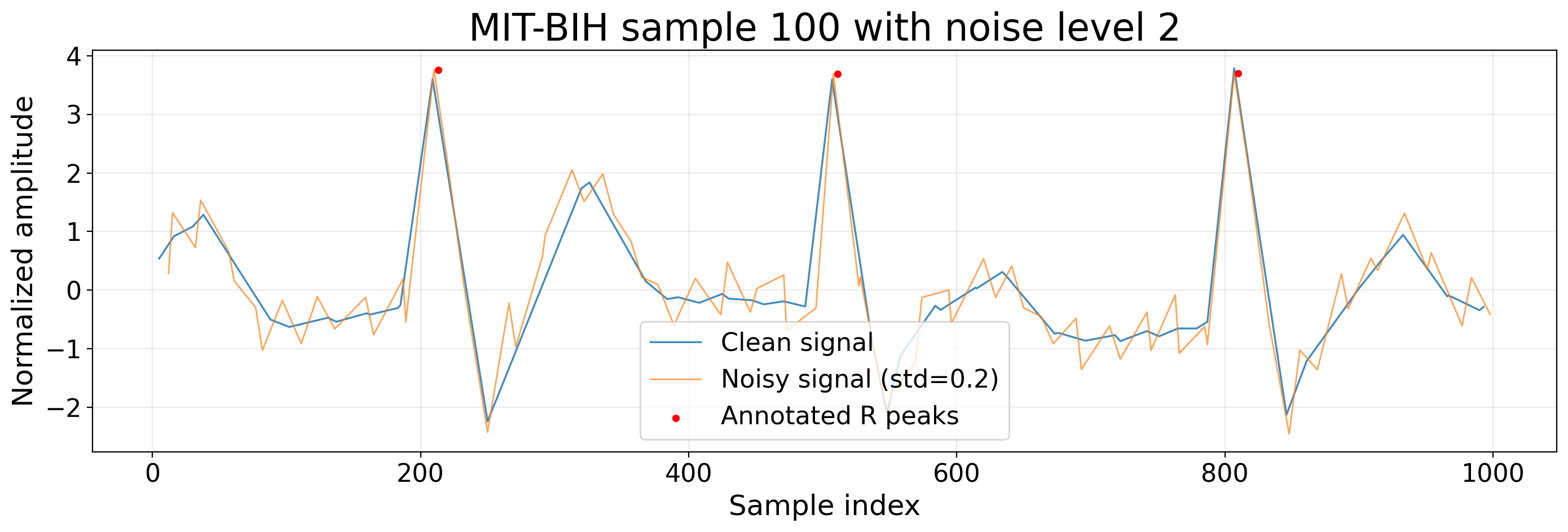}
         \caption{Noise level $\sigma=0.2$}
         \label{fig:sub2}
     \end{subfigure}
     \hfill
     \begin{subfigure}[b]{0.3\textwidth}
         \centering
         \includegraphics[width=\textwidth]{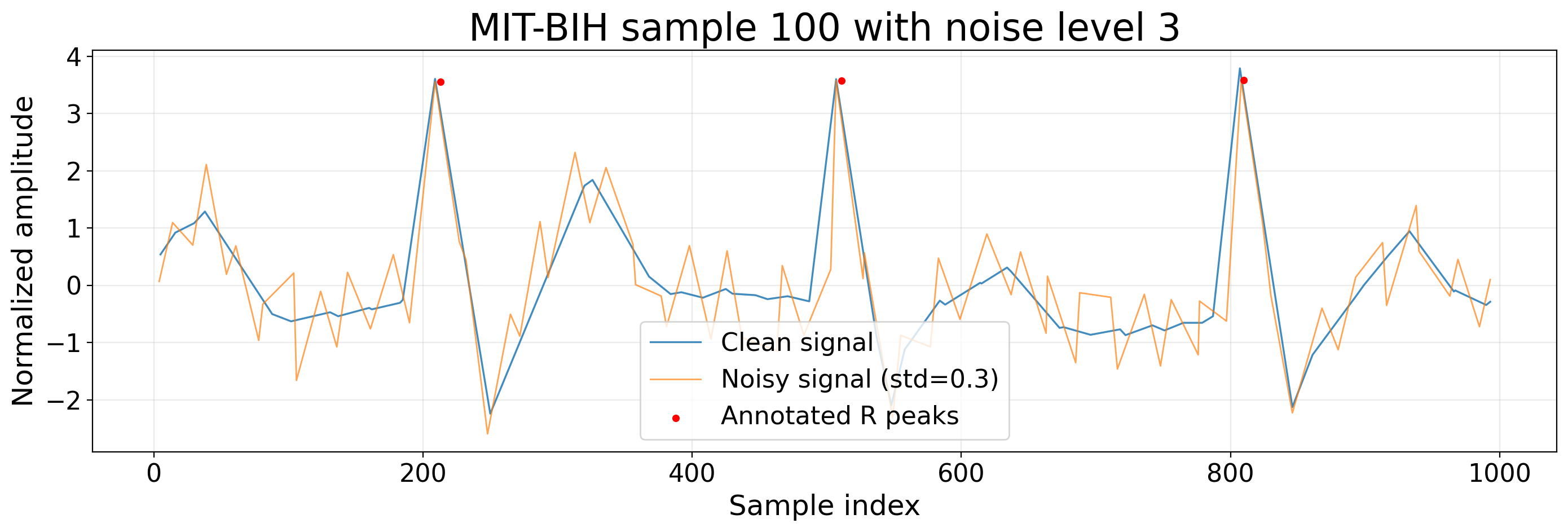}
         \caption{Noise level $\sigma=0.3$}
         \label{fig:sub3}
     \end{subfigure}

     \par\medskip

     \begin{subfigure}[b]{0.3\textwidth}
         \centering
         \includegraphics[width=\textwidth]{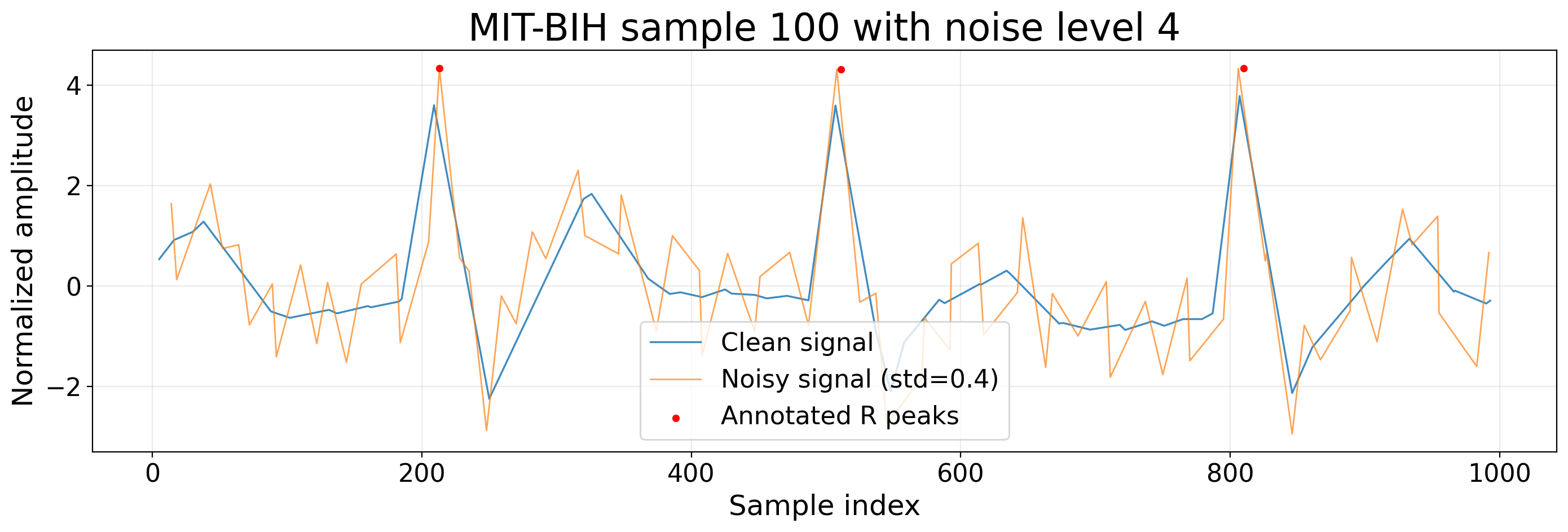}
         \caption{Noise level $\sigma=0.4$}
         \label{fig:sub4}
     \end{subfigure}
     \hspace{0.5cm}
     \begin{subfigure}[b]{0.3\textwidth}
         \centering
         \includegraphics[width=\textwidth]{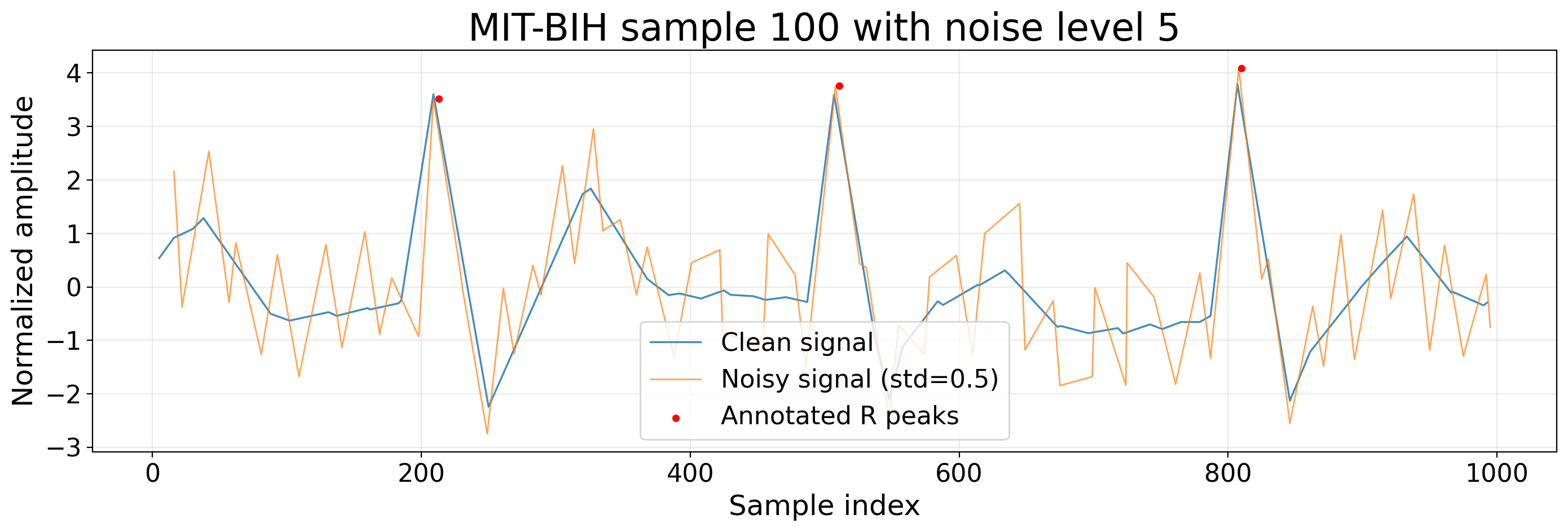}
         \caption{Noise level $\sigma=0.5$}
         \label{fig:sub5}
     \end{subfigure}

     \caption{Visualization of signal quality under different additive white Gaussian noise levels. Increasing noise progressively distorts the waveform and makes peak detection more challenging.}
     \Description{Five ECG signal plots showing progressively stronger additive Gaussian noise levels.}
     \label{fig:noise}
 \end{figure*}

To partially probe robustness to noise that may arise in less controlled settings, we conducted
a controlled additive-noise stress test by adding zero-mean white Gaussian noise to the input signal with standard deviations of 0.1, 0.2, 0.3, 0.4, and 0.5. The quantitative results are reported in Table~\ref{tab:noise_performance_std}, and representative visualizations are shown in Fig.~\ref{fig:noise}. We include this experiment as a controlled stress test of noise tolerance, rather than as a substitute for fully free-living evaluation.

As the perturbation level increases, performance degrades gradually and monotonically across all evaluation metrics. For heart-rate estimation, the MAE increases from $0.86 \pm 0.08$ BPM on clean signals to $2.99 \pm 0.24$ BPM at the highest noise level, while HR MAPE rises from $1.24 \pm 0.15\%$ to $4.34 \pm 0.34\%$. A similar trend is observed for HRV estimation: HRV MAE increases from $9.97 \pm 0.17$ ms to $25.95 \pm 1.55$ ms, and HRV MAPE rises from $61.10 \pm 0.91\%$ to $181.12 \pm 29.66\%$ as the perturbation becomes more severe.

Peak-detection performance also declines with increasing noise, but remains comparatively stable overall. Specifically, the F1 score decreases from $0.9729 \pm 0.0021$ under clean conditions to $0.9343 \pm 0.0011$ at the highest noise level. Precision drops from $0.9710 \pm 0.0025$ to $0.9239 \pm 0.0012$, and recall decreases from $0.9749 \pm 0.0023$ to $0.9448 \pm 0.0012$. These results suggest that the detector tolerates moderate additive perturbations reasonably well for beat detection and heart-rate estimation, whereas HRV-related metrics are more sensitive as the corruption level increases.

Overall, this analysis provides supporting evidence that the proposed model remains functional under controlled additive-noise perturbations, particularly for peak detection and HR estimation. At the same time, we do not interpret this experiment as a replacement for fully in-the-wild validation, where motion artifacts, posture transitions, contact variability, sensor displacement, and longer-term domain shift may introduce more complex failure modes than additive Gaussian noise alone.

\begin{table*}[t]
\centering
\caption{Performance of Peak-Detector under controlled additive Gaussian noise perturbations.}
\label{tab:noise_performance_std}
\resizebox{\textwidth}{!}{%
\begin{tabular}{@{}cccccccc@{}}
\toprule
\textbf{Noise Level} & \textbf{Heart MAE (BPM)} & \textbf{HRV MAE (ms)} & \textbf{HR MAPE (\%)} & \textbf{HRV MAPE (\%)} & \textbf{F1} & \textbf{Precision} & \textbf{Recall} \\ \midrule
\textbf{Clean} & \textbf{0.86 $\pm$ 0.08} & \textbf{9.97 $\pm$ 0.17} & \textbf{1.24 $\pm$ 0.15} & \textbf{61.10 $\pm$ 0.91} & \textbf{0.9729 $\pm$ 0.0021} & \textbf{0.9710 $\pm$ 0.0025} & \textbf{0.9749 $\pm$ 0.0023} \\ \midrule
1 & 0.93 $\pm$ 0.09 & 10.74 $\pm$ 0.47 & 1.32 $\pm$ 0.12 & 73.10 $\pm$ 7.10  & 0.9722 $\pm$ 0.0013 & 0.9701 $\pm$ 0.0006 & 0.9742 $\pm$ 0.0021 \\
2 & 1.05 $\pm$ 0.13 & 11.86 $\pm$ 1.01 & 1.48 $\pm$ 0.18 & 83.98 $\pm$ 14.15 & 0.9686 $\pm$ 0.0015 & 0.9668 $\pm$ 0.0011 & 0.9703 $\pm$ 0.0021 \\
3 & 1.41 $\pm$ 0.16 & 15.24 $\pm$ 1.31 & 2.02 $\pm$ 0.24 & 95.60 $\pm$ 10.39 & 0.9616 $\pm$ 0.0021 & 0.9580 $\pm$ 0.0027 & 0.9652 $\pm$ 0.0018 \\
4 & 1.94 $\pm$ 0.13 & 18.57 $\pm$ 1.31 & 2.80 $\pm$ 0.16 & 128.50 $\pm$ 24.44 & 0.9498 $\pm$ 0.0026 & 0.9441 $\pm$ 0.0022 & 0.9555 $\pm$ 0.0030 \\
5 & 2.99 $\pm$ 0.24 & 25.95 $\pm$ 1.55 & 4.34 $\pm$ 0.34 & 181.12 $\pm$ 29.66 & 0.9343 $\pm$ 0.0011 & 0.9239 $\pm$ 0.0012 & 0.9448 $\pm$ 0.0012 \\ \bottomrule
\end{tabular}%
}
\end{table*}

\section{FAILURE CASE ANALYSIS}
\label{sec:failure case analysis}
\begin{figure}[htbp]
     \centering
     \begin{subfigure}[b]{0.49\textwidth}
         \centering
         \includegraphics[width=\textwidth]{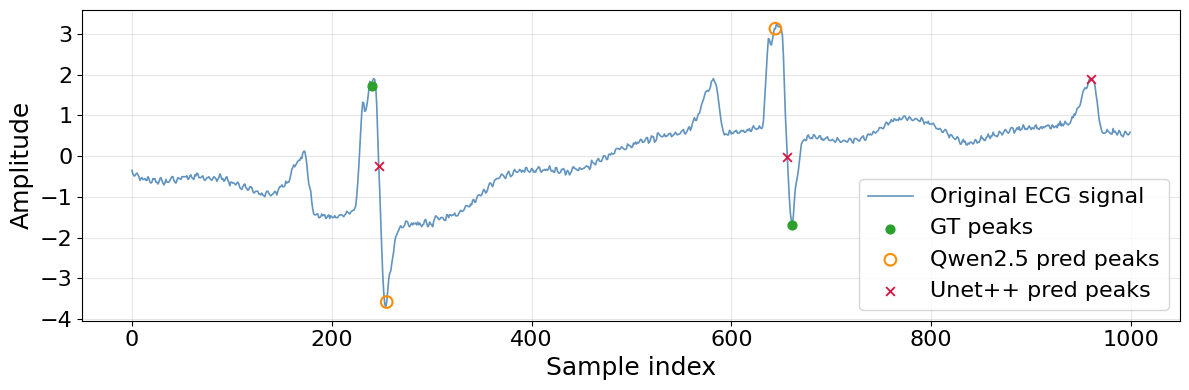}
         \caption{S-peak misidentification}
         \label{fig:first}
     \end{subfigure}
     \hfill
     \begin{subfigure}[b]{0.49\textwidth}
         \centering
         \includegraphics[width=\textwidth]{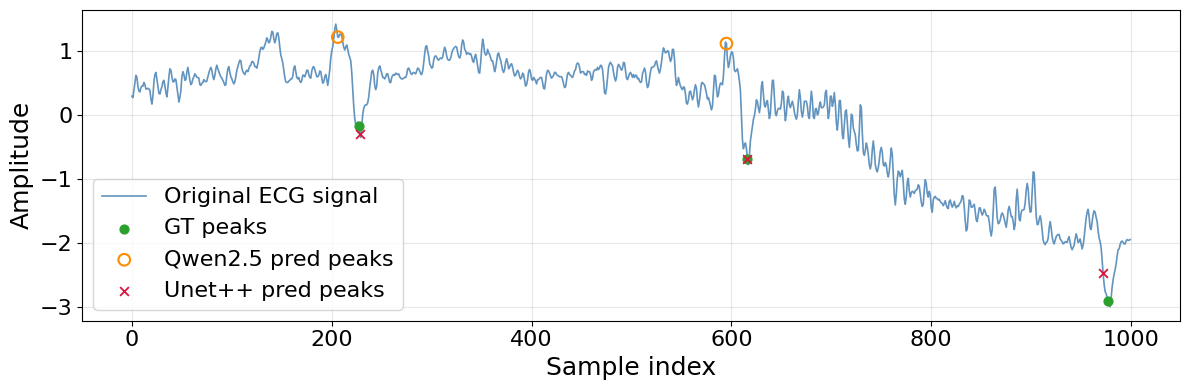}
         \caption{Motion artifact distortion}
         \label{fig:second}
     \end{subfigure}
     \hfill
        \caption{Analysis of failure cases in the MIT-BIH dataset: (a) morphological confusion between S-peaks and R-peaks, and (b) detection failures caused by extreme signal corruption.}
        \Description{Two failure-case plots showing S-peak misidentification and motion-artifact distortion.}
        \label{fig:errors}
\end{figure}

This section examines the representative failure modes of Peak-Detector on the MIT-BIH Arrhythmia Database, as illustrated in Fig. \ref{fig:errors}.

In the first segment (Fig. \ref{fig:first}), the model misidentifies the S-wave as an R-peak. This error is primarily attributed to the prominent negative amplitude of the S-wave, which exhibits a morphology similar to that of an inverted R-peak, leading to a false positive. In the subsequent segments (Fig. \ref{fig:second}), the model fails to localize the peaks accurately due to severe signal distortion. In these instances, the underlying QRS complex is obscured by high-frequency noise and baseline wander, resulting in a significantly reduced signal-to-noise ratio (SNR) that makes reliable peak selection more difficult.

Specialized ECG models such as the UNet++ can sometimes achieve smaller coordinate error on cleaner ECG signals because they are optimized for precise local QRS localization. However, proximity alone does not guarantee physiologically correct fiducial localization: a prediction may fall between adjacent landmarks (e.g., between the R and S peaks) while still appearing close to the annotation. In contrast, traditional peak detectors usually align more closely with anatomically meaningful morphological features. This observation suggests that future improvements should place greater emphasis on anatomically meaningful fiducial alignment, rather than optimizing only global localization error.

\section{ABLATION STUDY}
\label{sec:ablation_study}

To dissect the individual contributions of each component within the Peak-Detector framework, we conducted a systematic ablation study on the BCG Arrhythmia dataset, with the results detailed in Table~\ref{tab:ablation_study}. Our analysis begins with the baseline performance of a general-purpose, instruction-tuned LLM (\texttt{qwen2.5-3B-instruct}), which yields a remarkably low F1-score of 0.1690. This confirms that standard LLMs are ill-suited for this precise numerical inference task without specialized adaptation.

\begin{table}[htbp]
\centering
\normalsize
\caption{Ablation study results on the BCG Arrhythmia dataset, comparing precision, recall, F1-score, HR MAE (BPM), and HRV MAE under a 50\,ms matching tolerance. Best performance per block is highlighted in bold. ``-'' indicates no valid result could be obtained.}
\label{tab:ablation_study}
\resizebox{0.8\textwidth}{!}{%
\begin{tabular}{lccccc}
\toprule
& \multicolumn{5}{c}{BCG Arrhythmia Dataset} \\
\cmidrule(lr){2-6}
Method & Precision & Recall & F1-score & HR MAE (BPM) & HRV MAE (ms) \\
\midrule
\textbf{Peak-Detector (Full Framework)} & \textbf{0.9398}$\pm$\textbf{0.0290} & \textbf{0.9537}$\pm$\textbf{0.0393} & \textbf{0.9467}$\pm$\textbf{0.0338} & \textbf{1.43}$\pm$\textbf{0.95} & \textbf{36.85}$\pm$\textbf{22.71} \\
Peak-Detector (SFT-only) & $0.8995\pm0.0475$ & $0.8656\pm0.0427$ & $0.8822\pm0.0445$ & $2.69\pm0.89$ & $73.66\pm22.74$ \\
Peak-Detector (w/o Peak Representation) & $0.8079\pm0.0315$ & $0.8885\pm0.0542$ & $0.8460\pm0.0403$ & $7.87\pm2.23$ & $127.20\pm43.34$ \\
Peak-Detector (w/o Explanation) & $0.9531\pm0.0165$ & $0.9421\pm0.0186$ & $0.9476\pm0.0170$ & $1.87\pm0.99$ & $51.23\pm19.93$ \\
\midrule
\multicolumn{6}{l}{\textit{Comparison with General-Purpose LLMs}} \\
\midrule
qwen2.5-3B-instruct (Base Model) & $0.2684\pm0.0499$ & $0.1247\pm0.0140$ & $0.1690\pm0.0177$ & $32.06\pm8.03$ & $892.19\pm46.98$ \\
Claude-Sonnet-4.5 & $0.7256\pm0.0961$ & $0.9410\pm0.0243$ & $0.8169\pm0.0724$ & $23.26\pm6.72$ & $144.26\pm26.57$ \\
GPT-5 & $0.4908\pm0.0726$ & $0.9676\pm0.0270$ & $0.6490\pm0.0667$ & $66.32\pm11.76$ & $125.43\pm13.33$ \\
Gemini-2.5-Pro & $0.6672\pm0.0739$ & $0.9206\pm0.0918$ & $0.7723\pm0.0744$ & $28.92\pm7.03$ & $155.60\pm19.22$ \\
\bottomrule
\end{tabular}%
}
\end{table}

The introduction of our specialized \textbf{Peak Representation} is critical; removing this component (`w/o Peak Representation`) and relying on raw sequences limits the F1-score to 0.8460 and results in a significantly higher HR MAE of 7.87\,BPM, underscoring the difficulty LLMs face in processing raw numerical data efficiently. \textbf{Supervised Fine-Tuning (SFT)} on our dataset provides the most substantial performance leap, increasing the F1-score from the baseline 0.1690 to 0.8822. However, SFT alone is insufficient to reach expert-level performance, as evidenced by the performance gap between the `SFT-only` variant and the `Full Framework` (0.8822 vs. 0.9467 F1-score).

The \textbf{Full Framework}, which integrates explanation generation and optimization, achieves the best physiological consistency with an HR MAE of 1.43\,BPM. Notably, the ablation of the explanation task (`w/o Explanation`) yields a marginally higher F1-score of 0.9476 compared to the Full Framework's 0.9467. However, this comes at the cost of physiological accuracy, as the HR MAE degrades from 1.43 to 1.87\,BPM. This suggests that while the model can learn to detect peaks without generating explanations, compelling it to articulate reasoning (the explanation task) acts as a regularizer that enforces better alignment with the underlying cardiac rhythm, reducing heart rate estimation errors.

Finally, the specialized nature of our approach is highlighted when compared against powerful, general-purpose LLMs. Claude-Sonnet-4.5, Gemini-2.5-Pro, and GPT-5 achieve F1-scores of 0.8169, 0.7723, and 0.6490 respectively—significantly lower than our framework. Furthermore, their HR MAEs range from 23.26 to 66.32\,BPM, which is orders of magnitude higher than our model's 1.43\,BPM. This unequivocally demonstrates that our domain-specific approach—combining a novel signal representation with a targeted tuning strategy—is essential for achieving high-precision, physiologically valid performance.
\section{IMPACT OF RELATIVE TEMPORAL TOLERANCE}
\label{sec:relative_tolerance}

To mitigate the bias inherent in fixed temporal tolerances across varying heart rates, we additionally report a \textbf{rate-normalized} evaluation using an \underline{adaptive tolerance of $\pm 5\%$} of the local IBI , shown in Table~\ref{tab:results_relative}, and we apply the same scoring protocol to all methods for fair comparison.
Unlike fixed-window metrics, this adaptive criterion strictly tightens detection constraints during periods of tachycardia, thereby preventing the erroneous alignment of artifacts with true cardiac events.
All methods (our model and all baselines) are scored using the \emph{same} peak-matching protocol under each tolerance setting: one-to-one matching between predicted and ground-truth peaks, and TP is counted only when a predicted peak matches a ground-truth peak within the specified tolerance window. Therefore, the comparison is fair under both the fixed ($\pm 30$\,ms) and adaptive ($\pm 5\%$ IBI) metrics.
Under this adaptive criterion, \textbf{Peak-Detector} exhibited remarkable stability, showing minimal degradation compared to fixed-window results (e.g., maintaining an F1-score of 0.9698 on MIT-BIH). This resilience was characteristic of learning-based models (both Peak-Detector and deep-learning based), which generally maintained consistent performance across both tolerance paradigms, indicating the successful encoding of scale-invariant temporal features. In contrast, heuristic baselines such as Pan-Tompkins degrade substantially (e.g., F1 dropping from 0.4361 (fixed) to 0.2861 (relative) on MIT-BIH), underscoring the brittleness of static morphological thresholds when subjected to rate-modulated signal variability.

\begin{table}[t]
\centering
\caption{Peak Detection Performance Comparison across Signal Modalities and Baselines using a \textbf{Relative Temporal Tolerance} ($\pm 5\%$ of the local peak-to-peak interval). Lower values are better for MAE (HR(bpm), HRV(ms)) and MAPE, higher values are better for F1, Pre, and Rec. Best performance in each metric is \textbf{bold}, second best is \underline{underlined}. Standard deviations are shown with the same reduced formatting as Table~\ref{tab:results_combined}. Superscripts on Peak-Detector indicate significance against the second-best method: \textsuperscript{*} $p<0.05$, \textsuperscript{**} $p<0.01$. The $p$-value column is retained for reference.}
\label{tab:results_relative}
\resizebox{\textwidth}{!}{%
\begin{tabular}{cc|l|cccccccccccc}
\toprule
& & \textbf{Metric} & \textbf{Pan-T} & \textbf{Nabian} & \textbf{Elg} & \textbf{Bishop} & \textbf{Pino} & \textbf{Choi} & \textbf{CNN-SWT} & \textbf{1D-UNet++} & \textbf{FR-Net} & \textbf{Peak-Detector} & \textbf{$p$-value} \\
\midrule
\multirow{14}{*}{\rotatebox{90}{\scriptsize\textbf{ECG}}} & \multirow{7}{*}{\rotatebox{90}{\scriptsize\textbf{MIT-BIH}}}
 & HR MAE(bpm) & 1.06 & 4.23 & 3.81 & 79.75 & 3.59 & 7.01 & \val{1.19}{0.16} & \underline{\val{0.57}{0.07}} & \textbf{\val{0.48}{0.01}} & \valsig{0.86}{0.08}{**} & $2.23e-03$ \\
 &  & HRV MAE(ms) & 8.48 & 27.13 & 39.10 & 68.57 & 26.00 & 62.94 & \val{18.50}{0.57} & \underline{\val{6.50}{0.68}} & \textbf{\val{5.02}{0.43}} & \valsig{9.97}{0.17}{**} & $3.33e-05$ \\
 &  & HR MAPE(\%) & 1.48 & 5.44 & 5.63 & 105.98 & 4.45 & 9.61 & \val{1.48}{0.16} & \underline{\val{0.79}{0.07}} & \textbf{\val{0.59}{0.01}} & \valsig{1.24}{0.15}{**} & $3.14e-03$ \\
 &  & HRV MAPE(\%) & 33.09 & 43.66 & 157.56 & 268.89 & 78.28 & 240.54 & \val{75.08}{6.93} & \underline{\val{27.84}{4.71}} & \textbf{\val{15.10}{0.75}} & \valsig{61.10}{0.91}{**} & $5.68e-10$ \\
 &  & F1 & 0.966 & 0.781 & 0.141 & 0.542 & 0.893 & 0.717 & \val{0.980}{0.002} & \underline{\val{0.987}{0.001}} & \textbf{\val{0.992}{0.000}} & \valsig{0.970}{0.003}{**} & $7.10e-04$ \\
 &  & Pre & 0.971 & 0.942 & 0.144 & 0.425 & 0.917 & 0.745 & \val{0.984}{0.001} & \val{0.990}{0.001} & \val{0.995}{0.001} & \valsig{0.968}{0.003}{**} & $1.76e-04$ \\
 &  & Rec & 0.961 & 0.667 & 0.138 & 0.750 & 0.869 & 0.692 & \val{0.977}{0.003} & \val{0.984}{0.001} & \val{0.990}{0.001} & \valsig{0.972}{0.004}{**} & $1.38e-03$ \\
\cmidrule(l){2-14}
& \multirow{7}{*}{\rotatebox{90}{\scriptsize\textbf{Incart}}}
 & HR MAE(bpm) & 3.21 & 4.88 & 2.72 & 51.42 & 4.40 & 10.35 & \val{0.59}{0.04} & \textbf{\val{0.34}{0.06}} & \val{0.46}{0.11} & \val{0.42}{0.04} & $7.49e-02$ \\
 &  & HRV MAE(ms) & 45.22 & 47.70 & 40.00 & 73.26 & 46.16 & 87.01 & \val{8.82}{0.35} & \textbf{\val{5.70}{0.59}} & \underline{\val{5.90}{1.18}} & \valsig{18.79}{0.58}{**} & $6.64e-08$ \\
 &  & HR MAPE(\%) & 4.54 & 5.69 & 3.79 & 66.97 & 5.04 & 12.07 & \val{0.72}{0.05} & \textbf{\val{0.42}{0.08}} & \underline{\val{0.50}{0.09}} & \valsig{1.05}{0.08}{**} & $3.12e-05$ \\
 &  & HRV MAPE(\%) & 250.36 & 149.26 & 189.22 & 263.30 & 201.51 & 395.78 & \val{39.97}{1.80} & \textbf{\val{22.40}{2.19}} & \underline{\val{25.76}{8.49}} & \valsig{72.07}{6.41}{**} & $2.09e-04$ \\
 &  & F1 & 0.897 & 0.781 & 0.190 & 0.512 & 0.865 & 0.614 & \val{0.992}{0.001} & \textbf{\val{0.994}{0.001}} & \underline{\val{0.994}{0.001}} & \valsig{0.975}{0.001}{**} & $1.51e-06$ \\
 &  & Pre & 0.902 & 0.896 & 0.193 & 0.439 & 0.893 & 0.668 & \val{0.992}{0.001} & \val{0.996}{0.000} & \val{0.996}{0.001} & \valsig{0.974}{0.002}{**} & $6.83e-05$ \\
 &  & Rec & 0.893 & 0.692 & 0.187 & 0.617 & 0.838 & 0.568 & \val{0.993}{0.001} & \val{0.992}{0.001} & \val{0.991}{0.002} & \valsig{0.975}{0.001}{**} & $1.87e-07$ \\
\midrule
\multirow{14}{*}{\rotatebox{90}{\scriptsize\textbf{PPG}}} & \multirow{7}{*}{\rotatebox{90}{\scriptsize\textbf{BIDMC}}}
 & HR MAE(bpm) & 8.92 & 3.08 & 1.48 & 3.06 & 3.21 & 5.29 & \underline{\val{0.71}{0.08}} & \val{0.68}{0.45} & \val{1.68}{0.54} & \textbf{\valsig{0.35}{0.13}{**}} & $5.29e-03$ \\
 &  & HRV MAE(ms) & 172.34 & 28.05 & 14.72 & 17.11 & 29.67 & 42.07 & \underline{\val{9.28}{0.76}} & \val{10.51}{6.45} & \val{27.71}{10.86} & \textbf{\valsig{5.57}{2.34}{*}} & $4.46e-02$ \\
 &  & HR MAPE(\%) & 10.26 & 3.28 & 1.57 & 3.46 & 3.38 & 4.98 & \val{0.80}{0.08} & \underline{\val{0.75}{0.50}} & \val{1.84}{0.64} & \textbf{\valsig{0.40}{0.16}{**}} & $8.82e-03$ \\
 &  & HRV MAPE(\%) & 1118.19 & 61.21 & 34.57 & 31.67 & 94.50 & 230.03 & \textbf{\val{27.51}{2.59}} & \val{34.44}{8.28} & \val{73.19}{41.87} & \val{45.39}{59.81} & $5.92e-01$ \\
 &  & F1 & 0.356 & 0.926 & 0.955 & 0.933 & 0.944 & 0.930 & \val{0.989}{0.002} & \underline{\val{0.990}{0.006}} & \val{0.978}{0.006} & \textbf{\val{0.991}{0.004}} & $4.61e-01$ \\
 &  & Pre & 0.361 & 0.994 & 0.988 & 0.980 & 0.974 & 0.975 & \val{0.988}{0.002} & \val{0.990}{0.006} & \val{0.991}{0.005} & \val{0.989}{0.005} & $8.97e-01$ \\
 &  & Rec & 0.351 & 0.866 & 0.924 & 0.890 & 0.916 & 0.890 & \val{0.989}{0.002} & \val{0.990}{0.008} & \val{0.966}{0.008} & \val{0.993}{0.004} & $1.58e-01$ \\
\cmidrule(l){2-14}
& \multirow{7}{*}{\rotatebox{90}{\scriptsize\textbf{Capno}}}
 & HR MAE(bpm) & 4.48 & 2.99 & 0.69 & 1.12 & 8.52 & 6.26 & \val{0.72}{0.12} & \underline{\val{0.41}{0.19}} & \val{0.47}{0.16} & \textbf{\val{0.35}{0.28}} & $7.37e-01$ \\
 &  & HRV MAE(ms) & 67.80 & 15.17 & 8.60 & 9.90 & 111.56 & 39.71 & \val{11.80}{2.10} & \val{7.14}{2.33} & \underline{\val{6.73}{1.26}} & \textbf{\val{6.19}{4.11}} & $7.05e-01$ \\
 &  & HR MAPE(\%) & 6.41 & 2.57 & 0.83 & 1.45 & 8.32 & 5.54 & \val{0.86}{0.14} & \underline{\val{0.53}{0.21}} & \val{0.54}{0.19} & \textbf{\val{0.44}{0.33}} & $6.64e-01$ \\
 &  & HRV MAPE(\%) & 493.32 & 96.87 & 47.75 & 59.04 & 905.87 & 304.46 & \val{85.38}{15.52} & \underline{\val{37.57}{9.20}} & \val{41.96}{7.04} & \textbf{\val{34.58}{16.14}} & $7.61e-01$ \\
 &  & F1 & 0.334 & 0.839 & 0.928 & 0.875 & 0.599 & 0.903 & \val{0.989}{0.002} & \val{0.991}{0.004} & \underline{\val{0.992}{0.002}} & \textbf{\val{0.994}{0.003}} & $3.12e-01$ \\
 &  & Pre & 0.333 & 0.997 & 0.990 & 0.986 & 0.633 & 0.978 & \val{0.991}{0.003} & \val{0.989}{0.008} & \val{0.995}{0.001} & \val{0.995}{0.004} & $2.17e-01$ \\
 &  & Rec & 0.335 & 0.725 & 0.874 & 0.787 & 0.571 & 0.843 & \val{0.988}{0.001} & \val{0.994}{0.004} & \val{0.988}{0.003} & \val{0.993}{0.003} & $7.85e-01$ \\
\midrule
\multirow{14}{*}{\rotatebox{90}{\scriptsize\textbf{BCG}}} & \multirow{7}{*}{\rotatebox{90}{\scriptsize\textbf{Kansas}}}
 & HR MAE(bpm) & 30.57 & 4.90 & 21.33 & 142.28 & 2.91 & 4.77 & \underline{\val{2.74}{0.44}} & \val{2.71}{1.67} & \val{3.81}{0.56} & \textbf{\val{2.39}{1.92}} & $8.10e-01$ \\
 &  & HRV MAE(ms) & 310.85 & 60.10 & 110.03 & 92.73 & 65.59 & 77.69 & \underline{\val{51.95}{8.97}} & \val{55.31}{34.69} & \val{71.33}{15.50} & \textbf{\val{44.38}{30.27}} & $6.52e-01$ \\
 &  & HR MAPE(\%) & 50.51 & 7.81 & 34.80 & 223.55 & 4.34 & 7.52 & \val{4.06}{0.66} & \val{4.42}{2.77} & \val{5.39}{0.77} & \textbf{\val{3.94}{2.51}} & $8.06e-01$ \\
 &  & HRV MAPE(\%) & 481.80 & 61.21 & 230.22 & 86.49 & 68.92 & 125.44 & \val{31.97}{2.51} & \underline{\val{30.55}{17.19}} & \val{45.56}{7.61} & \textbf{\val{25.80}{21.45}} & $7.42e-01$ \\
 &  & F1 & 0.218 & 0.926 & 0.732 & 0.445 & 0.867 & 0.867 & \val{0.940}{0.013} & \val{0.935}{0.049} & \underline{\val{0.947}{0.012}} & \textbf{\val{0.957}{0.032}} & $4.77e-01$ \\
 &  & Pre & 0.177 & 0.892 & 0.620 & 0.289 & 0.814 & 0.825 & \val{0.931}{0.015} & \val{0.938}{0.042} & \val{0.964}{0.007} & \val{0.956}{0.026} & $5.05e-01$ \\
 &  & Rec & 0.284 & 0.964 & 0.896 & 0.983 & 0.927 & 0.913 & \val{0.950}{0.010} & \val{0.932}{0.060} & \val{0.930}{0.016} & \val{0.959}{0.041} & $4.94e-01$ \\
\cmidrule(l){2-14}
& \multirow{7}{*}{\rotatebox{90}{\scriptsize\textbf{Arrhy}}}
 & HR MAE(bpm) & 27.17 & 3.97 & 19.42 & 148.79 & 2.71 & 3.76 & \val{5.77}{0.81} & \val{3.34}{2.30} & \val{5.72}{1.51} & \textbf{\val{1.43}{0.95}} & $6.85e-02$ \\
 &  & HRV MAE(ms) & 338.98 & 57.44 & 128.91 & 67.36 & 75.96 & 68.12 & \val{92.44}{6.01} & \val{84.52}{32.98} & \val{130.43}{36.23} & \textbf{\valsig{36.85}{22.71}{*}} & $3.23e-02$ \\
 &  & HR MAPE(\%) & 41.00 & 6.40 & 30.93 & 233.37 & 3.94 & 5.88 & \val{8.97}{1.32} & \val{4.89}{2.73} & \val{7.77}{2.07} & \textbf{\val{2.14}{1.43}} & $8.09e-02$ \\
 &  & HRV MAPE(\%) & 1742.69 & 209.44 & 696.06 & 133.11 & 330.90 & 334.95 & \val{455.37}{95.15} & \val{193.74}{51.64} & \val{524.88}{200.78} & \textbf{\valsig{98.05}{64.46}{**}} & $2.33e-03$ \\
 &  & F1 & 0.583 & 0.899 & 0.801 & 0.466 & 0.915 & 0.897 & \val{0.934}{0.007} & \textbf{\val{0.949}{0.020}} & \val{0.929}{0.009} & \underline{\val{0.947}{0.034}} & $1.62e-01$ \\
 &  & Pre & 0.516 & 0.926 & 0.731 & 0.310 & 0.922 & 0.917 & \val{0.903}{0.016} & \val{0.936}{0.023} & \val{0.973}{0.004} & \val{0.940}{0.029} & $3.38e-01$ \\
 &  & Rec & 0.669 & 0.874 & 0.886 & 0.939 & 0.908 & 0.878 & \val{0.967}{0.009} & \val{0.963}{0.020} & \val{0.890}{0.018} & \val{0.954}{0.039} & $9.78e-02$ \\
\midrule
\multirow{7}{*}{\rotatebox{90}{\scriptsize\textbf{BSG}}} & \multirow{7}{*}{\rotatebox{90}{\scriptsize\textbf{ICU}}}
 & HR MAE(bpm) & 29.51 & 11.35 & 10.79 & 155.76 & 13.67 & 14.17 & \underline{\val{8.43}{1.64}} & \val{10.57}{1.25} & \val{16.25}{1.25} & \textbf{\val{8.17}{0.80}} & $7.89e-01$ \\
 &  & HRV MAE(ms) & 138.93 & 95.80 & 117.23 & 143.78 & 111.37 & 105.65 & \textbf{\val{67.26}{8.88}} & \val{84.52}{4.59} & \val{126.07}{6.24} & \underline{\valsig{83.67}{9.85}{*}} & $4.86e-02$ \\
 &  & HR MAPE(\%) & 43.07 & 12.68 & 13.79 & 164.92 & 14.70 & 15.25 & \textbf{\val{10.81}{2.51}} & \val{16.77}{2.16} & \val{18.25}{0.85} & \underline{\val{13.06}{1.63}} & $1.91e-01$ \\
 &  & HRV MAPE(\%) & 65.94 & 57.38 & 103.04 & 86.42 & 99.11 & 77.86 & \val{39.76}{2.67} & \textbf{\val{35.79}{6.26}} & \val{114.67}{26.67} & \underline{\val{39.04}{4.41}} & $7.91e-01$ \\
 &  & F1 & 0.297 & 0.834 & 0.146 & 0.516 & 0.730 & 0.745 & \textbf{\val{0.863}{0.023}} & \val{0.801}{0.027} & \val{0.820}{0.014} & \val{0.830}{0.016} & $5.66e-02$ \\
 &  & Pre & 0.244 & 0.863 & 0.137 & 0.362 & 0.746 & 0.778 & \val{0.848}{0.038} & \val{0.772}{0.030} & \val{0.953}{0.025} & \valsig{0.780}{0.029}{*} & $3.12e-02$ \\
 &  & Rec & 0.379 & 0.808 & 0.157 & 0.945 & 0.718 & 0.716 & \val{0.880}{0.013} & \val{0.832}{0.027} & \val{0.719}{0.013} & \val{0.887}{0.016} & $5.27e-01$ \\
\bottomrule
\end{tabular}%
}
\end{table}

\clearpage

\section{IMPACT OF MODEL SCALE}
\label{sec:scale}
To disentangle the performance benefits of our proposed architecture from mere parameter scaling, this section investigates the relationship between model capacity and Heart Rate (HR) estimation accuracy across the evaluated deep learning baselines. We systematically vary the trainable parameters of each model by adjusting channel dimensions and network depth to analyze the trade-off between computational cost and performance (measured by Mean Absolute Error, MAE).

\paragraph{Experimental Setup}
We implemented capacity scaling for three distinct architectures:
\begin{itemize}
    \item \textbf{FR-Net:} We scaled the model by modifying the base channel size $C \in \{32, 48, 64, 96, 128, 160\}$, resulting in a parameter size of $S \in \{2.5M, 4.3M, 6.7M, 12.6M, 20.4M, 29.9M\}$.
    \item \textbf{Unet++:} We explored variations in both both base filter count ($F$) and network depth ($D$), evaluating combinations of $(F, D) \in \{(8,3), (16,3), (16,4), (32,3), (32,4), (16,5)\}$, resulting in a parameter size of $S \in \{0.1M, 0.4M, 1.6M, 1.7M, 7.1M, 7.3M\}$.
    \item \textbf{CNN-SWT:} We evaluated six distinct channel progression configurations, ranging from a lightweight model ($\{8 \to 16 \to 32 \to 64 \to 128\}$) to a high-capacity variant ($\{48 \to 96 \to 192 \to 384 \to 768\}$), resulting in a parameter size of $S \in \{0.2M, 0.7M, 1.5M, 2.7M, 4.2M, 6.1M\}$.
\end{itemize}

\begin{figure*}[!t]

\centering

\begin{subfigure}[b]{0.32\textwidth}
    \centering
    \includegraphics[width=\textwidth]{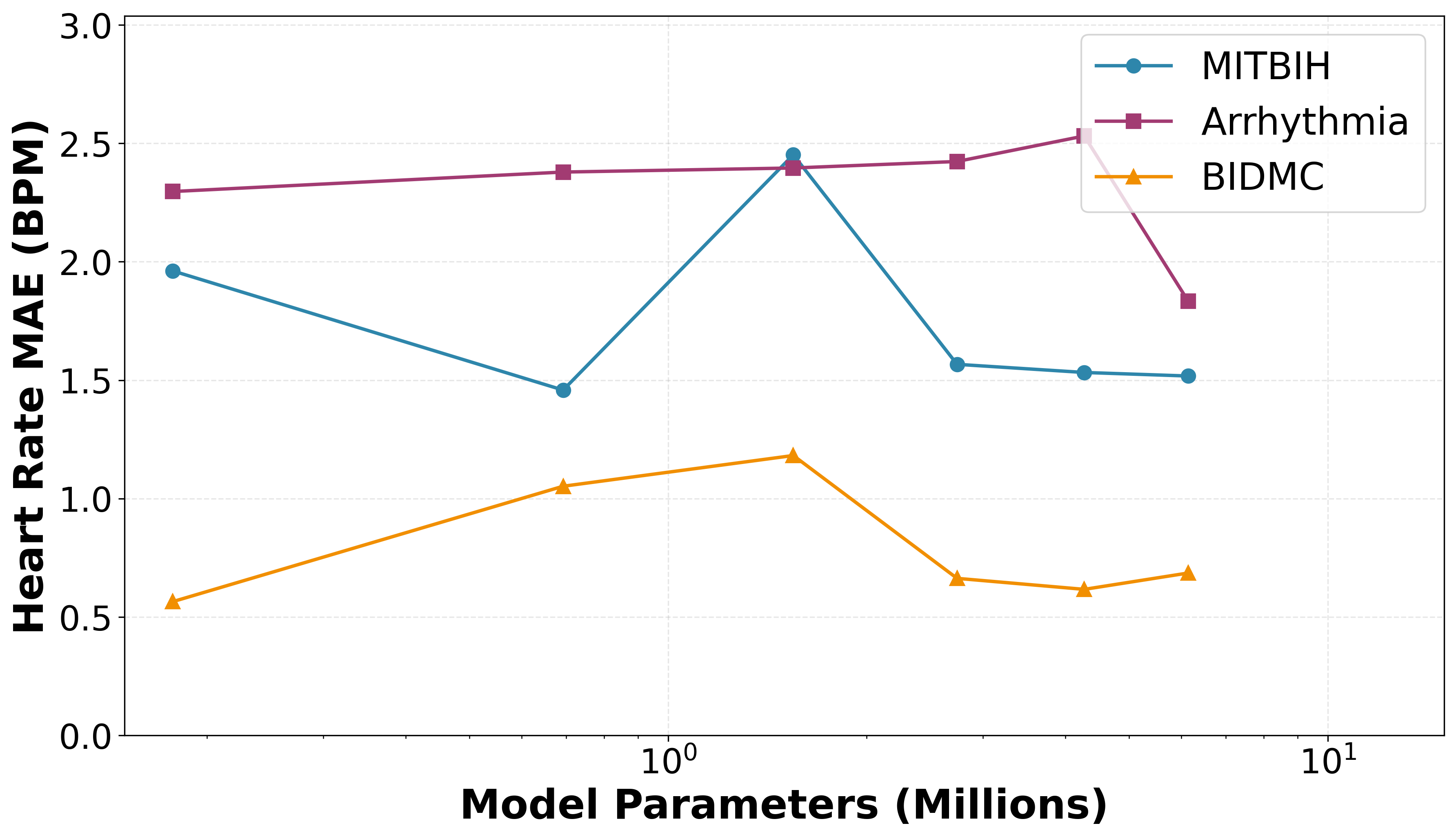}
    \caption{CNN-SWT}
    \label{fig:CNN-SWT_parameter}
\end{subfigure}
\hfill
\begin{subfigure}[b]{0.32\textwidth}
    \centering
    \includegraphics[width=\textwidth]{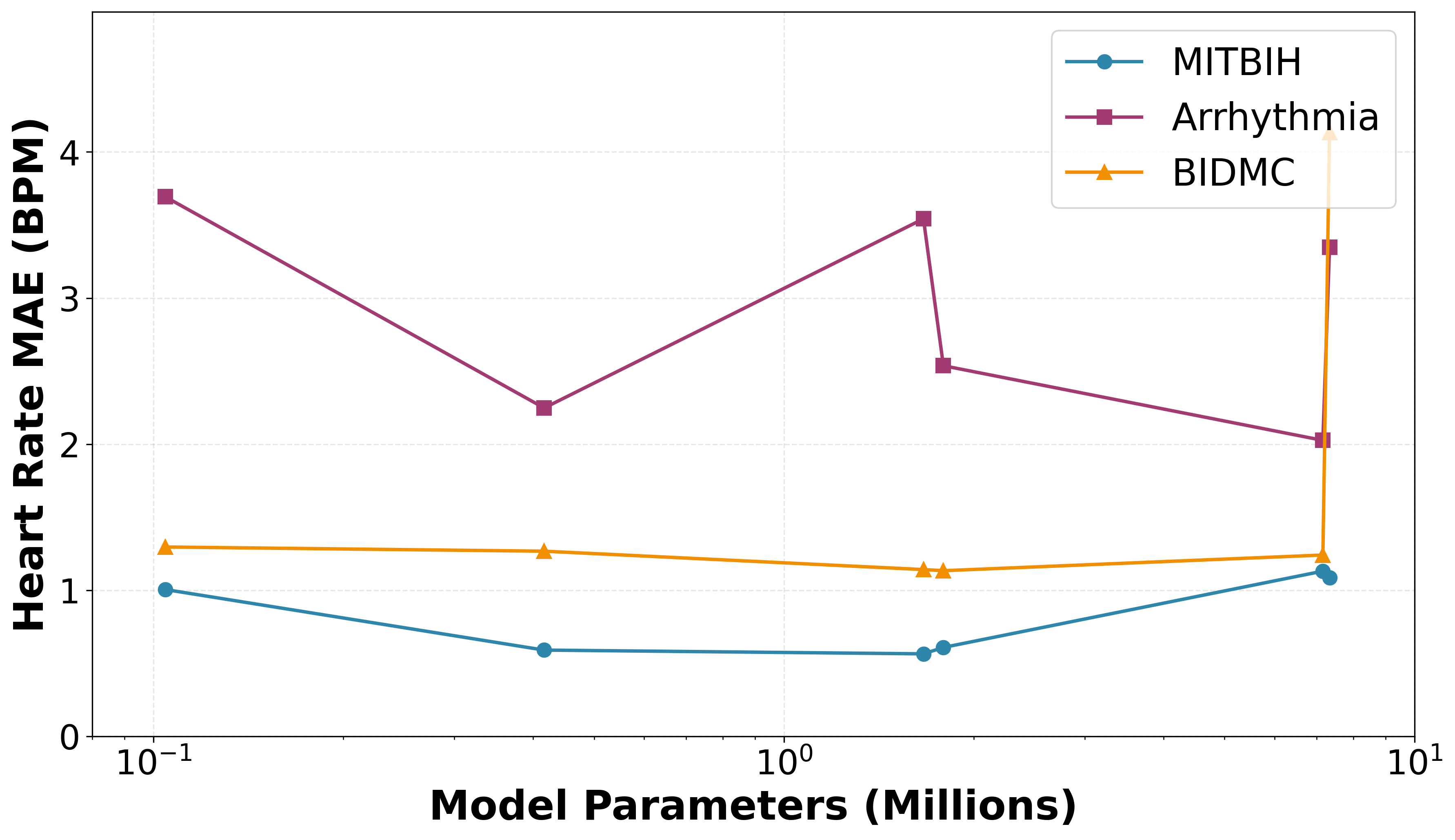}
    \caption{1D-Unet++}
    \label{fig:Unet++_parameter}
\end{subfigure}
\hfill
\begin{subfigure}[b]{0.32\textwidth}
    \centering
    \includegraphics[width=\textwidth]{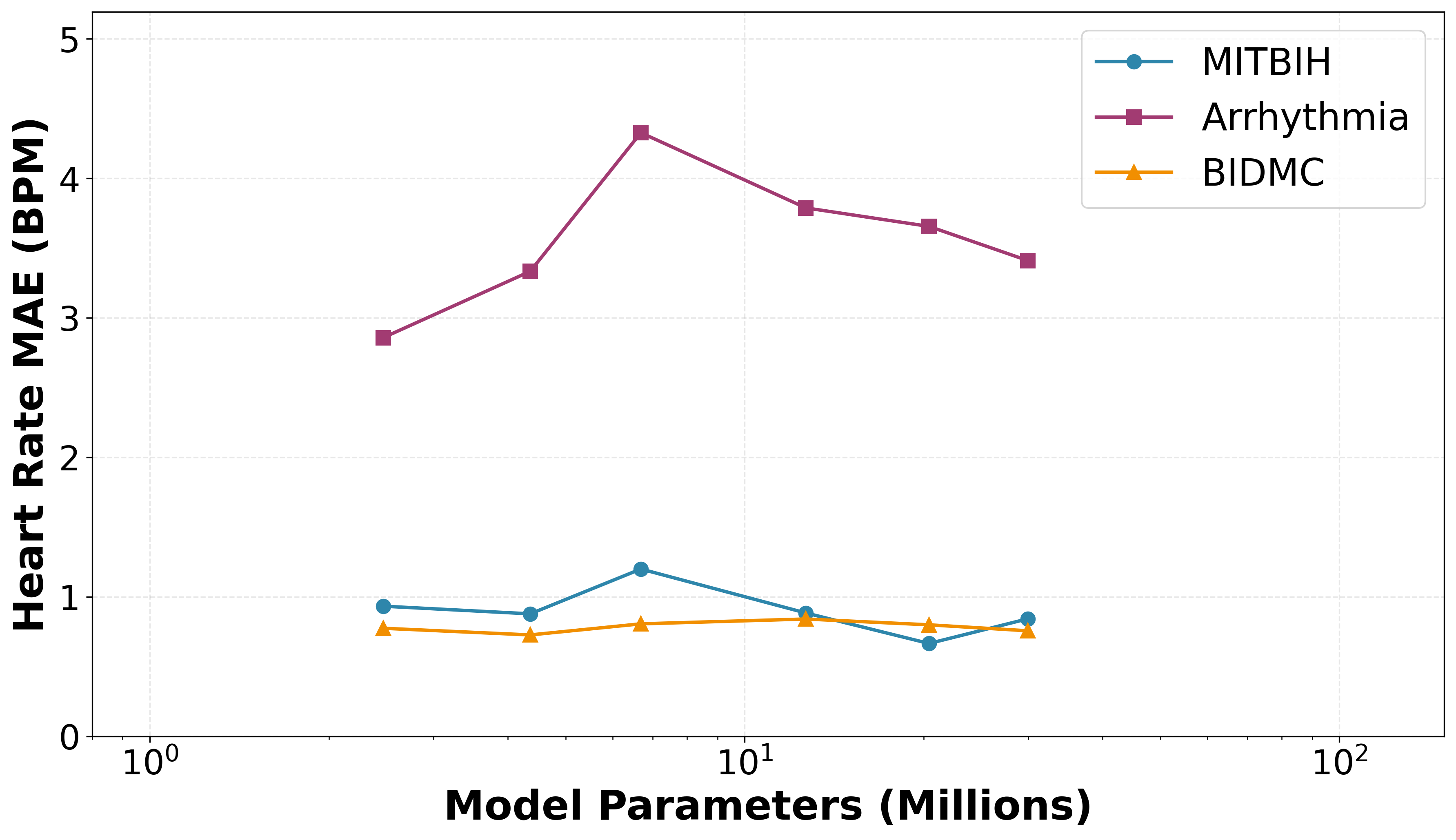}
    \caption{FR-Net}
    \label{fig:FR-Net_parameter}
\end{subfigure}
\caption{Impact of model scale on performance. We analyze the trade-off between the number of trainable parameters and Heart Rate Mean Absolute Error (HR MAE) for three architectures: (a) CNN-SWT, (b) Unet++, and (c) FR-Net.}
\Description{Three plots comparing model parameter counts against heart-rate mean absolute error for deep learning baselines.}
\label{fig:parameter_deep_learning}
\end{figure*}

\paragraph{Performance Analysis}
The relationship between parameter size and HR MAE for these baselines is illustrated in Fig.~\ref{fig:parameter_deep_learning}. Increasing model capacity can reduce MAE initially, but as scaling continues, we observe a performance plateau where improvements saturate and, in some instances, degrade. This trend indicates that indiscriminate parameter expansion in standard deep learning models yields diminishing returns and may introduce overfitting or optimization difficulties.

In contrast, Fig.~\ref{fig:scale} reports Peak-Detector performance when fine-tuning the same formulation on Qwen backbones ranging from 0.5B to 7B parameters under an identical protocol. On BCG Arrhythmia, larger backbones consistently improve HR/HRV estimation and detection quality in our tested range. 

\paragraph{Conclusion}
These findings suggest that the performance gap is not driven solely by model size; standard deep learning models appear unable to effectively utilize additional capacity beyond a certain threshold. Conversely, the Peak-Detector demonstrates a capability to leverage larger parameter spaces for continuous improvement. Furthermore, beyond quantitative superiority at scale, the Peak-Detector offers inherent explainability, a distinct advantage over black-box deep learning models.

\section{EVALUATION OF MODEL COMPRESSION TECHNIQUES}
\label{sec:compression_experiment}

To characterize the resource trade-offs of compressing Large Language Models (LLMs) in our setting, we conducted a comparative analysis of several memory-reduction strategies on the BCG Arrhythmia dataset. The primary goal of this experiment is to quantify the trade-off between reduced memory footprint and downstream physiological accuracy, rather than to claim mature real-time readiness for mobile or wearable deployment. All experiments were conducted with a batch size of 1 to provide a consistent evaluation setting across compression methods.

We evaluated three optimization strategies against a full-precision (FP16) baseline: Weight--Activation Quantization (W4A16), Structural Pruning (50\% sparsity), and Model Distillation using a 0.5B-parameter student model. The baseline Qwen2.5-3B model serves as the highest-fidelity configuration in this comparison, achieving a heart-rate mean absolute error (HR MAE) of 1.75 BPM with a memory footprint of 5.792 GB.

The experimental results, summarized in Table~\ref{tab:optimization_comparison}, highlight distinct trade-offs for each compression method:

\begin{itemize}
    \item \textbf{Weight--Activation Quantization (W4A16):} This method reduces memory usage by approximately 57\% (to 2.492 GB) by quantizing weights to 4-bit integers while maintaining 16-bit activations. This reduction is accompanied by a moderate increase in HR MAE to 2.85 BPM.
    
    \item \textbf{Structural Pruning (50\% Sparsity):} Pruning further reduces memory usage to 2.145 GB, but yields the largest degradation in accuracy among the tested methods, increasing HR MAE to 4.44 BPM. This result suggests that the model's performance on BCG signals is sensitive to the removal of structural capacity.
    
    \item \textbf{Model Distillation (0.5B):} Distillation achieves the largest memory reduction, lowering the footprint to 0.942 GB. However, the 0.5B student model produces an HR MAE of 3.26 BPM, indicating a substantial performance gap relative to the 3B teacher configuration under this task setting.
\end{itemize}

\begin{table}[htbp]
    \centering
    
    \caption{Experimental Comparison of LLM Optimization Methods for Mobile BCG Analysis (Qwen 2.5 3B). \textbf{HR MAE}: Heart Rate Mean Absolute Error.}
    \label{tab:optimization_comparison}
    \resizebox{0.7\columnwidth}{!}{
    \begin{tabular}{lcc}
        \toprule
        \textbf{Method} & \textbf{Training VRAM (GB)} & \textbf{Performance (HR MAE(s))} \\
        \midrule
        Baseline (FP16) & 5.792 & 1.75 BPM \\
        W4A16 & 2.492 & 2.85 BPM \\
        Pruning (50\% Sparsity)+W4A16 & 2.145 & 4.44 BPM \\
        Model Distillation (0.5B) & 0.942 & 3.26 BPM \\
        \bottomrule
    \end{tabular}}
\end{table}

Overall, these results show that aggressive compression can substantially reduce memory requirements, but at the cost of non-negligible degradation in HR accuracy. Under our test setting, this trade-off suggests that the current 3B configuration remains the strongest option when higher-fidelity peak analysis is required, while smaller compressed variants may be useful only in more resource-constrained scenarios where reduced memory usage is prioritized over accuracy. Accordingly, we interpret this analysis as evidence of a meaningful memory--accuracy trade-off in our LLM-based formulation, rather than as validation of strict real-time, always-on wearable deployment. In line with the broader deployment framing of this paper, the present results support \textit{Peak-Detector} most naturally as a selectively invoked, higher-fidelity analysis component, while more resource-aware variants remain an important direction for future work.

\section{VERIFICATION FRAMEWORK FOR LLM HALLUCINATION}
\label{sec:hallucination}

\begin{figure*}[!t]
\centering
\color{blue} 

\begin{subfigure}[b]{0.49\textwidth}
    \centering
    \includegraphics[width=\textwidth]{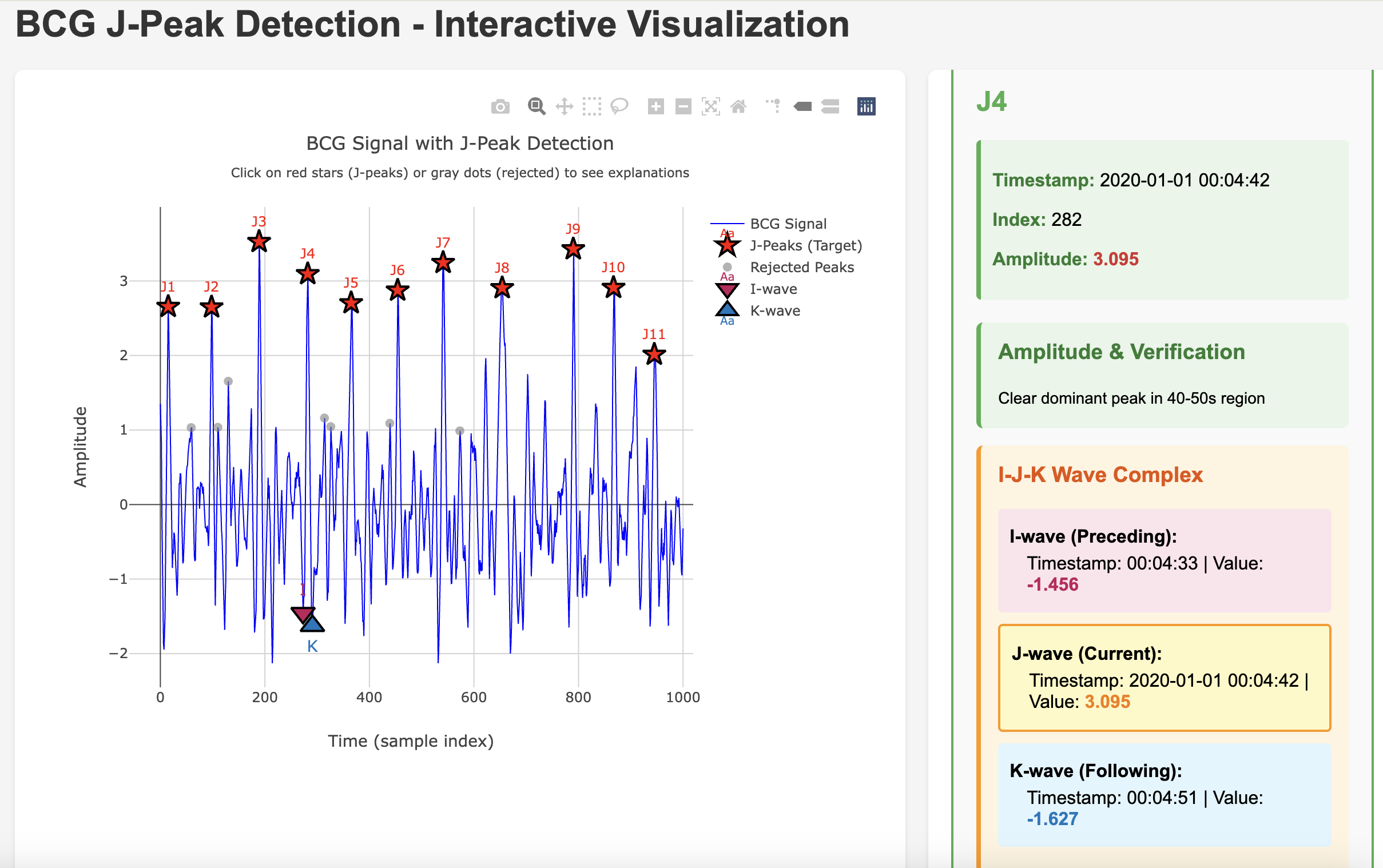}
    \caption{Target Peaks}
    \label{fig:Interactive_Visualization_of_Target_Peak}
\end{subfigure}
\hfill
\begin{subfigure}[b]{0.49\textwidth}
    \centering
    \includegraphics[width=\textwidth]{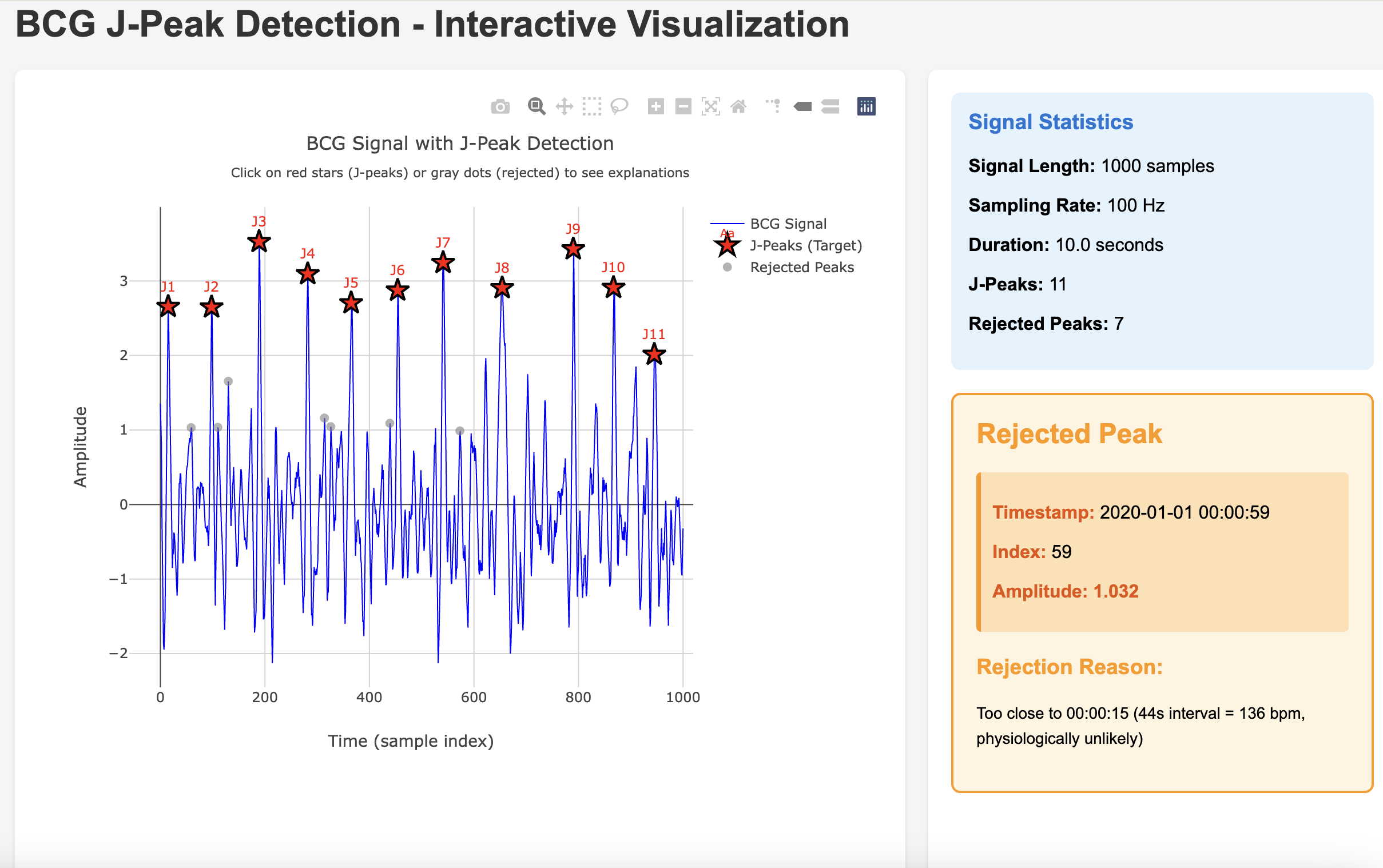}
    \caption{Rejected Peaks}
    \label{fig:Interactive_Visualization_of_Rejected_Peaks}
\end{subfigure}
\caption{Interactive verification interface. (a) Visualization of valid Target Peaks with morphological details. (b) Visualization of Rejected Peaks with associated exclusion criteria.}
\Description{Two interface screenshots showing accepted target peaks and rejected candidate peaks with explanations.}
\label{fig:Interactive_Visualization}
\end{figure*}

Hallucination ~\cite{Huang_2025} is a known risk when LLMs generate free-form rationales. In our data-construction pipeline, the teacher LLM is \textbf{not} responsible for peak labeling: it is explicitly provided with (i) the candidate extrema list from our Peak Representation and (ii) the \textit{ground-truth} peak list, and is only asked to explain \textit{why} the ground-truth peaks should be selected from the candidates. Therefore, hallucination risk is confined to the \textit{explanation text}, mainly in two forms: (a) \emph{objective inconsistencies} (e.g., timestamps/amplitudes/intervals that contradict the provided metadata), and (b) \emph{semantic inaccuracies} (e.g., logically unsupported rationale). To ensure explanations are faithful and usable for instruction tuning, we apply a three-stage verification pipeline before adding any teacher-generated explanation into the Peak-Explanation dataset:

\begin{enumerate}
    \item \textbf{Visualization-Assisted Verification Interface:} We parse the teacher LLM output and render it as an interactive overlay on the waveform (Fig.~\ref{fig:Interactive_Visualization}). The interactive user interface (UI) explicitly highlights \emph{LLM-selected target peaks} and \emph{LLM-rejected candidate peaks} together with their timestamps, indices, amplitudes, and the corresponding justification, enabling fast sanity checks by human reviewers.
    
    \item \textbf{Rule-Based Factual Consistency Check:} We \underline{automatically} verify objective claims and formatting in the explanation:  (i) the output peak list matches the provided ground truth; (ii) every referenced timestamp/index exists in the candidate list; (iii) stated amplitudes and inter-beat intervals are consistent with the underlying numeric values (within a small tolerance); and (iv) the output follows the required template. Explanations with any factual inconsistency are discarded.
    
    \item \textbf{Qualitative Human Review:} For semantic quality, human reviewers inspect the explanations using the UI and label them a \textit{Concise} (accurate and clear), \textit{Ambiguous} (technically correct but a bit vague), or \textit{Incorrect} (logical fallacies or misleading rationale).
\end{enumerate}

Specifically, in Step 1, as illustrated in Fig.~\ref{fig:Interactive_Visualization_of_Target_Peak}, the interface for valid target peaks aggregates critical metadata, including the timestamp, signal index, peak amplitude, and a detailed justification of amplitude dominance, alongside the associated fiducial complex morphology (e.g. I-J-K). Conversely, for filtered candidates, the UI displays the timestamp, index, and amplitude, while explicitly highlighting the rejection rationale, primarily focusing on physiological interval violations, shown in Fig.~\ref{fig:Interactive_Visualization_of_Rejected_Peaks}. Note that these peaks was rejected based on the LLM’s explanation for candidate peaks, which differs from the rejection criteria used in the verification framework.

\textbf{Verification Results.}
The quantitative outcomes of the verification framework are summarized in Table~\ref{tab:combined_verification_results}. The \textit{Rule-Based Factual Consistency Check} (Step 2) was applied to the full dataset to guarantee complete adherence to objective signal constraints. Conversely, for the \textit{Qualitative Human Review} (Step 3), we employed random sampling ($N=250$ per dataset) to derive a statistically representative estimate of semantic quality without the prohibitive cost of exhaustive manual annotation. In Step 2, we observed \textbf{0\% rejection rate} across all objective checks  the model demonstrated robust adherence to the ground truth signal attributes, indicating that the teacher LLM consistently grounded the numerical reasoning in the provided signal metadata without hallucinating objective parameters.

\begin{table}[ht]
\centering

\caption{Comprehensive Verification Results. The table shows the attrition of hallucinations during the automated check (Step 2) and the semantic quality distribution of the remaining explanations during the human review (Step 3).}
\label{tab:combined_verification_results}
\resizebox{0.5\columnwidth}{!}{
\begin{tabular}{lr}
\toprule
\textbf{Verification Stage / Category} & \textbf{Percentage} \\
\midrule
\multicolumn{2}{l}{\textbf{Step 2: Rule-Based Factual Consistency Check}} \\
\textit{(Percentages relative to total generated explanations)} & \\
\addlinespace
\hspace{3mm} Rejection: Temporal Mismatch & 0\% \\
\hspace{3mm} Rejection: Amplitude Discrepancy & 0\% \\
\hspace{3mm} Rejection: Physiological Interval Violation & 0\% \\
\hspace{3mm} Rejection: Formatting/Syntax Errors & 0\% \\
\cmidrule{2-2}
\textbf{Total Rejected (Step 2)} & \textbf{0\%} \\
\textbf{Passed to Human Review} & \textbf{100.0\%} \\
\midrule
\multicolumn{2}{l}{\textbf{Step 3: Qualitative Human Review}} \\
\textit{(Percentages relative to explanations passing Step 2)} & \\
\addlinespace
\hspace{3mm} \textbf{Concise} (Accurate and clear) & 91.7\% \\
\hspace{3mm} \textbf{Ambiguous} (Technically correct but vague) & 8.3\% \\
\hspace{3mm} \textbf{Incorrect} (Logical fallacies present) & 0\% \\
\cmidrule{2-2}
\textbf{Total Evaluated} & \textbf{100.0\%} \\
\bottomrule
\end{tabular}}
\end{table}

Step 3 evaluated the semantic utility of the generated content as shown in Table~\ref{tab:combined_verification_results}. The assessment revealed that the vast majority of explanations (91.7\%) were \textit{Concise}, providing accurate and clear justifications for the detected peaks. A minor fraction (8.3\%) were classified as \textit{Ambiguous}, where the reasoning was technically correct but lacked specificity. Notably, 0\% observed \textit{Incorrect} cases with $N=250$, suggesting that the model maintained logical coherence and avoided fabricating rationale that contradicted the signal morphology. These analyses suggests that semantic hallucination is rare in practice after our verification. As a result, we retain all teacher-generated explanations (including ambiguous ones) for training, given that exhaustive manual vetting is impossible.

\section{CROSS-DATASET GENERALIZATION ANALYSIS}
\label{sec:Cross_Modality}

To assess the generalization capabilities and robustness of Peak-Detector to domain shift, we conducted a comprehensive cross-dataset evaluation. Models were trained on a single source dataset and evaluated on all others. The F1-score results for Peak-Detector and the strongest baseline, FR-Net, are presented in Table~\ref{tab:cross_domain} and Table~\ref{tab:cross_domain_FRnet}, respectively.

\textbf{In-Domain Analysis.}
When evaluated on an in-domain but unseen dataset (e.g., training on MIT-BIH and testing on Incart), Peak-Detector's F1-score drops slightly from 0.9805 to 0.9580. This minor degradation, observed across all modalities, suggests that while the model learns modality-specific features effectively, it is also susceptible to dataset-specific biases, even within the same signal type.

\begin{table}[htbp!]
\centering
\caption{Cross-Dataset Generalization Performance (F1-score) for Peak-Detector. Rows indicate the training (source) dataset and columns indicate the testing (target) dataset.}
\label{tab:cross_domain}
\resizebox{0.65\columnwidth}{!}{
\begin{tabular}{l|cccccc}
\toprule
\textbf{Source / Target} & \textbf{MIT-BIH} & \textbf{Incart} & \textbf{BIDMC} & \textbf{Capnobase} & \textbf{Kansas} & \textbf{BCG Arry.} \\
\midrule
\textbf{MIT-BIH} & 0.9729 & 0.9580 & 0.6304 & 0.7466 & 0.4401 & 0.6440 \\
\textbf{Incart} & 0.9677 & 0.9805 & 0.2550 & 0.4494 & 0.3002 & 0.5376 \\
\textbf{BIDMC} & 0.3597 & 0.3726 & 0.9928 & 0.9564 & 0.3410 & 0.8040 \\
\textbf{Capnobase} & 0.4871 & 0.4129 & 0.9380 & 0.9912 & 0.3141 & 0.5957 \\
\textbf{Kansas} & 0.4217 & 0.5672 & 0.8656 & 0.8592 & 0.9544 & 0.8912 \\
\textbf{BCG Arry.} & 0.4230 & 0.5337 & 0.8639 & 0.7218 & 0.7556 & 0.9545 \\
\hline
\textbf{Combined Model} & 0.9681 & 0.9342 & 0.9877 & 0.9142 & 0.9379 & 0.8930 \\
\bottomrule
\end{tabular}}
\end{table}

\begin{table}[ht]
\centering
\caption{Cross-Dataset Generalization Performance (F1-score) for FR-Net. Rows indicate the training (source) dataset and columns indicate the testing (target) dataset.}
\label{tab:cross_domain_FRnet}
\resizebox{0.65\columnwidth}{!}{
\begin{tabular}{l|cccccc}
\toprule
\textbf{Source / Target} & \textbf{MIT-BIH} & \textbf{Incart} & \textbf{BIDMC} & \textbf{Capnobase} & \textbf{Kansas} & \textbf{BCG Arry.} \\
\midrule
\textbf{MIT-BIH} & 0.9802 & 0.8760 & 0.8436 & 0.1005 & 0.6273 & 0.8683 \\
\textbf{Incart} & 0.9570 & 0.9897 & 0.6838 & 0.0019 & 0.1070 & 0.5477 \\
\textbf{BIDMC} & 0.4312 & 0.3859 & 0.9679 & 0.9870 & 0.5050 & 0.8151 \\
\textbf{Capnobase} & 0.2320 & 0.0966 & 0.9458 & 0.9844 & 0.3229 & 0.2982 \\
\textbf{Kansas} & 0.6329 & 0.6717 & 0.8565 & 0.8858 & 0.9230 & 0.8795 \\
\textbf{BCG Arry.} & 0.8460 & 0.7228 & 0.8765 & 0.7792 & 0.8088 & 0.9120 \\
\hline
\textbf{Combined Model} & 0.9690 & 0.8816 & 0.8688 & 0.2467 & 0.9347 & 0.8879 \\
\bottomrule
\end{tabular}}
\end{table}

\textbf{Cross-Domain Analysis.}
Performance degrades more significantly in cross-domain scenarios, highlighting that models learn features specific to the physiological origin of each modality (electrical vs. optical vs. mechanical). For instance, Peak-Detector trained on ECG data (MIT-BIH) achieves an F1-score of 0.6304 when tested on PPG data (BIDMC). Crucially, even this attenuated performance is substantially higher than that of an untrained base model (F1-score of 0.1712, see Table~\ref{tab:ablation_study}), indicating that the framework learns a foundational, transferable knowledge of what constitutes a "peak." In comparison, FR-Net suffers a far more severe performance degradation in similar scenarios. When trained on MIT-BIH (ECG) and tested on Capnobase (PPG), FR-Net’s F1-score plummets to 0.1005, a near-total failure. In the same scenario, Peak-Detector maintains a functional F1-score of 0.7466. This stark difference suggests that while FR-Net excels at learning modality-specific features, its representations are less transferable, whereas Peak-Detector's language-based reasoning provides greater out-of-domain robustness.

\textbf{Performance of a Combined Model.}
The most compelling evidence for our framework's potential as a universal peak detector is the performance of the Combined Model, trained on a mixed dataset (MIT-BIH, BIDMC, and Kansas). As shown in the final row of Table~\ref{tab:cross_domain}, this single model achieves robust and high-quality performance across all six test sets, closely approaching the results of the specialized, individually trained models (e.g., 0.9877 on BIDMC vs. 0.9928 for the specialized model). In contrast, the combined FR-Net model exhibits poor generalization. While it performs well on the modalities included in its mixed training (ECG and BCG), its performance on the unseen PPG datasets is significantly worse than that of a specialized PPG-trained FR-Net (e.g., an F1-score of 0.2467 on Capnobase, compared to 0.9844 for the specialized model). This outcome strongly suggests that the Peak-Detector framework can be effectively scaled with diverse data to create a single, powerful, and truly generalizable model for cross-modal analysis—a capability not demonstrated by the conventional deep learning architecture.

\section{EXPLORATORY DEVICE FEASIBILITY AND DEPLOYMENT BOUNDARIES}
\label{sec:device_limitation}

To characterize the deployment boundaries of \textit{Peak-Detector}, we analyze the relationship between available hardware resources and LLM parameter scale across representative device tiers. As summarized in Table~\ref{tab:device-llm-capacity}, computational budgets vary substantially across the edge-to-server spectrum. Prior studies suggest that resource-constrained IoT platforms can support only relatively small models, whereas modern smartphones and other consumer devices may accommodate somewhat larger configurations under specialized hardware acceleration~\cite{nguyen2025evaluation,li2025apple,team2023gemini}. We include this comparison primarily to contextualize deployment trade-offs, rather than to claim mature real-time readiness for wearable or mobile execution.

The baseline \textit{Peak-Detector} uses a 3B-parameter model as the primary high-fidelity configuration evaluated in this work. We further examined compression through quantization, pruning, and distillation to study the memory--accuracy trade-off under more constrained settings (Appendix~\ref{sec:compression_experiment}). These experiments show that substantial reductions in memory footprint are possible, but they are accompanied by non-negligible degradation in downstream physiological accuracy. Accordingly, we do not interpret these results as evidence of comparable edge performance, strict real-time mobile deployment, or always-on suitability for battery-constrained wearables.

Instead, we view the current evidence as supporting \textit{Peak-Detector} most strongly in a selective, higher-fidelity analysis role within a tiered ubiquitous sensing pipeline. In such a setting, lightweight front-end methods can perform continuous low-power screening, while \textit{Peak-Detector} is invoked only for flagged windows, uncertain segments, or retrospective summaries that require interpretable rationales and more reliable peak localization. Developing more resource-aware variants that better preserve both detection quality and explanation quality under edge constraints remains an important direction for future work.

\begin{table}[ht]
\centering
\caption{LLM Inference Capacity by Device Category (Standard 4-bit Quantization)}
\label{tab:device-llm-capacity}
\resizebox{0.34\columnwidth}{!}{
\begin{tabular}{@{}ll@{}}
\toprule
\textbf{Device Category} & \textbf{Parameter} \\ \midrule
IoT / Edge~\cite{nguyen2025evaluation} & $\leq$ 1.5B \\
Smartphone~\cite{li2025apple,team2023gemini} & 1B -- 3B \\
Desktop Workstation~\cite{georganas2025pushing,blackstein2025rtx} & 12B -- 70B \\
Enterprise AI Server~\cite{makin2025sustainable,dong2025beyond} & 70B+ \\ \bottomrule
\end{tabular}}
\end{table}
\section{IMPLICATIONS FOR ARRHYTHMIC PEAK RECOVERY}
\label{sec:implication}

To examine the relevance of \textit{Peak-Detector} to arrhythmia-sensitive settings, we evaluated its ability to recover peaks associated with irregular cardiac events, which constitute an important upstream prerequisite for downstream rhythm analysis. We quantify this behavior using \textbf{Arrhythmic Peak Recall}, defined as the proportion of correctly identified arrhythmic beats within a 50\,ms tolerance window. The evaluation focuses on Atrial Premature Contractions (APCs) and Ventricular Premature Contractions (VPCs), both of which introduce irregular timing patterns that can challenge peak localization.

\begin{figure}[ht]
\centering
\begin{minipage}{0.45\textwidth}
    \centering
    \includegraphics[width=\textwidth]{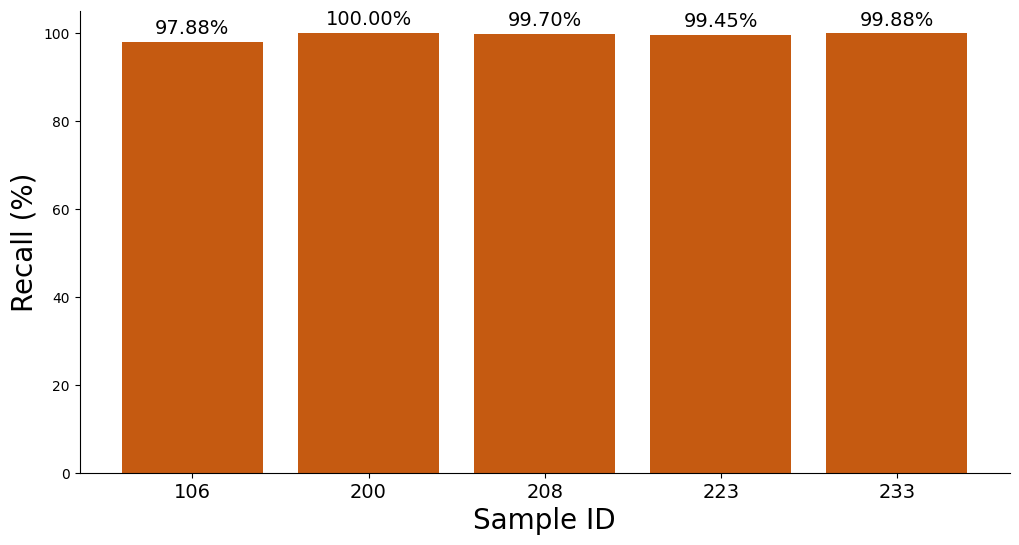}
    \caption{MIT-BIH Arrhythmia Recall}
    \label{fig:arrhythmia Recall}
\end{minipage}
\hfill
\begin{minipage}{0.51\textwidth}
    \centering
    \includegraphics[width=\textwidth]{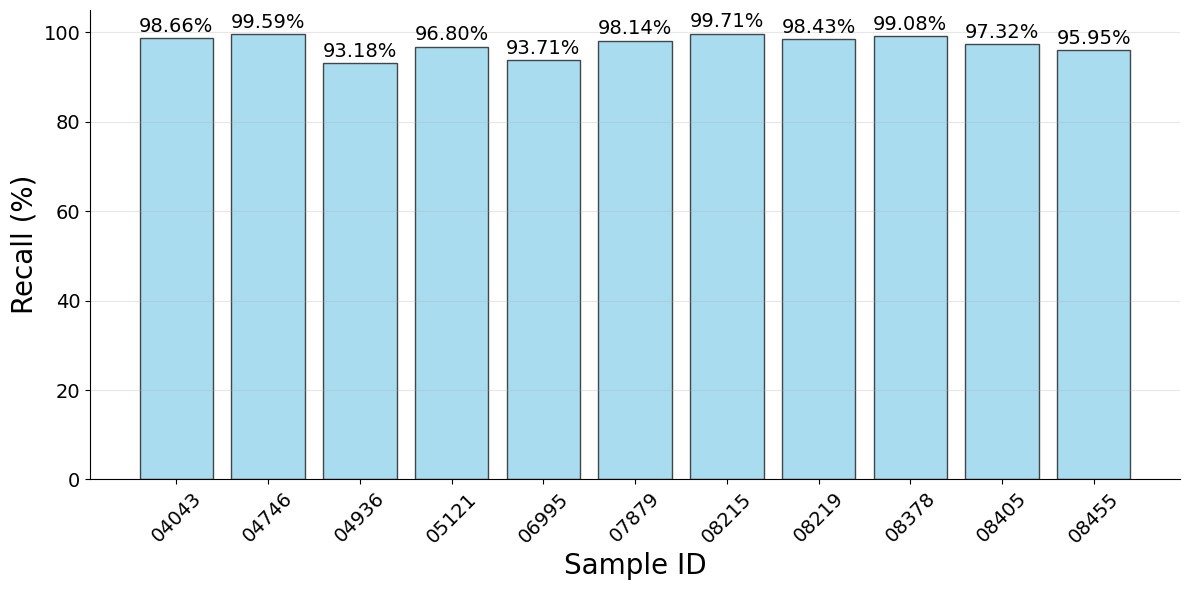}
    \caption{MIT-BIH Atrial Fibrillation Recall}
    \label{fig:AFib recall}
\end{minipage}
\Description{Two recall plots showing arrhythmic peak recovery on MIT-BIH arrhythmia and atrial fibrillation datasets.}
\end{figure}

To assess both in-distribution and out-of-distribution behavior, the model was trained exclusively on the \textbf{MIT-BIH Arrhythmia Database} and evaluated on both the same dataset and the unseen \textbf{MIT-BIH Atrial Fibrillation Database}. As illustrated in Figures~\ref{fig:arrhythmia Recall} and~\ref{fig:AFib recall}, \textit{Peak-Detector} achieves high arrhythmic peak recall across subjects. In the in-distribution setting, recall exceeds 99\% for most subjects, while on the unseen atrial fibrillation dataset it remains above 97\% for most subjects.

These results suggest that \textit{Peak-Detector} can maintain strong peak recovery performance even in the presence of irregular rhythms, and that the learned representation transfers reasonably well to unseen arrhythmia-related conditions. We interpret this analysis as evidence of downstream relevance rather than as a full validation of downstream arrhythmia analysis. In particular, while accurate peak recovery is an important prerequisite for rhythm analysis, end-to-end arrhythmia classification involves additional considerations beyond the scope of the present work.

\section{VISUALIZATION}
\label{app: visualization}

\begin{table}[htbp]
\centering
\caption{Comparison of J-Peak Detection Results and Explanations from Teacher LLM and Fine-tuned Peak-Detector}
\label{tab: explanation}
\scriptsize
\resizebox{1\columnwidth}{!}{
\begin{tabular}{|p{8cm}|p{8.5cm}|}
\hline
\textbf{Peak-Detector Generated Output} & \textbf{Teacher LLM Generated Output} \\
\hline
J: [2020-01-01 00:00:17, 2020-01-01 00:01:51, 2020-01-01 00:03:22, 2020-01-01 00:04:57, 2020-01-01 00:06:34, 2020-01-01 00:08:10, 2020-01-01 00:09:47, 2020-01-01 00:11:25, 2020-01-01 00:13:06, 2020-01-01 00:14:44, 2020-01-01 00:16:20] & J: [2020-01-01 00:00:17, 2020-01-01 00:01:51, 2020-01-01 00:03:22, 2020-01-01 00:04:57, 2020-01-01 00:06:34, 2020-01-01 00:08:10, 2020-01-01 00:09:47, 2020-01-01 00:11:25, 2020-01-01 00:13:06, 2020-01-01 00:14:44, 2020-01-01 00:16:20]\\
\hline
\textbf{Explanation:}

\textbf{Step 1: Morphological Characteristics}

The J-peak is defined as the **largest headward (positive) wave** occurring late in systole. Examining the detected peaks, I identify candidates with prominent positive amplitudes:
\begin{itemize}
\item 2020-01-01 00:00:17: \textbf{2.915030} \checkmark
\item 2020-01-01 00:01:51: \textbf{2.786782} \checkmark
\item 2020-01-01 00:03:22: \textbf{2.842781} \checkmark
\item 2020-01-01 00:04:57: \textbf{3.200422} \checkmark (highest amplitude)
\item 2020-01-01 00:06:34: \textbf{2.703152} \checkmark
\item 2020-01-01 00:08:10: \textbf{3.031469} \checkmark
\item 2020-01-01 00:09:47: \textbf{2.775080} \checkmark
\item 2020-01-01 00:11:25: \textbf{2.718626} \checkmark
\item 2020-01-01 00:13:06: \textbf{2.302004} \checkmark
\item 2020-01-01 00:14:44: \textbf{2.987384} \checkmark
\item 2020-01-01 00:16:20: \textbf{3.164307} \checkmark
\end{itemize}
These peaks exhibit amplitudes >2.0, significantly higher than other positive deflections (<1.5), indicating they represent true cardiac events rather than noise.

\textbf{Step 2: Temporal Relationships and Physiological Intervals}

\textbf{Inter-beat intervals (RR-equivalents):}
\begin{itemize}
\item 00:00:17 $\rightarrow$ 00:01:51: \textbf{94 seconds} (0.94s)
\item 00:01:51 $\rightarrow$ 00:03:22: \textbf{91 seconds} (0.91s)
\item 00:03:22 $\rightarrow$ 00:04:57: \textbf{95 seconds} (0.95s)
\item 00:04:57 $\rightarrow$ 00:06:34: \textbf{97 seconds} (0.97s)
\item 00:06:34 $\rightarrow$ 00:08:10: \textbf{96 seconds} (0.96s)
\item 00:08:10 $\rightarrow$ 00:09:47: \textbf{97 seconds} (0.97s)
\item 00:09:47 $\rightarrow$ 00:11:25: \textbf{98 seconds} (0.98s)
\item 00:11:25 $\rightarrow$ 00:13:06: \textbf{101 seconds} (1.01s)
\item 00:13:06 $\rightarrow$ 00:14:44: \textbf{98 seconds} (0.98s)
\item 00:14:44 $\rightarrow$ 00:16:20: \textbf{96 seconds} (0.96s)
\end{itemize}

Mean interval: $\sim$0.97s (Heart rate: $\sim$62 bpm)

This indicates bradycardia with arrhythmia, consistent with the signal description. The variability in intervals (91--101 seconds) confirms arrhythmic pattern.

\textbf{Step 3: Signal-to-Noise Ratio and Amplitude Criteria}

Amplitude threshold analysis:
\begin{itemize}
\item True J-peaks: 2.915--3.200 (SNR excellent)
\item Other positive peaks: 0.096--1.950 (likely I-waves, K-waves, or noise)
\item Negative deflections: $-0.191$ to $-2.100$ (I-waves and K-waves)
\end{itemize}

The 2:1 amplitude ratio between true J-peaks and other features provides clear discrimination.

\textbf{Step 4: Waveform Context (I-J-K Complex)}

Each identified J-peak follows the characteristic pattern:
\begin{itemize}
\item \textbf{Preceding I-wave} (negative deflection): e.g., $-1.943758$ before $2.915030$
\item \textbf{J-peak} (dominant positive): The selected peaks
\item \textbf{Following K-wave} (negative deflection): e.g., $-1.597640$ after $2.786782$
\end{itemize}

Example I-J-K complex (00:04:57 J-peak):
\begin{itemize}
\item I-wave: 00:04:49 ($-1.795014$)
\item J-peak: 00:04:57 ($3.200422$)
\item K-wave: 00:05:06 ($-1.629486$)
\end{itemize}

\textbf{Step 5: Exclusion of False Positives}

Why smaller positive peaks are NOT J-peaks:
\begin{itemize}
\item 00:00:49 (1.614529): Too small, occurs too soon after 00:00:17 (31s interval = 194 bpm, physiologically impossible)
\item 00:02:21 (1.366664): Low amplitude, inconsistent timing
\item 00:02:55 (1.562412): Only 54s after previous J-peak, too frequent
\item 00:07:52 (0.997150): Below amplitude threshold, irregular pattern
\end{itemize}

Why negative peaks are NOT J-peaks:
\begin{itemize}
\item All negative deflections represent I-waves or K-waves, part of the I-J-K complex
\end{itemize}
&
\textbf{Explanation:}

\textbf{Step 1: Morphological Characteristics} 

The selected J-peaks represent the largest positive (headward) deflections in the BCG signal, consistent with the physiological definition of the J-wave as the most prominent systolic component:

Peak Time: 00:00:17, Amplitude: 2.915030 \checkmark J-peak -- Dominant positive peak in first cardiac cycle

Peak Time: 00:01:51, Amplitude: 2.786782 \checkmark J-peak -- Highest amplitude in surrounding 10s window

Peak Time: 00:03:22, Amplitude: 2.842781 \checkmark J-peak -- Clear dominant positive deflection

Peak Time: 00:04:57, Amplitude: 3.200422 \checkmark J-peak -- Maximum amplitude in entire signal

Peak Time: 00:06:34, Amplitude: 2.703152 \checkmark J-peak -- Prominent systolic wave

Peak Time: 00:08:10, Amplitude: 3.031469 \checkmark J-peak -- Second highest amplitude overall

Peak Time: 00:09:47, Amplitude: 2.775080 \checkmark J-peak -- Dominant in cycle

Peak Time: 00:11:25, Amplitude: 2.718626 \checkmark J-peak -- Clear J-wave morphology

Peak Time: 00:13:06, Amplitude: 2.302004 \checkmark J-peak -- Highest in irregular segment

Peak Time: 00:14:44, Amplitude: 2.987384 \checkmark J-peak -- Strong systolic component

Peak Time: 00:16:20, Amplitude: 3.164307 \checkmark J-peak -- Dominant terminal peak

Rejected candidates (smaller positive peaks):

00:00:49 (1.614529), 00:01:22 (0.887839), 00:02:21 (1.366664) -- All significantly smaller than nearby dominant peaks. These represent either I-waves, K-waves, or noise artifacts.

\textbf{Step2: Temporal Relationships - Physiological RR Intervals}

Analysis of inter-peak intervals reveals arrhythmic pattern consistent with the signal description:

Interval Peak 1$\rightarrow$2: Duration 94s, Heart Rate 64 bpm -- Bradycardic/irregular

Interval Peak 2$\rightarrow$3: Duration 91s, Heart Rate 66 bpm -- Irregular rhythm

Interval Peak 3$\rightarrow$4: Duration 95s, Heart Rate 63 bpm -- Consistent with arrhythmia

Interval Peak 4$\rightarrow$5: Duration 97s, Heart Rate 62 bpm -- Variable intervals

Interval Peak 5$\rightarrow$6: Duration 96s, Heart Rate 63 bpm -- Arrhythmic pattern

Interval Peak 6$\rightarrow$7: Duration 97s, Heart Rate 62 bpm -- Irregular

Interval Peak 7$\rightarrow$8: Duration 98s, Heart Rate 61 bpm -- Bradyarrhythmia

Interval Peak 8$\rightarrow$9: Duration 100s, Heart Rate 60 bpm -- Extended interval

Interval Peak 9$\rightarrow$10: Duration 98s, Heart Rate 61 bpm -- Variable

Interval Peak 10$\rightarrow$11: Duration 96s, Heart Rate 63 bpm -- Irregular rhythm

Key observations:

Mean interval: $\sim$96.2s (approximately 62 bpm)

Coefficient of variation: Low, but intervals show arrhythmic irregularity

No intervals $<$40s or $>$120s, ruling out extreme tachycardia/bradycardia

Pattern consistent with sinus arrhythmia or atrial fibrillation with controlled ventricular response

\textbf{Step3: Signal-to-Noise Ratio and Amplitude Criteria}

Amplitude threshold analysis:

Selected J-peaks: 2.30 to 3.20 (mean: 2.84)

Rejected positive peaks: 0.06 to 1.64 (mean: 0.68)

Clear bimodal distribution with $>$1.5 amplitude separation

Signal-to-noise ratio: $\sim$4:1 for true J-peaks vs. artifacts

Negative deflections (I-waves and K-waves):

Range: $-0.19$ to $-2.10$

Properly excluded as J-peak is defined as positive/headward deflection

These represent I-waves (preceding J) and K-waves (following J)

\textbf{Step4: Contextual Waveform Features - IJK Complex}

Each identified J-peak shows characteristic IJK morphology:

Example cycle (00:04:57 J-peak):

I-wave: 00:04:49 ($-1.795014$) -- Negative deflection preceding J

J-peak: 00:04:57 (3.200422) -- Dominant positive peak \checkmark

K-wave: 00:05:06 ($-1.629486$) -- Negative deflection following J

Pattern verification across all cycles:

1. Each J-peak is preceded by negative I-wave (0.5--1.0s before)

2. Each J-peak is followed by negative K-wave (0.5--1.5s after)

3. Smaller positive deflections between cycles represent:

L-waves (late diastolic oscillations)

M-waves (atrial components)

Respiratory artifacts
\\
\hline
\end{tabular}}
\end{table}



\end{document}